\documentclass{article}
\pdfoutput=1
\PassOptionsToPackage{numbers}{natbib}

\usepackage[preprint,nonatbib]{neurips_2020}

\usepackage[utf8]{inputenc} 
\usepackage[T1]{fontenc}    
\usepackage{hyperref}
\usepackage{url}            
\usepackage{booktabs}       
\usepackage{amsfonts}       
\usepackage{nicefrac}       
\usepackage{microtype}      
\usepackage{graphicx}
\usepackage{lipsum}
\usepackage{graphics}
\usepackage{comment}
\usepackage{amsmath,amssymb} 
\usepackage{color}
\usepackage{times}
\usepackage{epsfig}
\usepackage{grffile}
\usepackage{tabularx}
\usepackage{algorithm}
\usepackage{algorithmic}
\usepackage[algo2e,ruled,vlined]{algorithm2e}
\usepackage{multirow}
\usepackage{subcaption}
\usepackage{appendix}

\title{PCAAE: Principal Component Analysis Autoencoder for organising the latent space of generative networks}

\author{
  Chi-Hieu Pham, Saïd Ladjal\thanks{Alphabetical order, equal contribution of authors}, Alasdair Newson\footnotemark[1] \\
  LTCI, Télécom Paris, Université Paris Saclay\\
  19 Place Marguerite Perey, 91120 Palaiseau\\
  \texttt{$\{$ chi-hieu.pham, said.ladjal, anewson $\}$@telecom-paris.fr} \\
}

\begin{document}
\maketitle

\begin{abstract}

Autoencoders and generative models produce some of the most spectacular deep learning results to date. However, understanding and controlling the latent space of these models presents a considerable challenge. Drawing inspiration from principal component analysis and autoencoder, we propose the Principal Component Analysis Autoencoder (PCAAE). This is a novel autoencoder whose latent space verifies two properties. Firstly, the dimensions are organised in decreasing importance with respect to the data at hand. Secondly, the components of the latent space are statistically independent. We achieve this by progressively increasing the latent space during training, and with a covariance loss applied to the latent codes. The resulting autoencoder produces a latent space which separates the intrinsic attributes of the data into different components of the latent space, in a completely unsupervised manner. We also describe an extension of our approach to the case of powerful, pre-trained GANs. We show results on both synthetic examples of shapes and on a state-of-the-art GAN. For example, we are able to separate the color shade scale of hair and skin, pose of faces and the gender in the CelebA, without accessing any labels. We compare the PCAAE with other state-of-the-art approaches, in particular with respect to the ability to disentangle attributes in the latent space. We hope that this approach will contribute to better understanding of the intrinsic latent spaces of powerful deep generative models.

\end{abstract}

\section{Introduction}
\label{sec:intro}

The recent impressive results of deep generative models and autoencoder-type models rely on a core idea: uncovering a compact, powerful latent space where the original high-dimensional data can be better synthesised or manipulated. Some of the most astounding recent synthesis results in deep learning have come from generative models such as generative autoencoders \cite{Kingma2014Auto,van2016conditional,sonderby2016ladder,tolstikhin2018wasserstein,huang2018introvae,heljakka2020towards} or Generative Adversarial Networks (GANs) \cite{goodfellow2014generative,salimans2016improved,karras2017progressive,srivastava2017veegan,karras2018style,choi2017stargan,zhu2017toward,pham2019simultaneous}. However, in spite of their undoubted efficiency, the latent spaces created by these models are difficult to interpret. In particular, a common problem is that these spaces are \emph{entangled}: several image characteristics are often combined into one dimension of the latent space, making navigation and understanding difficult. Certain previous approaches have attempted to disentangle the space in a semi-supervised manner, that requires knowledge about the true underlying factors of the data \cite{kingma2014semi,reed2014learning,mathieu2016disentangling,siddharth2017learning, denton2017unsupervised,hsu2017unsupervised}. However, we would like to achieve this organisation of the latent space in a non-supervised approach, letting the data tell us what variability exists in the database.

\begin{figure}[t]
    \centering
    \includegraphics[width=0.7\textwidth]{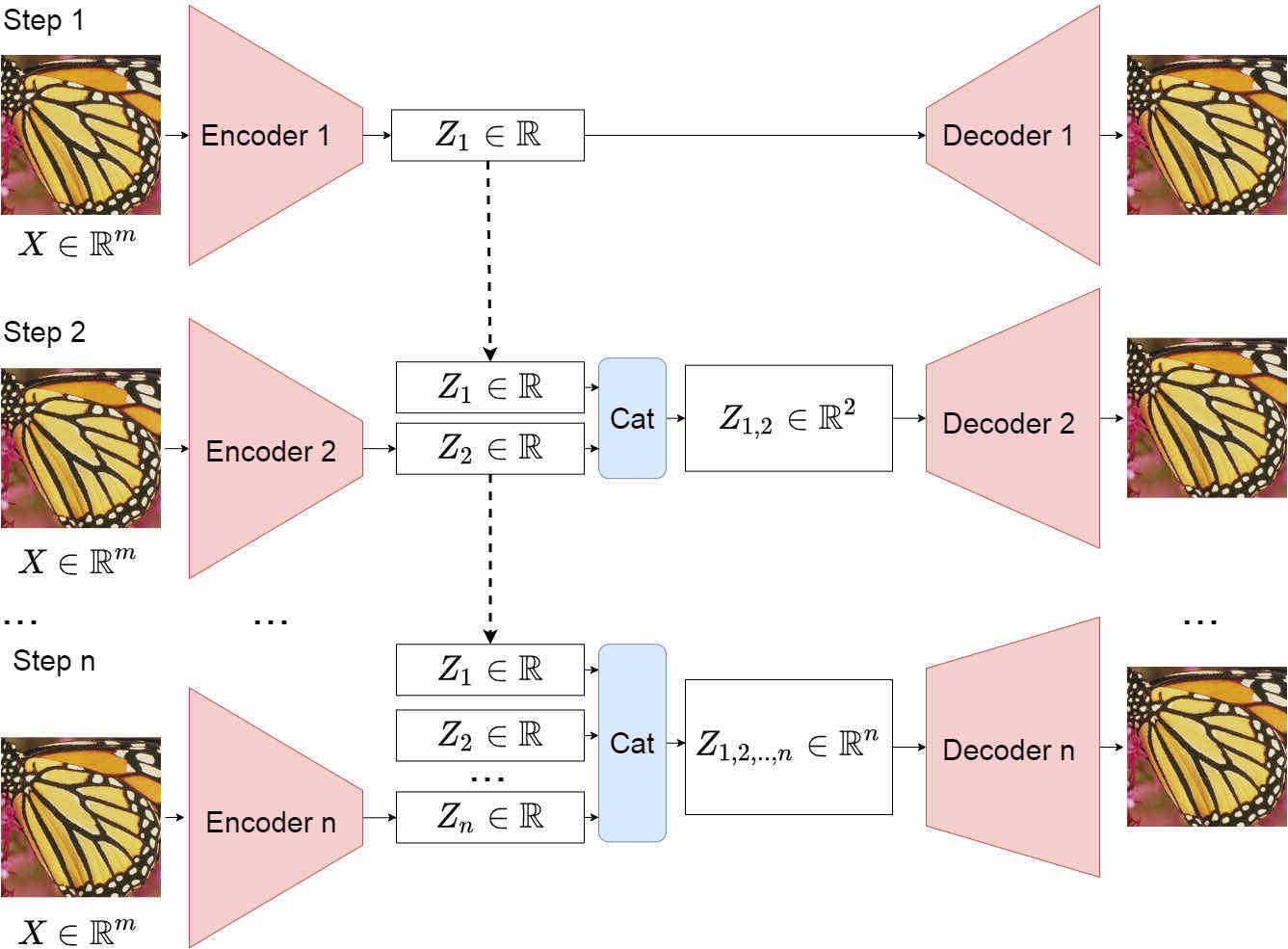}
    \label{fig:autoencoder}
    \caption{Architecture of our PCAAE. At the $n^{th}$ step, the PCAAE takes all previous pre-trained encoders and ignores those decoders. The parameters of these encoders are fixed. Their output are concatenated with those of the $n^{th}$ encoder and trained with the $n^{th}$ decoder.}
    \vspace{-3pt}
    \label{fig:diagram}
    \vspace{-12pt}
\end{figure}

In this work, we propose a network which we refer to as the ``Principal Component Analysis Autoencoder'' (PCAAE). An autoencoder is a neural network consisting of two sub-networks : an encoder and a decoder. These networks project data to and from the lower-dimensional latent space. Ideally, we would like this latent space to be interpretable and navigable. We propose to achieve this by creating an autoencoder which shares some of the desirable characteristics of the PCA. The classical PCA is a linear transformation to a space with two main properties. Firstly, the axes are organised in order of decreasing variability. So, along the first axis lies the greatest variability of the data, along the second orthogonal axis lies the second-greatest variability, and so on and so forth. Secondly, the axes are orthogonal to each other, which is necessary for interpretation and manipulation. Ideally, we would like to have the best of both worlds, ie. the power of a non-linear transformation (a neural network here) with the aforementioned properties of PCA. This is precisely the objective of this work. More precisely, our goal is to propose an autoencoder with the following two properties: i) the latent space components (axes) are ordered in terms of decreasing importance and ii) each component of a code is statistically independent from the other components.

To achieve this, we start by training an autoencoder with a latent space of size 1. Once this is trained, we fix the values of this first element in the latent space, and train an autoencoder with a latent space of size 2, where only the second component is trained. At each step, the decoder is discarded, and a new one is trained from scratch. This continues until we reach the required latent space size (see Figure~\ref{fig:diagram} for an illustration of this approach). Secondly, we add a latent space covariance loss term to the autoencoder loss to ensure that each component is statistically independent. If the intrinsic characteristics of the data are distributed independently throughout the dataset, then this will be reflected in the PCAAE latent space.  The final objective is to create an autoencoder whose latent space efficiently separates (disentangles) independent characteristics of the data being considered. For example, this could be properties such as size, shape or colour, or more high-level characteristics such as gender or hair colour in the case of images of faces. We achieve this without any reference to labels relative to these characteristics. Instead, we aim to discover the latter in a completely unsupervised fashion, through the data itself.

\begin{figure}[htb!]
    \centering
    \includegraphics[width=\textwidth]{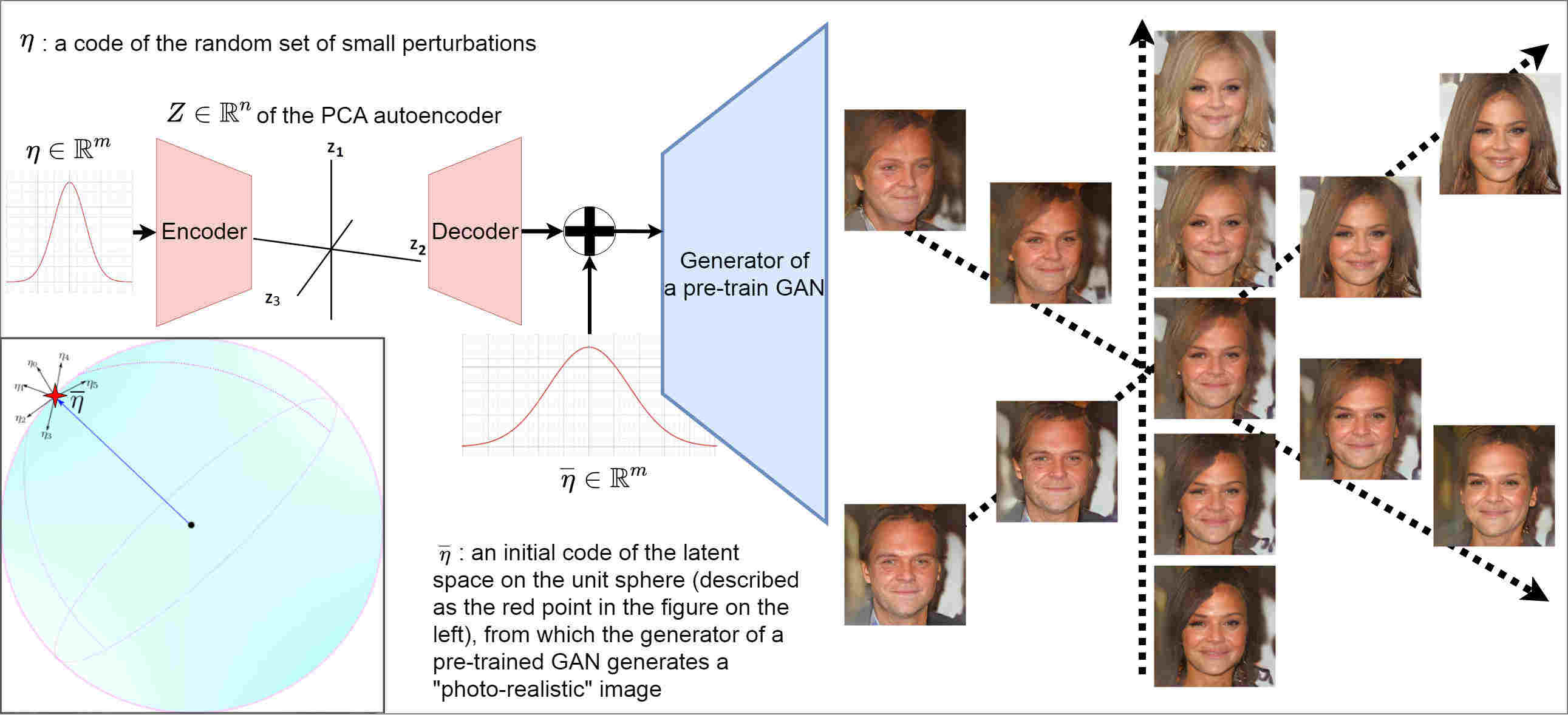}
    \caption{PCAAE is applied for navigating in the latent space of a pre-trained GAN. Each component of the PCAAE attempts to control one attribute of generated images.}
    \label{fig:PCA_GAN}
    \vspace{-10pt}
\end{figure}

To summarise, in this paper we propose the following contributions: 
\begin{itemize}
\item An algorithm to create a autoencoder with a latent space where the components of the latent code are ordered in terms of decreasing importance to the data; 
\item We use a covariance loss term to encourage the components of the latent space to be statistically independent to decrease entanglement; 
\item  We show how the PCAAE can be used to organise and disentagle the latent space of a pre-trained generative network such as a GAN.
\end{itemize}

In other words, we wish both to impose an order on and disentangle the latent space. We demonstrate the efficiency of our autoencoder on synthetic examples of images of geometric shapes as well as on the more complex data of the CelebA dataset. In the first case, we show that the resulting autoencoder retrieves meaningful axes that can be manipulated to change different geometric characeristics (size, rotation) of the shapes.
In the second, we automatically discover properties such as hair colour, gender and pose. We emphasise that this is done in a completely unsupervised manner, without any access to the labellings of these characteristics. While other approaches to disentanglement exist, they mainly focus on supervised settings where the labels are available. In this work, we wish to discover these underlying properties automatically, by letting the data indicate its different variable characteristics.

\section{Previous work}
\label{sec:previousWork}

Broadly speaking, there are two main categories of networks which are used for image editing and synthesis: autoencoders  \cite{Kingma2014Auto,ranzato2008sparse,Glorot2011Deep,Makhzani2013K} and GANs \cite{radford2015unsupervised,arjovsky2017wasserstein,gulrajani2017improved,chen2016infogan,odena2017conditional,yan2016attribute2image,mukherjee2019clustergan,delannoy2020segsrgan}. The goal of autoencoders is to compress and decompress data to and from a compact, powerful latent space. GANs, on the other hand, fix the latent space with an a priori distribution (for example Gaussian), and attempt to create realistic data with the parallel action of a generator and an adversarial network. While the models have produced impressive results, understanding and interpreting their latent spaces is now an extremely hot topic. Ideally, we would like to understand what kinds of hidden representations the model has learned. More precisely, the latent space should be disentangled so that one latent code represents one factor of the variation in the formation of the data space.

Many previous works concerning such models have the goal of improving the compactness and power of latent spaces. Firstly, a commonly remarked behaviour of autoencoders is that they fill up all the space allowed in their latent space, which is detrimental to interpretation and manipulation. A common solution to this problem is to allow the autoencoder more space than is likely necessary, and then try to impose some sort of structure on the latent space. Sparse autoencoders \cite{ranzato2008sparse,Glorot2011Deep,Makhzani2013K}, for example, attempt to have as few active (non-zero) neurons or specify a maximum number of non-zero values as possible in the network. However, while this forces compactness, the autoencoder can still entangle several data characteristics in a single latent component. The well-known generative autoencoder such as variational autoencoder \cite{Kingma2014Auto}, Wasserstein autoencoder \cite{tolstikhin2018wasserstein} creates an autoencoder whose latent space is encouraged to follow a certain predefined distribution. While this is very useful for the purposes of synthesis, this does not in itself improve the interpretability of the latent space components, which can mix several characteristics. 

Many previous works exist on the specifc task of disentangling latent representations. Rifai et al \cite{rifai2012disentangling} employ contractive autoencoders to learn locally invariant features at multiple resolutions, which is then given to a ``contractive discriminant analysis'' block for the purpose of emotion prediction. Reed et al \cite{reed2014learning} propose a Boltzmann machine to discover underlying variation in data with two strategies. Firstly, they include the data labels in their cost function for the Boltzmann machine, and secondly, they ``clamp'' (impose) a code for two data points which are known to share some characteristics. The work of Cheung et al \cite{cheung2014discovering}, Kumar et al \cite{kumar2018variational} and Lezama \cite{lezama2019overcoming} are the most similar previous works to ours, in certain aspects. In particular, these works employ some form of covariance loss. Cheung et al use a semi-supervised autoencoder to output an image and at the same time predict a class. Kumar et al propose the covariance loss for the latent space to decorrelate its dimensions, leading to match the moments of the distributions of data and the latent space. Lezama uses a loss on the Jacobian of an autoencoder output with respect to the latent code, to encourage the code to follow the desired class, as well as a prediction loss using binary classes. Lample et al proposed Fader networks \cite{lample2017fader}, which try to isolate a single image characteristic in a single latent component, with an innovative use of a discriminator network. This produces a network where the characteristic can be effectively controlled with a slider. In the case of the work of $\beta$-VAE$_B$ \cite{higgins2017beta}, $\beta$-VAE$_H$ \cite{burgess2018understanding}, FactorVAE \cite{kim2018disentangling} and $\beta$-TCVAE \cite{chen2018isolating}, propose frameworks or regularisation to disentangle VAE by modeling and weighting the Kullback-Leibler divergence term to encourage factorised representations in the latent space.


\section{Principal Component Analysis Autoencoder}
\label{sec:pcaAutoencoder}
Before describing the PCAAE, we first set out some notation. Let $\mathcal{X}$ be the data space, in general, we will consider images of size $m=s \times s$, so $\mathcal{X} = \mathbb{R}^{m}$. We note with $\mathcal{Z}=\mathbb{R}^n$ the latent space, $n$ being the dimensionality of this latent space. We denote the encoder with $E:\mathcal{X} \rightarrow \mathcal{Z}$, and the decoder with $D:\mathcal{Z}\rightarrow \mathcal{X}$. We denote with $z_i$ the $i_{th}$ component of $z$. Let $y = D \circ E(x)$ be the output of the autoencoder. The standard autoencoder loss, also called the \emph{reconstruction loss}, is given by:
\begin{equation}
    \lVert x - D \circ E(x) \rVert^2_2.
\label{eq:reconstructionLoss}
\end{equation}

Now, we describe the core idea and algorithm of PCAAE. As we explained above, there are two central questions we must address in order to define the PCAAE : 1) What do we mean by ``decreasing importance'' of the latent space components, and how can we impose this? 2) How can we enforce independence of the latent components?

In the case of the standard PCA, importance refers to the variability of the data along an axis. Such a definition is difficult to use with an autoencoder since, in general, all the dimensions in the latent space are filled during training. Thus, it is not useful to simply carry out a PCA on the latent space. Therefore, we impose a notion of importance by training a series of autoencoders of increasing latent space size, starting with a latent space of size 1 (a scalar). In this first autoencoder, we can suppose that the information of greatest ``importance'' will be encoded, in the sense of the cost of the $\ell_2$ autoencoder loss. We then increase the size of the latent space by 1, while maintaining the same first component from the previous training: only the second component is trained. This is repeated iteratively until a certain predefined dimension $n$ is attained (as described in Figure \ref{fig:diagram}). Note that at each iteration, the previous decoder is thrown away, and a new one is trained from scratch. Indeed, we wish to impose some structure on the latent space via the training of the encoder, but the decoder must be allowed to do as it sees fit. 

We address the second question, how can we impose independence on the latent codes, in the following manner. \emph{We require that the covariance matrix of the vector $z$ to be as close as possible to the identity matrix}. In order to reduce the computational burden we can, without loss of generality, impose a \emph{batch normalisation} \cite{ioffe2015batch} (BN) layer to the vector $z$, without any learning associated to it. That is to say impose each component of $z$ to be of 0 mean and of variance 1. The magnitude of the off-diagonal entries of the covariance matrix can then be simply expressed as $
\sum_{i\neq j}\left(\mathbb{E}(z_iz_j)\right)^2
$ where $i$ and $j$ range through the dimensions of $z$. We recall that we are adding a new dimension to our latent space while freezing the first dimensions. Therefore, imposing the independence between the components of the vector $z$ boils down to minimizing
\begin{equation}
\label{eqn:lcovexp}    
\sum_{i<k}\left(\mathbb{E}(z_iz_k)\right)^2
\end{equation}
where $k$ is the current dimension being added. This, in turn, can be translated in the loss term, by replacing the expectation by a mean over the whole dataset and keeping only the terms depending on the example $x$ :
\begin{equation}
\label{eqn:Lcov}
\mathcal{L}_{\text{cov}}(z_i(x)z_k(x)) =\sum_{i=1}^{k-1}\left[\frac{1}{N^2}z_k(x)^2z_i(x)^2+\frac{2}{N}z_i(x)z_k(x)\frac{\sum_{x'\neq x}z_i(x')z_k(x')}{N}\right] 
\end{equation}
where $N$ is the size of the dataset.  Since $N^2$ is much larger than $N$, the first term in the brackets can be discarded. This loss can be effectively implemented by replacing the term $\sum_{x'\neq x}z_i(x')z_k(x')/N$ by a running mean, similarly to what is done in the case of a BN. Alternatively, one can simply compute the quantity in \eqref{eqn:lcovexp} over a mini-batch and use it instead of the previous formulation. 
Finally, the loss function of the $i^{th}$ autoencoder is :
\begin{equation}
\mathcal{L}_{AE_i}(x) = \lVert x - D_i \circ E_i(x)\rVert_2^2  \; + \; \sum_{j=1}^{i-1} \lambda_{cov}\mathcal{L}_{\text{cov}}(E_i(x),E_j(x))
\label{Equ:AEi}
\end{equation}

\section{PCAAE for GAN}
\label{sec:PCA_GAN}
The objective of the generator of GANs is to find a mapping from the latent distribution $p_z$ into the image data distribution $p_{data}$. Ideally, we would like each latent component to correspond to one factor of variation in the data. In practice, the latent representations of GANs are entangled (see supplementary material for experimental proof of this). In order to organise and disentangle this latent representation, we apply the PCAE to the latent space of a pre-trained GAN. Indeed, we do not intend to create a new GAN architecture which can compete with state-of-the-art generators such as PGAN, rather we propose to use our PCAAE to better understand and organise the latent space of a high quality, pre-trained GAN. In other words, since the problem of simultaneously learning and organising the latent space is too difficult, we propose to learn first and then organise afterwards. The learning part is done during the training of the high-quality GAN.

Let us highlight that the strategy we propose can be easily adapted to analyse any GAN, and we have chosen PGAN in the current work due to its impressive performances. The input sample to the PGAN lives in $\mathbb{R}^{512}$ (the PGAN latent space), or more precisely is a random sample from the normal gaussian distribution in dimension 512. Since the input is normalised in the first operation of PGAN's generator during testing, we can assume that the latent codes are drawn uniformly from a sphere, which is not convex. To make the job of the autoencoder easier, and since the latent space is not convex, we will apply our tool locally around a given point from the latent space. More precisely, let $\overline{\eta}$ be a fixed point of this sphere (see Figure~\ref{fig:PCA_GAN}). Let $G$ be the generator of PGAN and $\eta$ be a small perturbation vector (drawn randomly). Our goal is to design a low dimensional autoencoder $E,D$ that minimizes the following loss~:
 
 \begin{equation}
      \label{eqn:PCAAEGAN}
	\mathcal{L}(\eta) = \lVert G(\eta+\overline{\eta}) - G (D \circ E(\eta)+\overline{\eta})\rVert_2^2  \; + \; \mathcal{L}_{\text{cov}}(\eta),
\end{equation}

 In other words, the autoencoder's goal is to produce a vector $D(E(\eta))$ which leads to an image that is as close as possible to $G(\eta+\overline{\eta})$. This vector $D(E(\eta))$ will have passed through the low dimensional internal representation of the autoencoder, which is well-organised. The covariance loss $\mathcal{L}_{\text{cov}}$ in Equation~\eqref{eqn:PCAAEGAN} is defined as in \eqref{eqn:Lcov} and will encourage disentanglement of the latent space of the pair $E,D$. We apply the same training strategy that consists in iteratively increasing the number of latent components, while freezing the first components. This training process is illustrated in Figure~\ref{fig:PCA_GAN}. The pair $E,D$ is, a priori, specific to a fixed $\overline{\eta}$ although some pairs $E,D$ have been found to work around other points (see supplementary material).

\begin{figure}
    \centering

    \begin{minipage}{.49\textwidth}
    \begin{tabularx}{\linewidth}{ccc}
    \includegraphics[width=0.31\linewidth, trim={0.5cm 0.8cm 3.0cm 3.1cm}, clip]{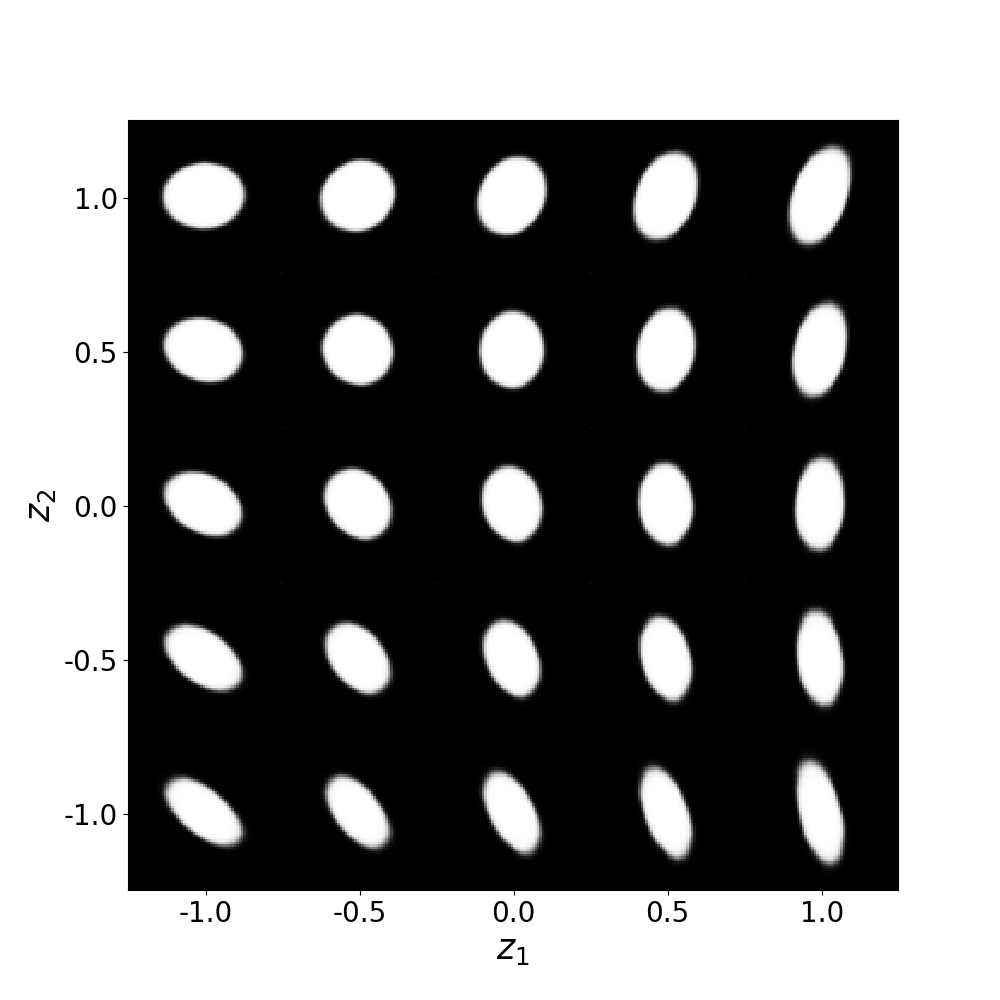}
    \includegraphics[width=0.31\linewidth, trim={0.5cm 0.8cm 3.0cm 3.1cm}, clip]{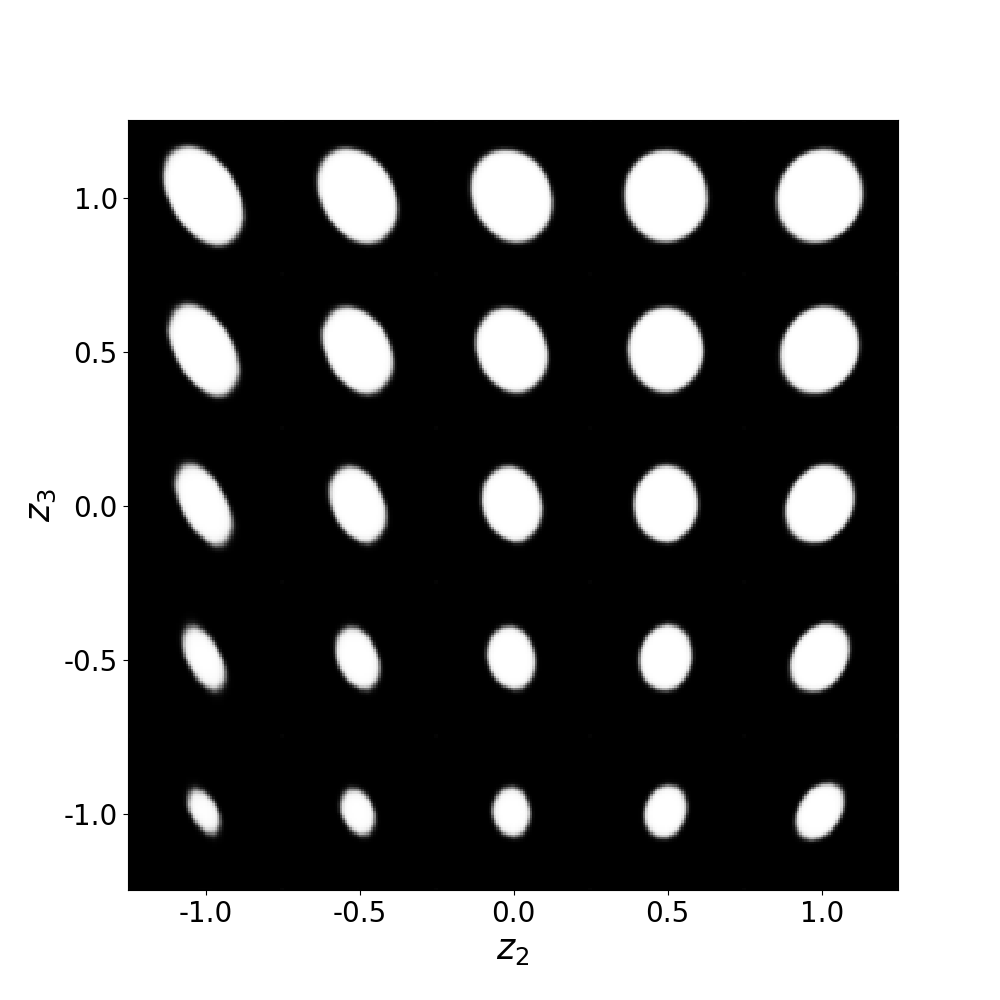}
    \includegraphics[width=0.31\linewidth, trim={0.5cm 0.8cm 3.0cm 3.1cm}, clip]{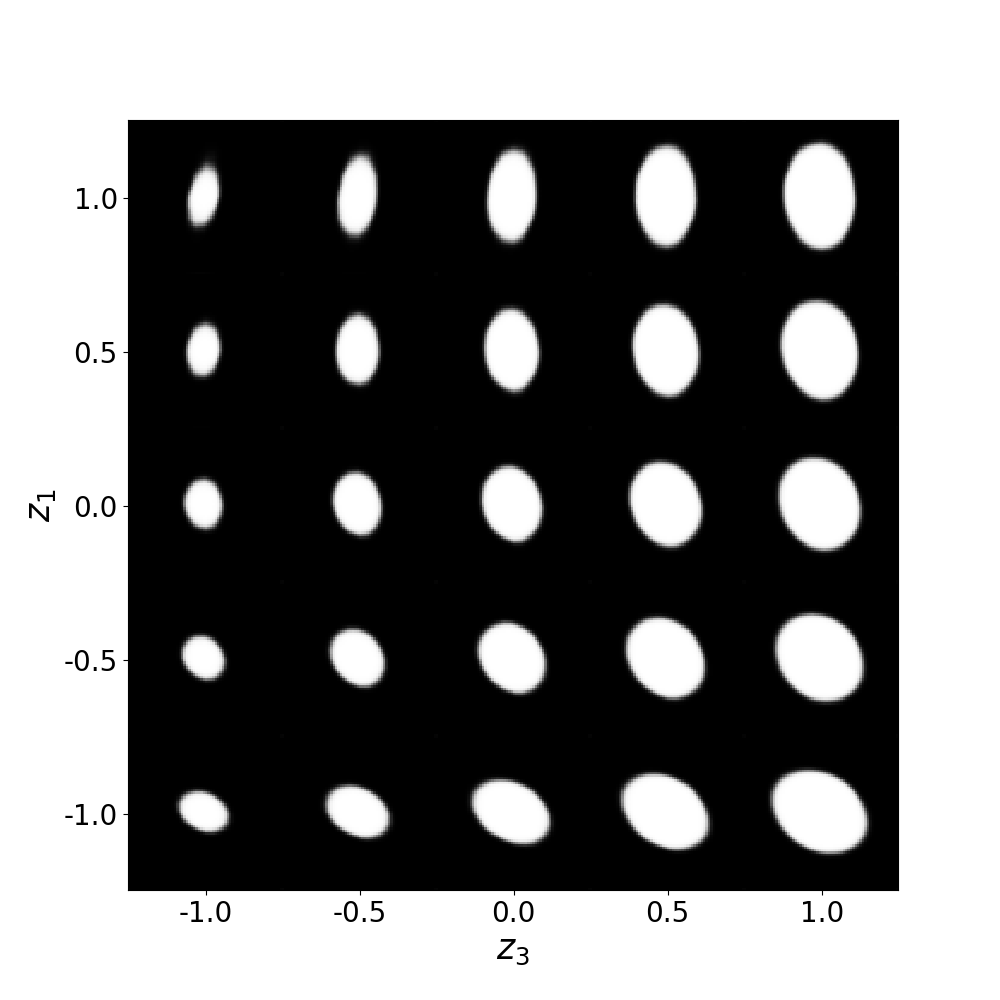}
    \end{tabularx}
    \vspace{-7pt}
    \subcaption{\normalsize VAE}
    \end{minipage}
    \hfill
    \begin{minipage}{.49\textwidth}
    \begin{tabularx}{\linewidth}{ccc}
    \includegraphics[width=0.31\linewidth, trim={0.5cm 0.8cm 3.0cm 3.1cm}, clip]{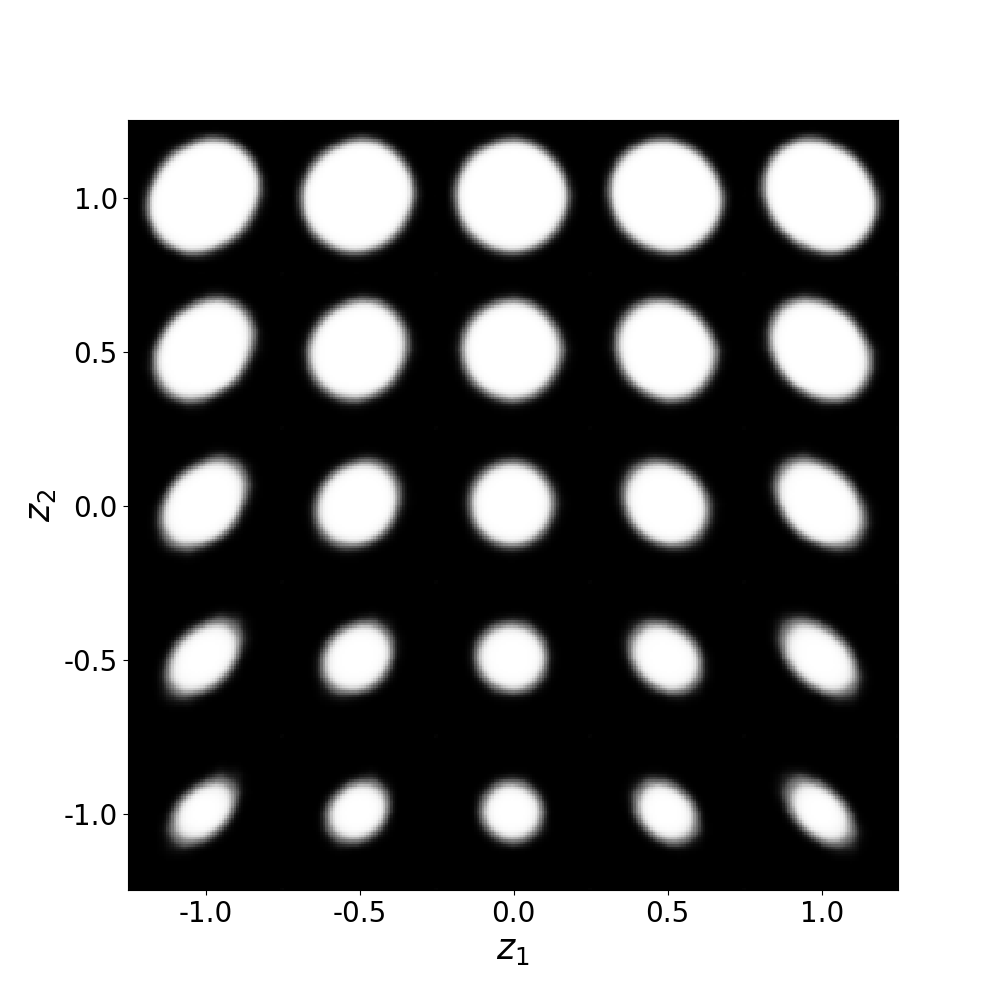}
    \includegraphics[width=0.31\linewidth, trim={0.5cm 0.8cm 3.0cm 3.1cm}, clip]{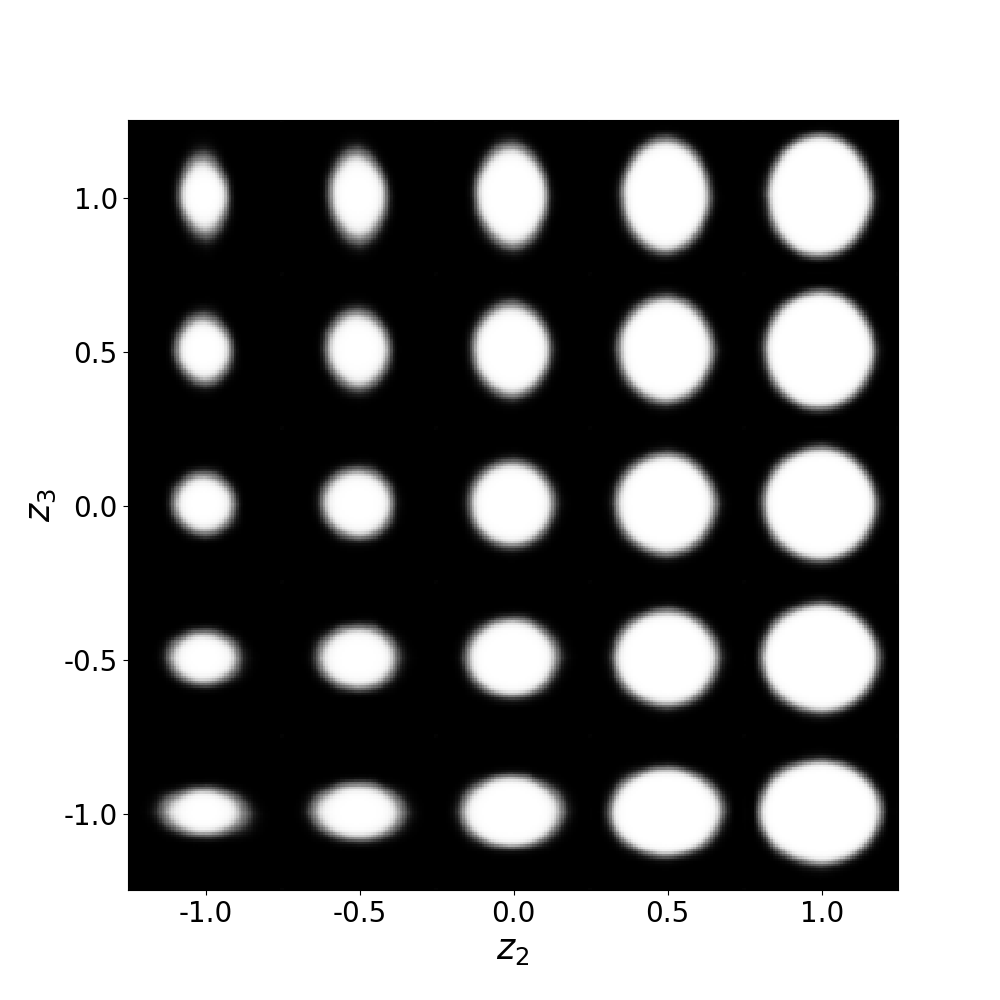}
    
    \includegraphics[width=0.31\linewidth, trim={0.5cm 0.8cm 3.0cm 3.1cm}, clip]{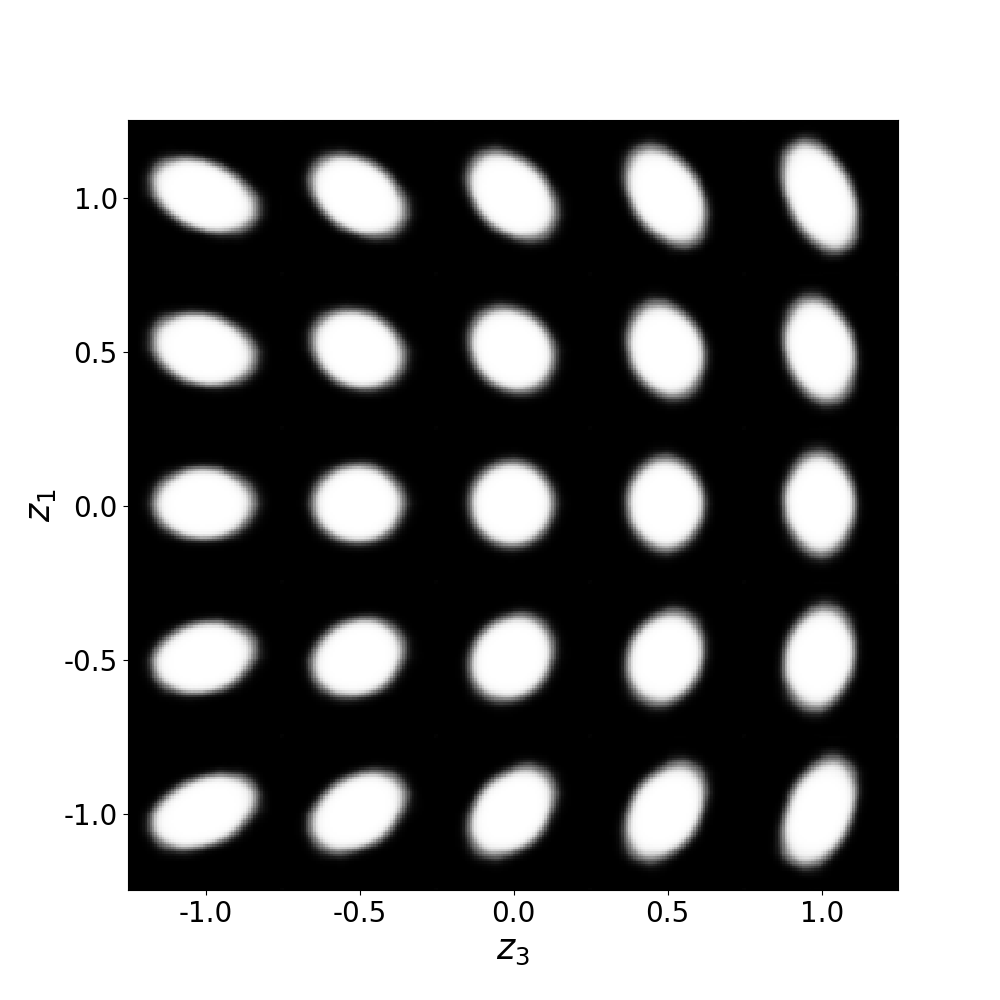}
    \end{tabularx}
    \vspace{-7pt}
    \subcaption{\normalsize $\beta$-VAE}
    \end{minipage}
    \\
    \begin{minipage}{.49\textwidth}
    \begin{tabularx}{\linewidth}{ccc}
    \includegraphics[width=0.31\linewidth, trim={0.5cm 0.8cm 3.0cm 3.1cm}, clip]{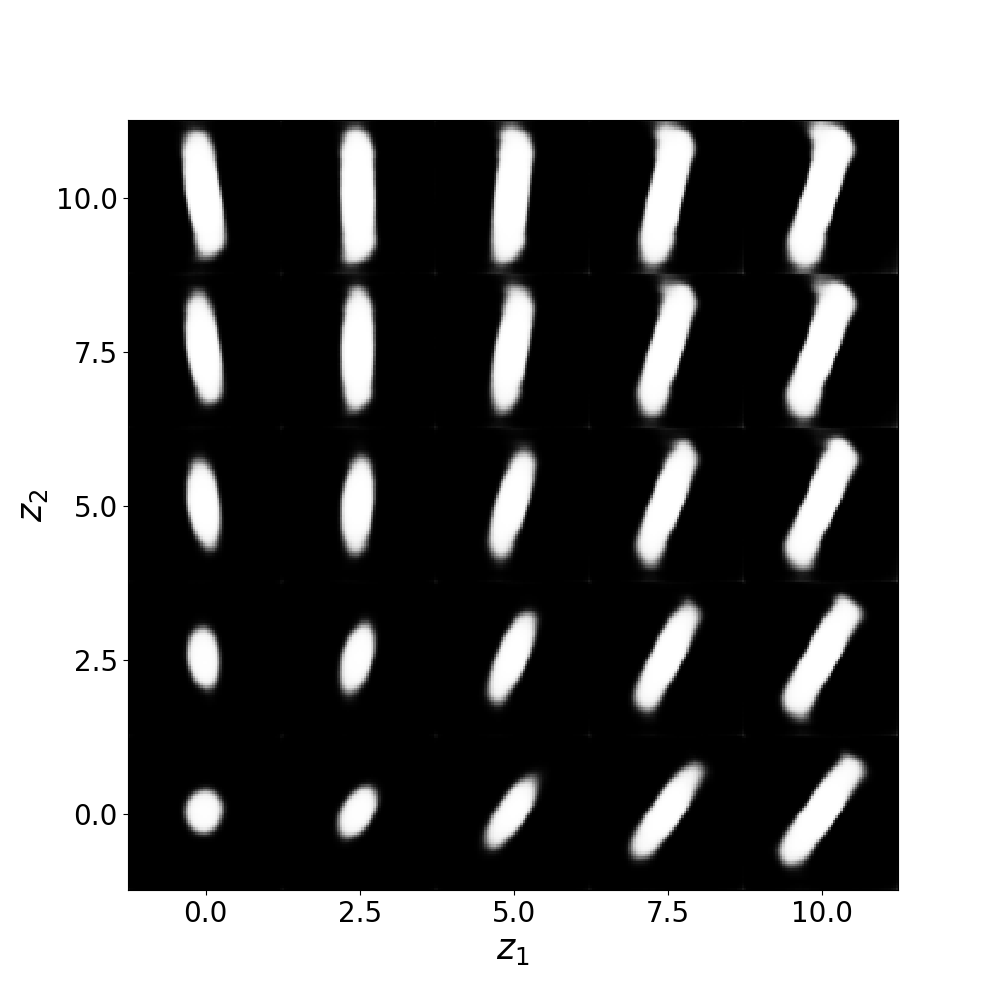}
    
    \includegraphics[width=0.31\linewidth, trim={0.5cm 0.8cm 3.0cm 3.1cm}, clip]{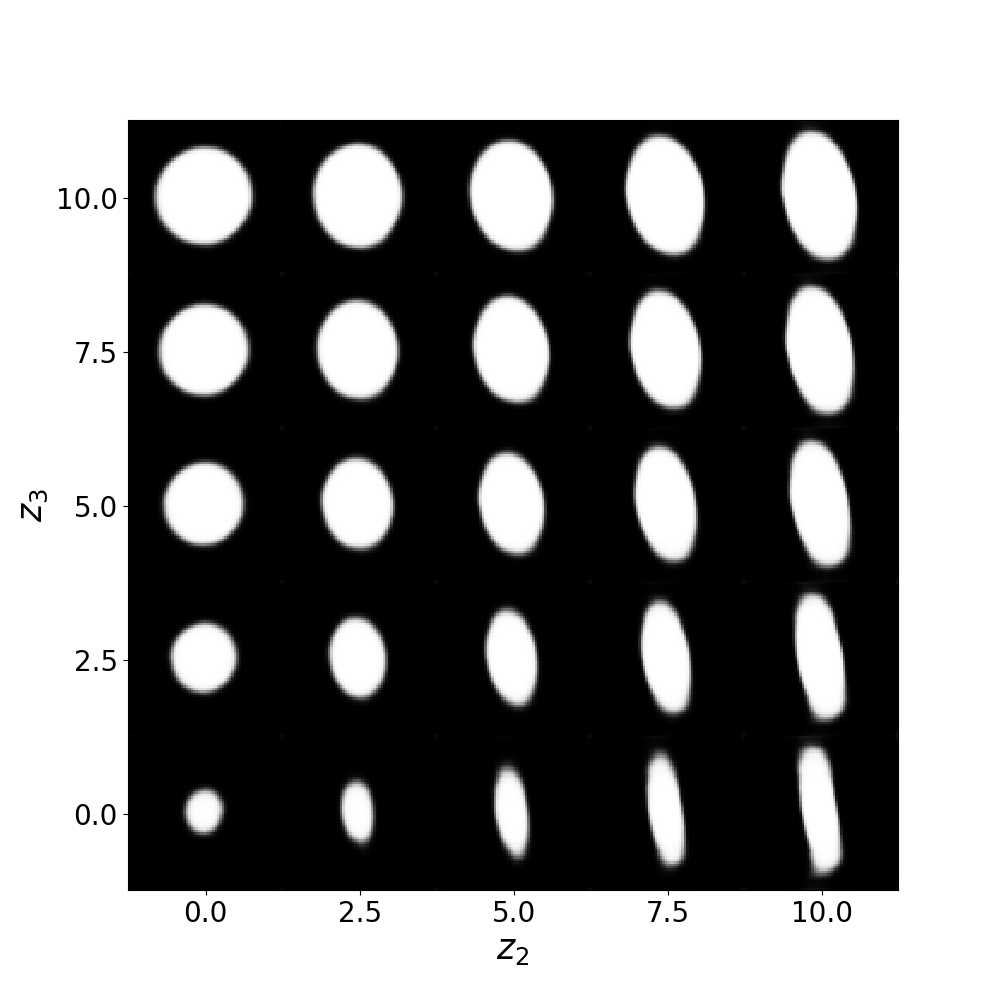}
    
    \includegraphics[width=0.31\linewidth, trim={0.5cm 0.8cm 3.0cm 3.1cm}, clip]{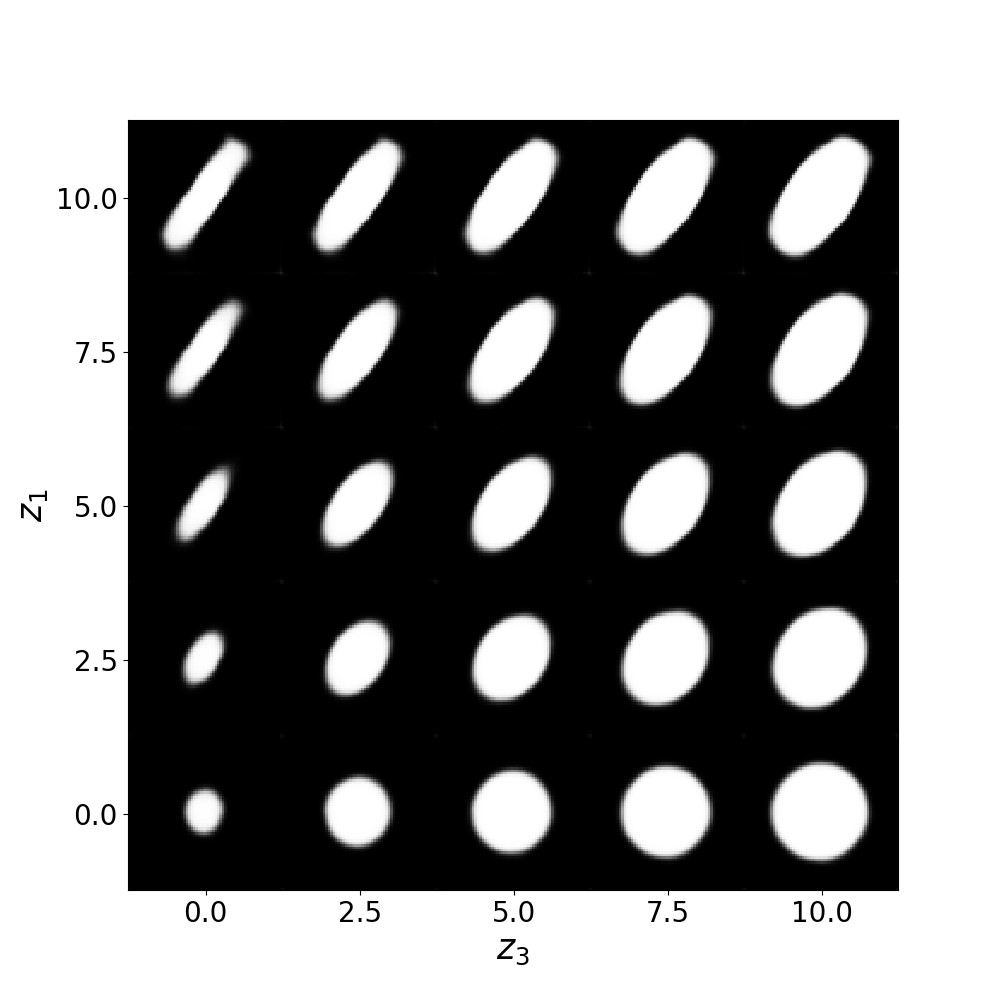}
    \end{tabularx}
    \vspace{-7pt}
    \subcaption{\normalsize $\beta$-TCVAE}
    \end{minipage}
    \hfill
    \begin{minipage}{.49\textwidth}
    \begin{tabularx}{\linewidth}{ccc}
    \includegraphics[width=0.31\linewidth, trim={0.5cm 0.8cm 3.0cm 3.1cm}, clip]{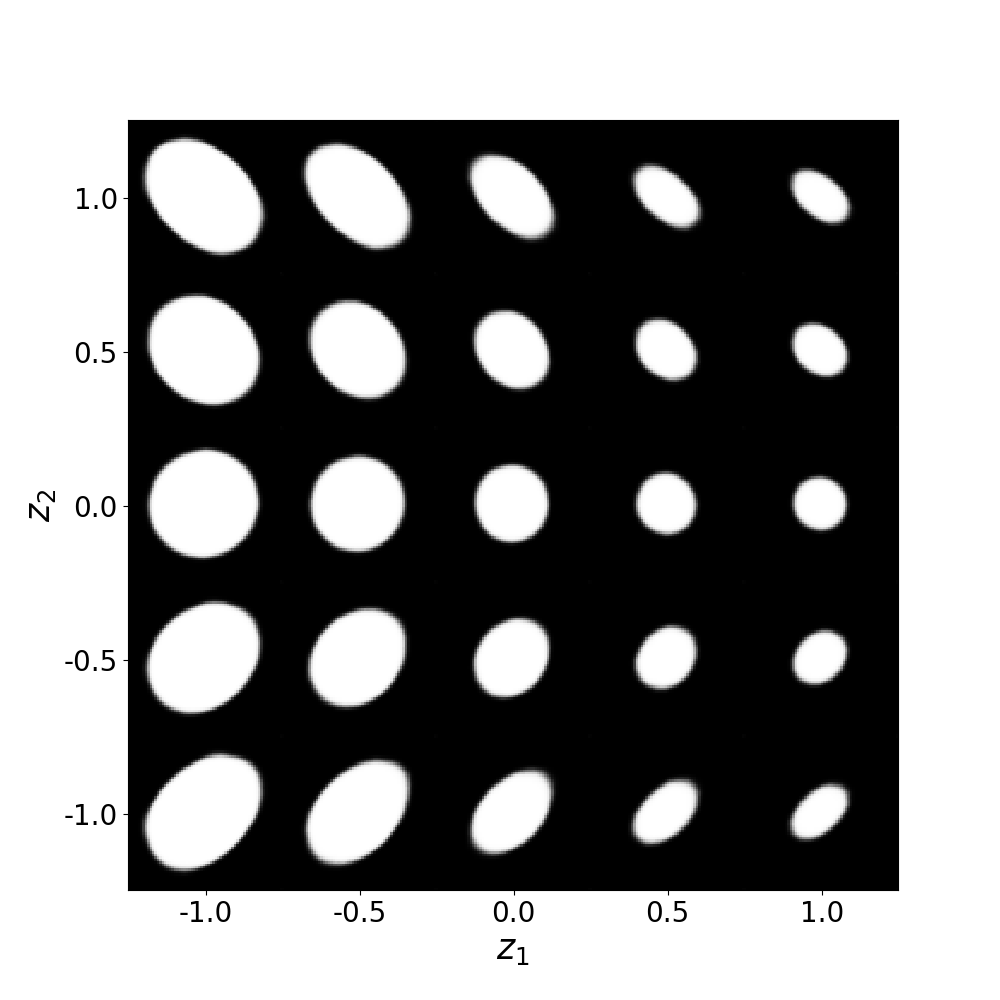}
    \includegraphics[width=0.31\linewidth, trim={0.5cm 0.8cm 3.0cm 3.1cm}, clip]{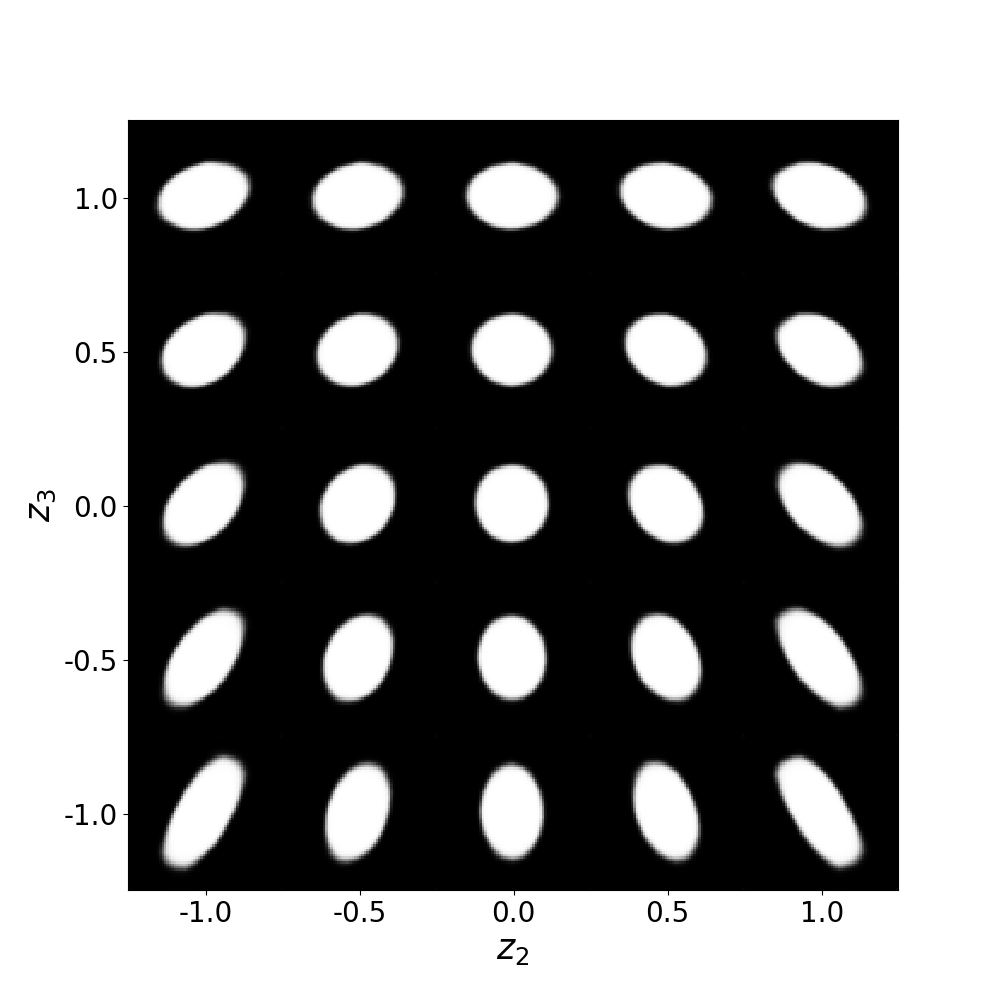}
    \includegraphics[width=0.31\linewidth, trim={0.5cm 0.8cm 3.0cm 3.1cm}, clip]{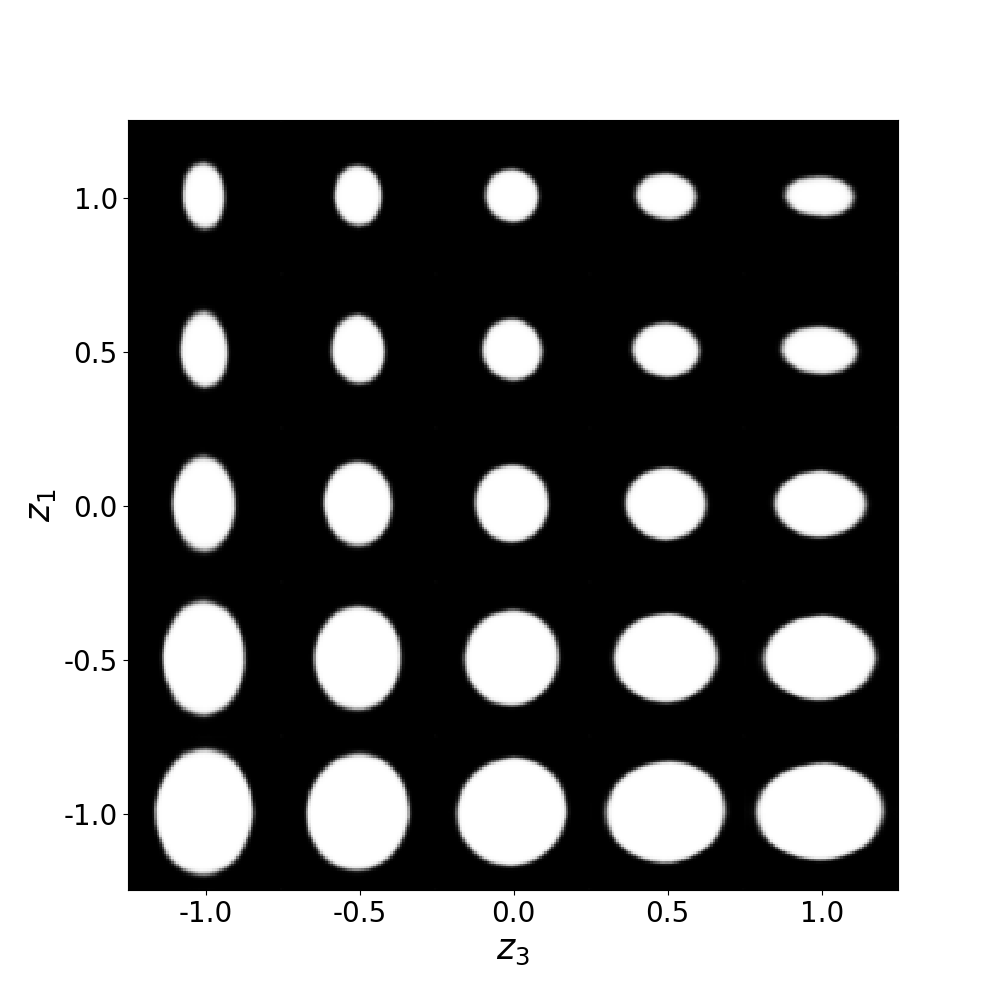}
    \end{tabularx}
    \vspace{-7pt}
    \subcaption{\normalsize PCAAE}
    \end{minipage}
    \vspace{-7pt}
    \caption{Interpolation in latent space w.r.t image reconstruction, ellipses with rotation (three parameters) of VAE, $\beta$-VAE, $\beta$-TCVAE and our proposed method. The PCAAE can create a meaningful latent space where different geometric attributes are separated (\textit{i.e.} the $1^{st}$ component corresponds to the surface and the next two parameters are the ratios of the ellipses' axes in different directions). (Please see supplementary material for higher quality images).
    }
    
    \label{fig:3D_ellipses_rotation}
\end{figure}{}

\section{Results}
\label{sec:results}

In this section, we present the results of our PCAAE, and we compare with those of VAE \cite{kingma2014adam}, $\beta$-VAE$_B$ \cite{higgins2017beta}, $\beta$-VAE$_H$ \cite{burgess2018understanding}, FactorVAE \cite{kim2018disentangling} and $\beta$-TCVAE \cite{chen2018isolating}. Note that other approaches to disentangling the latent space use data labels, which we wish to avoid here : our goal is to discover the variability of the data in an unsupervised fashion.

\subsection{Disentanglement evaluation}

\begin{table}[t]
\centering
\begin{tabular}{|c|c|c|c|}
\hline
\multirow{2}{*}{}  & \multicolumn{3}{c|}{PCAAE} \\
\cline{2-4}
& A & R1 & R2 \\
\hline
 $Z_1$ & \textbf{0.97} & 0.15 & 0.16 \\
 $Z_2$ & 0.00 & 0.06 & \textbf{0.61}  \\
 $Z_3$ & 0.00 & \textbf{0.72} & 0.06 \\
 \hline
\end{tabular}{}
 \hspace{1pt}
\begin{tabular}{|c|c|c|c|c|c|c|}
\hline  
\multirow{2}{*}{}  &  \multicolumn{6}{c|}{Area of ellipses} \\
\cline{2-7}
& \small  AE & \small  VAE &  \small $\beta$-VAE$_B$ & \small $\beta$-TCVAE & \small FactorVAE & \small PCAAE\\

\hline        
 $Z_1$ &  0.51 & 0.00 &  0.15 &  0.10 & 0.01 & \textbf{0.97} \\
 $Z_2$ &  0.00 & 0.31 &  0.90 &  0.14 &  0.07 & 0.03 \\
 $Z_3$ &  0.64 & 0.83 &  0.06 &  0.86 & 0.89 & 0.00 \\
\hline 
\end{tabular}{}
\caption{Evaluation of the absolute PCC between the attributes of ellipses w.r.t. three components ($Z_1$, $Z_2$ and $Z_3$) of the trained latent space. We consider three attributes: the area (A), the ratio of two diameters towards vertical and horizontal directions (R1), the ratio of two diameters towards diagonal directions (R2). In the left table, bold font denotes the largest value among the components. In the right table, the strongest correlation (the PCAAE's) is in bold font. We can see that each component of PCAAE is strongly correlated with only one ellipse attribute.
}
\label{table:evaluation_ellipses}
\vspace{-10pt}
\end{table}

We propose to use the absolute Pearson correlation coefficient (PCC) as a disentanglement evaluation to verify the relationship between the attributes of image data and the components of the trained latent space. Given a pair of random variables $(Attr(X),z_i)$ where $Attr(X)$ is the attribute of image $X$ and $z_i$ denotes the $i^{th}$ component of the latent space that represents the data, the absolute PCC  $\rho(Attr(X),z_i) $ is computed as:
\begin{equation}
    \rho(Attr(X),z_i) = \left| \frac{cov(Attr(X),z_i)}{\sigma_{Attr(X)} \sigma_{z_i}} \right|= \left| \frac{\mathbb{E}[(Attr(X)-\mu_{Attr(X)})(z_i-\mu_{z_i})]}{\sigma_{Attr(X)} \sigma_{z_i}} \right|
    \label{PCC}
\end{equation}{}
where $\sigma_{Attr(X)}$ and $\sigma_{z_i}$ denote the standard deviation of ${Attr(X)}$ and $z_i$, respectively. $\mu_{Attr(X)}$ and $\mu_{z_i}$ is the mean of ${Attr(X)}$ and $z_i$, respectively. The absolute PCC ranges from 0 to 1.
One dimension of the disentangled latent space, which corresponds to a attribute of data, show the much larger value than others.

\subsection{Experimental setup and results on synthetic data}
\label{subsec:architecture_dataset}

In order to find out whether our PCAAE is able to capture meaningful components which correspond to the parameters of visual objects, we have first tested our algorithm on synthetic data of binary images of geometric shapes which are centred in the image, with a single shape per image. We have created images of ellipses in the case of three parameters (two axes, and rotation). The two ellipse axes are sampled from a uniform distribution on the interval $(0,\frac{n}{2})$, and the rotation from a uniform distribution on the interval $(0,\frac{\pi}{2})$. In these experiments, we set $n$ (maximum autoencoder dimension) to 3 (the number of parameters used to create the dataset). A drawback of using data with binary images of shapes is that we have a limited number of centred parametric shapes that we can create, even though we sample the parameters from a continuous space. To solve this problem, we blur the binary shapes slightly with a Gaussian filter with $\sigma=0.8$ pixels, allowing us to create as many images as we wish.

Figure~\ref{fig:3D_ellipses_rotation} shows decoded images of interpolated points in the latent space, in the case of ellipses. Table~\ref{table:evaluation_ellipses} shows the numeric evaluation based on the absolute PCC between the attributes of ellipses with respect to three parameter of the trained latent space. We observe that the latent space of our PCAAE corresponds to three principal attributes of ellipses : area (A), the ratio of two diameters towards vertical and horizontal directions (R1), the ratio of two diameters towards diagonal directions (R2). The compared methods also create a meaningful latent space whereas AE and VAE learn a latent space where the intrinsic parameters of the ellipses are mixed up. While these are not the parameters with which we created the images (indeed, the autoencoder has absolutely no way of knowing what representation to choose, and we cannot impose one), they are indeed independent; for a given surface, the ratio between the axes is an independent parameter, and vice versa. This gives us a way to interpolate in the latent space in a meaningful manner. These independent parameters are sufficient to describe the ellipse, and each axis is hierarchically more interpretable and navigable than in the case of other methods. For more results, see the supplementary material.

\begin{table}[t]
\begin{tabular}{|c|c|c|c|c|c|c|c|c|c|c|c|c|}
\hline  
\multirow{2}{*}{Co.}  &  \multicolumn{3}{|c|}{AE} &  \multicolumn{3}{|c|}{$\beta$-TCVAE} &  \multicolumn{3}{|c|}{FactorVAE} & \multicolumn{3}{|c|}{PCAAE}\\
\cline{2-13}
& HC & HP & GE & HC & HP & GE & HC & HP & GE & HC & HP & GE\\
\hline        
$Z_1$  & 0.20          & \textbf{0.53} & 0.02          & 0.35          & \textbf{0.80} & 0.04          & 0.07          & 0.46          & \textbf{0.53} & \textbf{0.70} & 0.03          & 0.14          \\
$Z_2$  & \textbf{0.36} & 0.21          & 0.04          & 0.05          & 0.26          & 0.25          & 0.04          & 0.17          & 0.03          & 0.07          & \textbf{0.80} & 0.13          \\
$Z_3$  & 0.28          & 0.36          & 0.23          & 0.13          & 0.04          & 0.04          & 0.09          & \textbf{0.66} & 0.37          & 0.16          & 0.15          & \textbf{0.56} \\
$Z_4$  & 0.20          & 0.19          & \textbf{0.58} & \textbf{0.53} & 0.19          & \textbf{0.53} & \textbf{0.70} & 0.11          & 0.14          & 0.03          & 0.24          & 0.05          \\
$Z_5$  & 0.34          & 0.21          & 0.49          & 0.07          & 0.15          & 0.33          & 0.01          & 0.28          & 0.12          & 0.05          & 0.22          & 0.09          \\
\hline
\end{tabular}{}
\caption{Quantitative evaluation of the correlation of latent components with high-level attributes. We have calculated the PCC between the latent components of AE, VAE-based methods and PCAAE, and three attributes: head pose (HP), hair colour (HC) and gender (GE). We can see that the components of PCAAE are correlated with one attribute only. (See supplementary material for more results)
}
\vspace{-10pt}
\label{tab:table_evaluation_gan}
\end{table}

\subsection{Experimental setup and results of the PCAAE applied to the latent space of PGAN}

To show the use of our PCAAE on more high-level data, we take a pre-trained model of PGAN \cite{karras2017progressive} \footnote{Pytorch GAN zoo: \url{https://github.com/facebookresearch/pytorch_GAN_zoo}} trained with the CelebA dataset \cite{liu2015faceattributes}. Note that the pre-trained generator is fixed during the training of our PCAAE. The latent space of PGAN is entangled (we show experiments to support this in the supplementary material), so that a variation along one parameter of this initial code in the latent space can modify several characteristics of the generated images. The latent space size of this pre-trained network is 512. An initial code $\overline{\eta}$ , from which the network generates a photo-realistic image, is chosen. In order to create the set of random perturbations, we sample from a Gaussian distribution $\eta \sim \mathcal{N}(\mu,\sigma^{2})$.

In order to evaluate the disentanglement of the latent space of other methods and ours, we use pre-trained classifiers to determine an attribute of generated images. We choose three main attributes which the classifiers \cite{Face} can recognize well, corresponding to the head pose (\textit{i.e.} turning left to right), hair colour (\textit{i.e.} black, brunette and blond) and gender. Note that in the supplementary material, we also display the results of our PCAAE directly to the Celeba data. This gives very blurry results (since the task is very difficult, as mentioned above), similar to the results of $\beta$-VAE \cite{higgins2017beta} (Figure 4 of their paper), which lead us to our approach to using the PCAAE applied to GANs.

\begin{figure}[t]
    \centering

    \begin{minipage}{\textwidth}
    \begin{tabularx}{\linewidth}{ccc}
    \includegraphics[width=0.24\linewidth, trim={1.7cm 1.0cm 3.0cm 3.1cm}, clip]{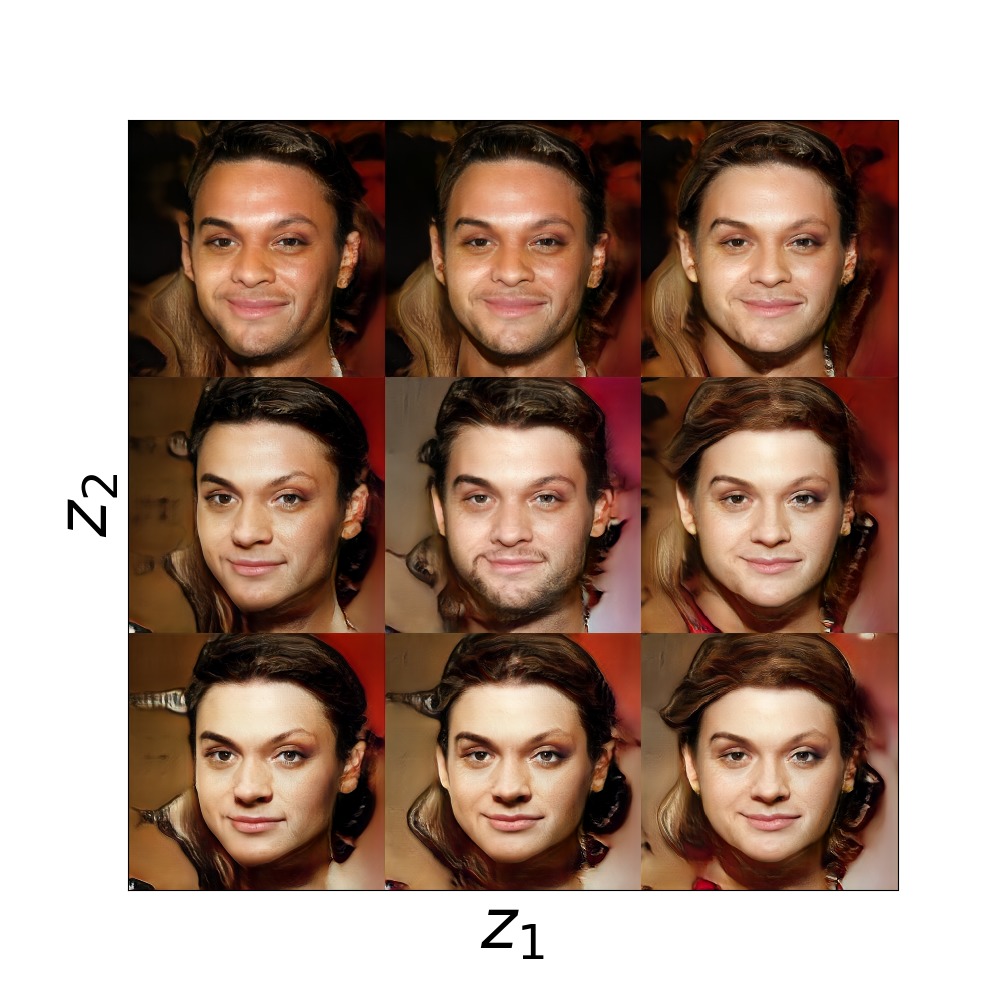}
    \includegraphics[width=0.24\linewidth, trim={1.7cm 1.0cm 3.0cm 3.1cm}, clip]{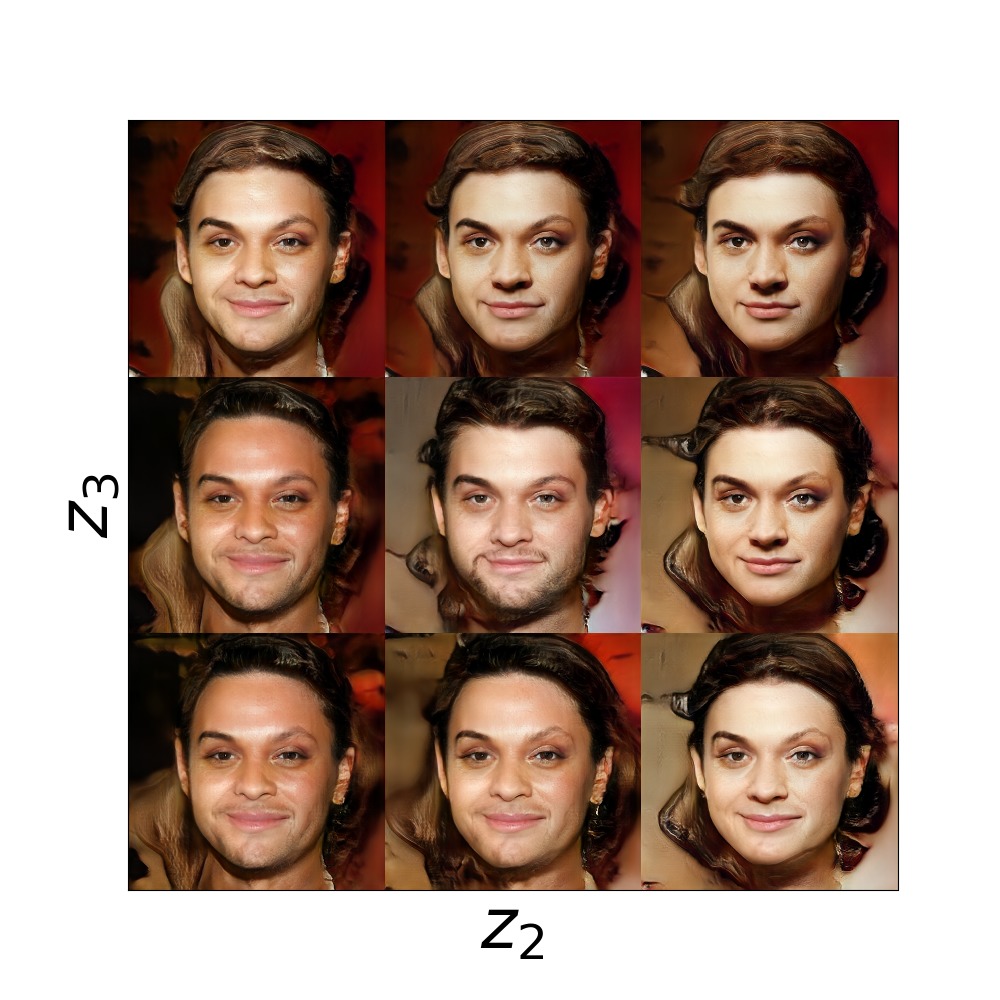}
    \includegraphics[width=0.24\linewidth, trim={1.7cm 1.0cm 3.0cm 3.1cm}, clip]{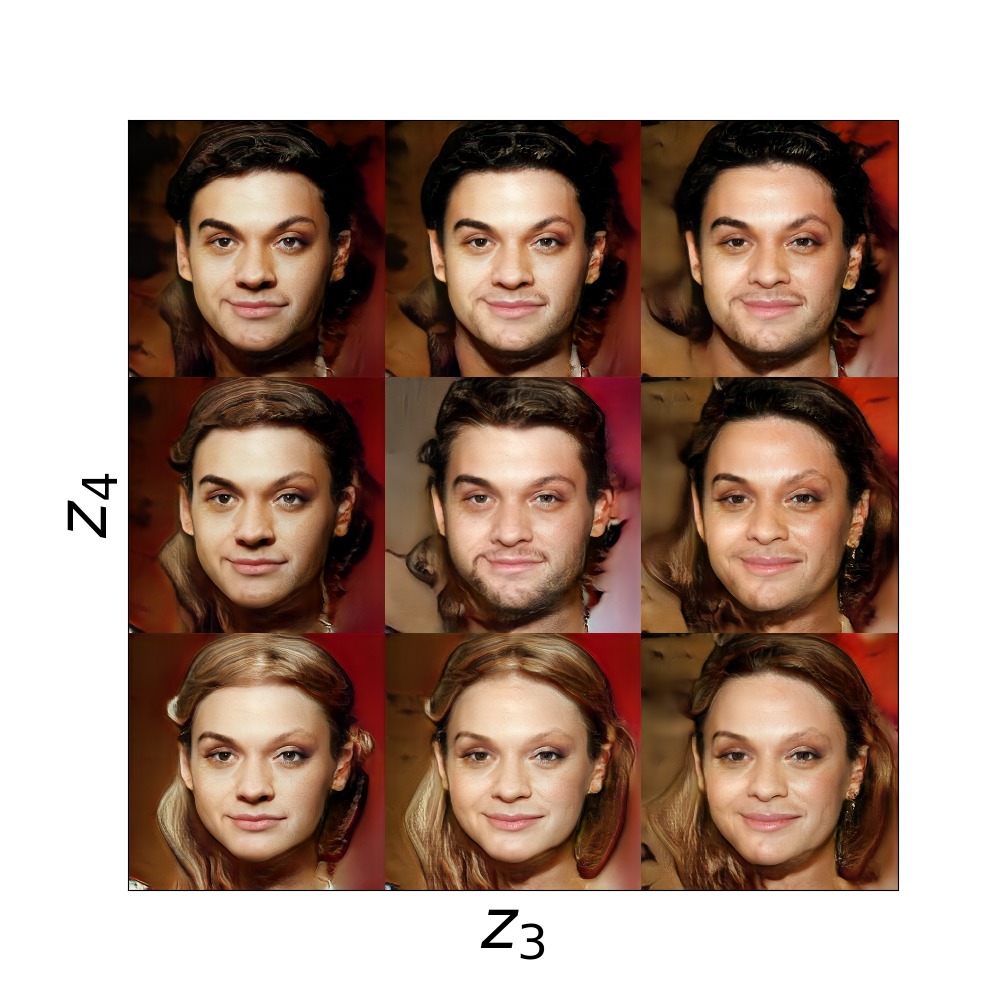}
    \includegraphics[width=0.24\linewidth, trim={1.7cm 1.0cm 3.0cm 3.1cm}, clip]{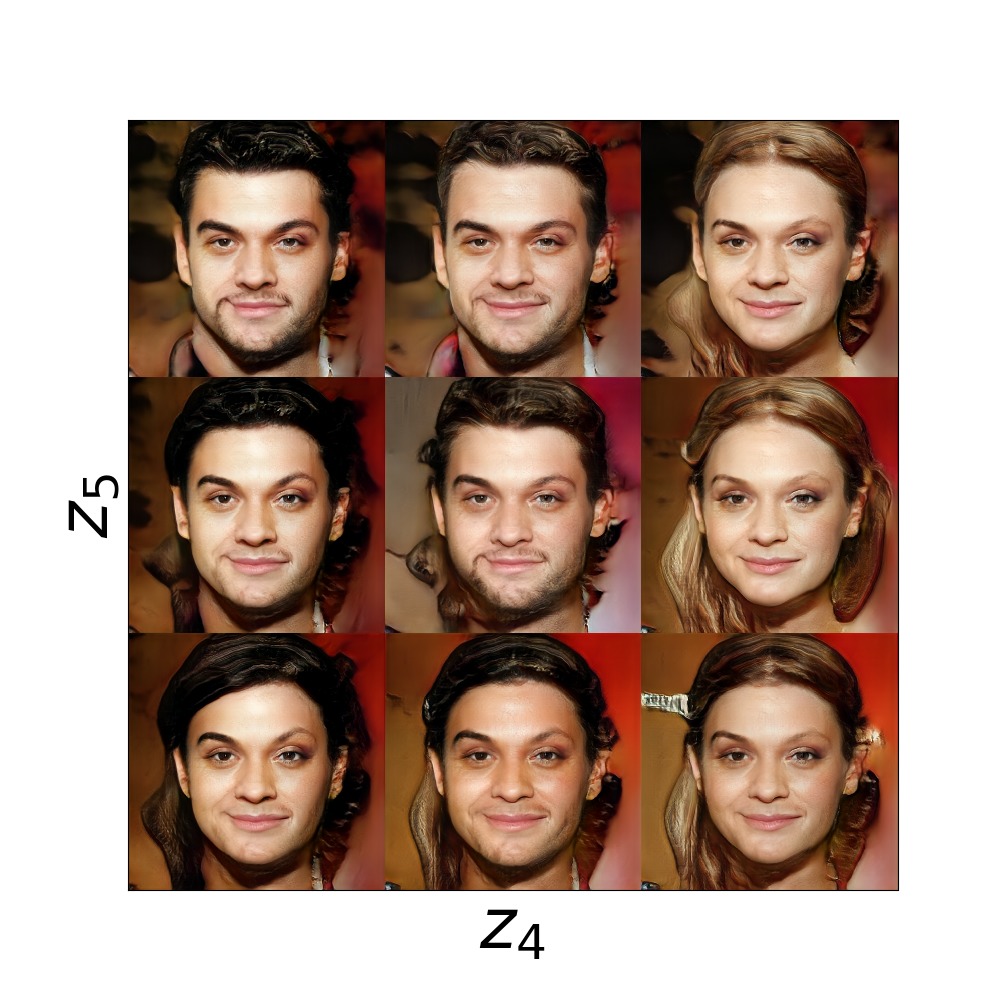}
    \end{tabularx}
    \vspace{-9pt}
    \subcaption{\normalsize $\beta$-TCVAE}
    \end{minipage}
      \\ 
    \begin{minipage}{\textwidth}
    \begin{tabularx}{\linewidth}{ccc}
    \includegraphics[width=0.24\linewidth, trim={1.7cm 1.0cm 3.0cm 3.1cm}, clip]{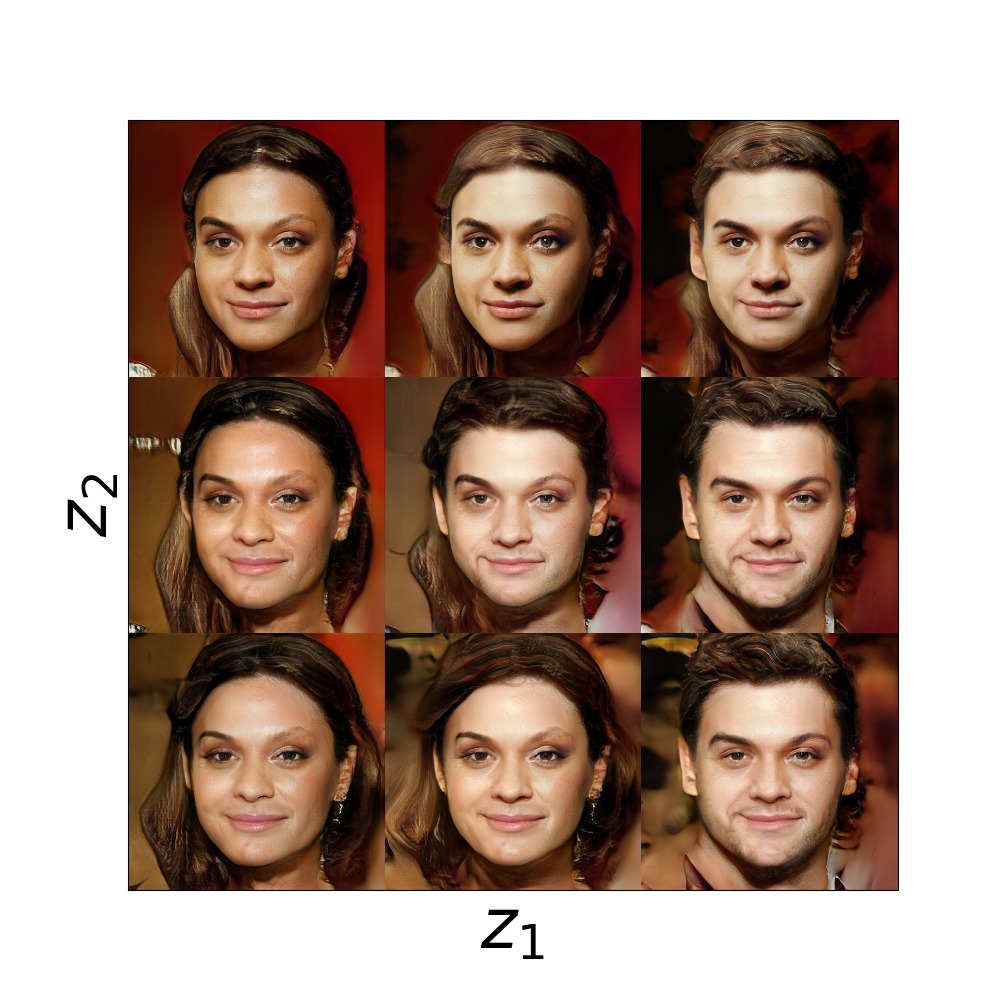}
    \includegraphics[width=0.24\linewidth, trim={1.7cm 1.0cm 3.0cm 3.1cm}, clip]{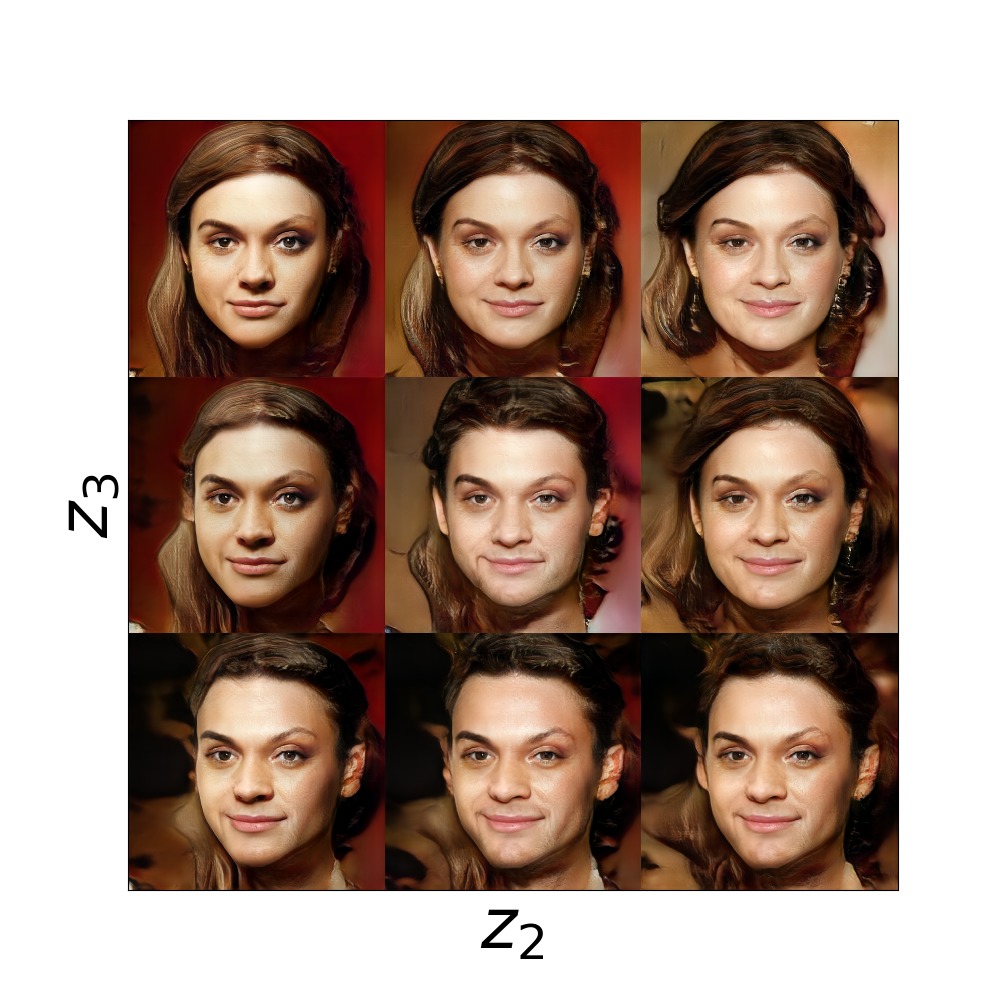}
    \includegraphics[width=0.24\linewidth, trim={1.7cm 1.0cm 3.0cm 3.1cm}, clip]{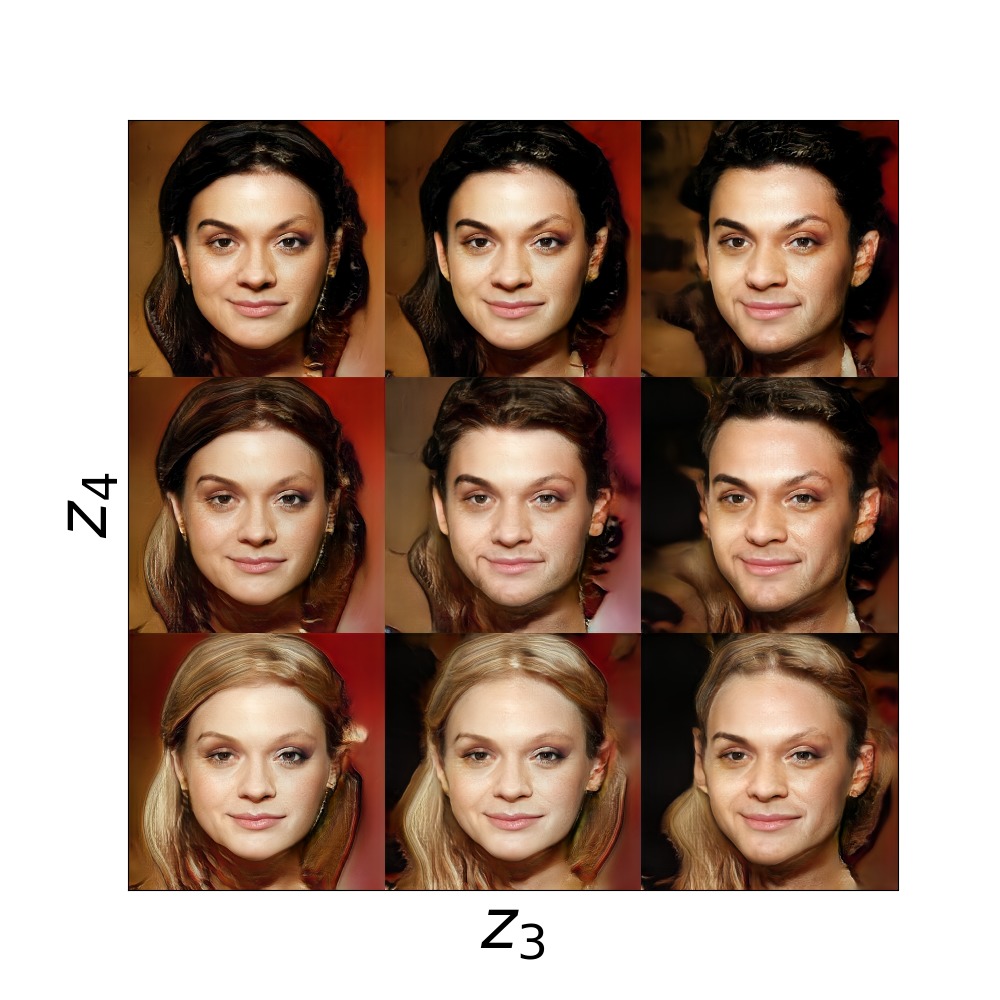}
    \includegraphics[width=0.24\linewidth, trim={1.7cm 1.0cm 3.0cm 3.1cm}, clip]{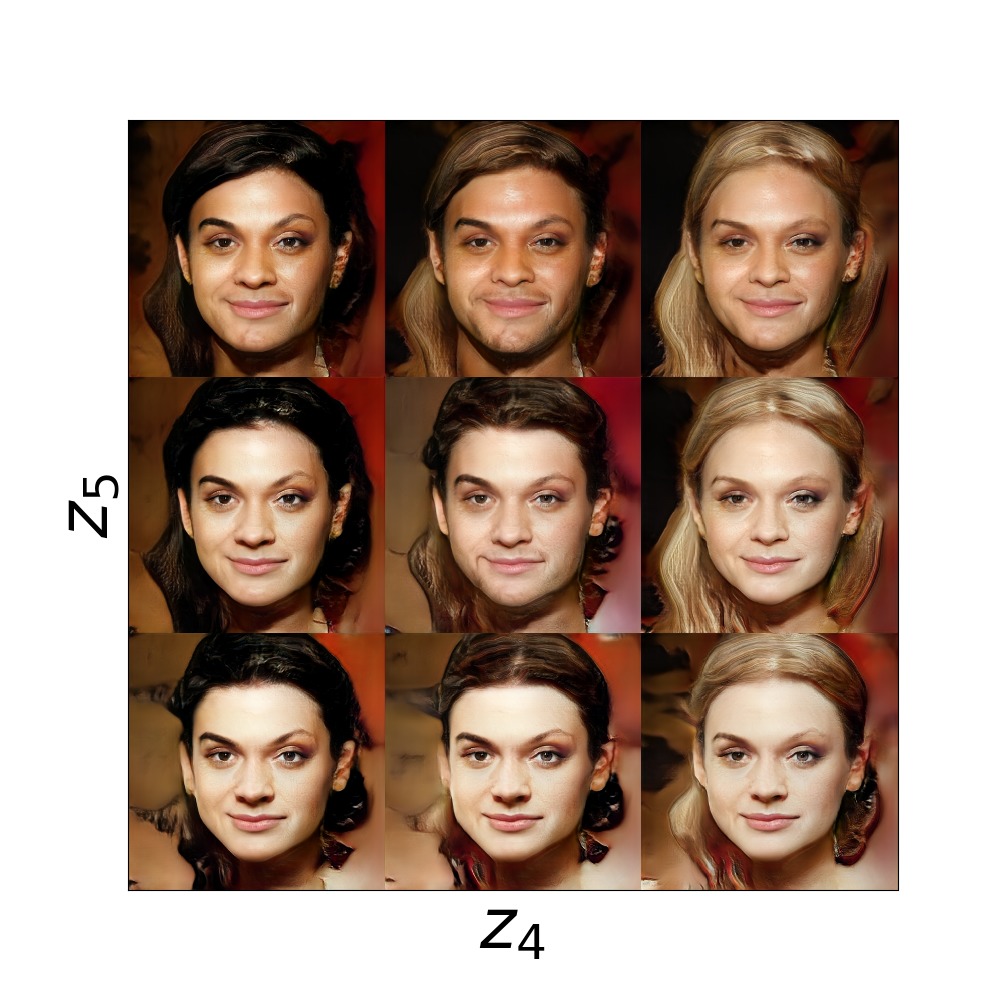}
    \end{tabularx}
    \vspace{-9pt}
    \subcaption{\normalsize FactorVAE}
    \end{minipage}
    \\ 
    \begin{minipage}{\textwidth}
    \begin{tabularx}{\linewidth}{ccc}
    \includegraphics[width=0.24\linewidth, trim={1.7cm 1.0cm 3.0cm 3.1cm}, clip]{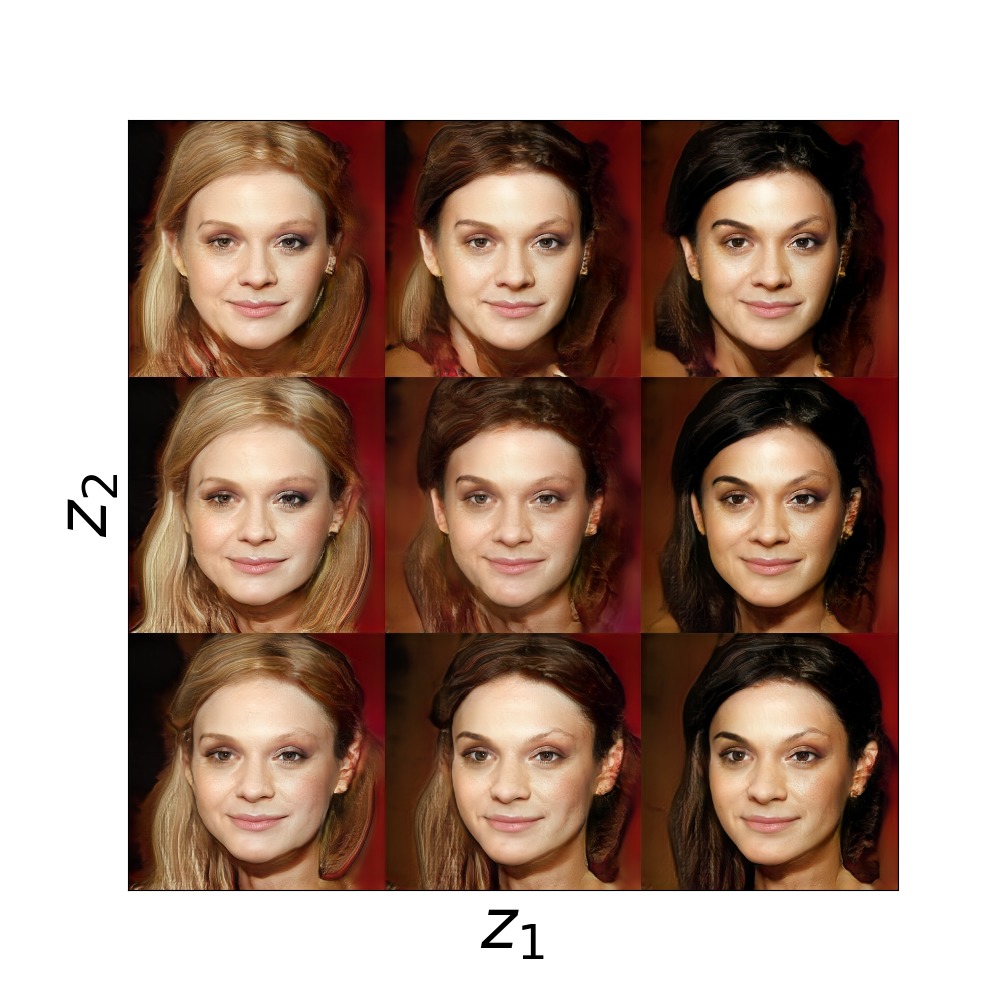}
    \includegraphics[width=0.24\linewidth, trim={1.7cm 1.0cm 3.0cm 3.1cm}, clip]{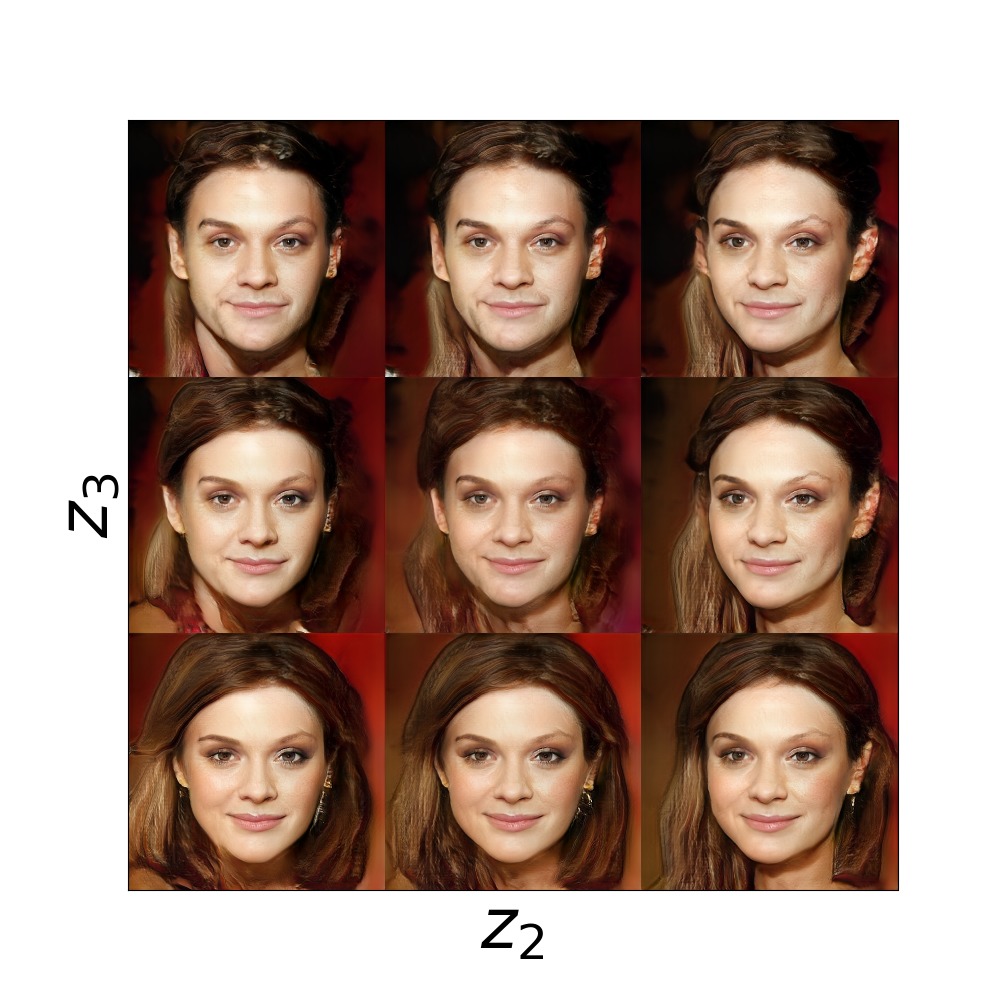}
    \includegraphics[width=0.24\linewidth, trim={1.7cm 1.0cm 3.0cm 3.1cm}, clip]{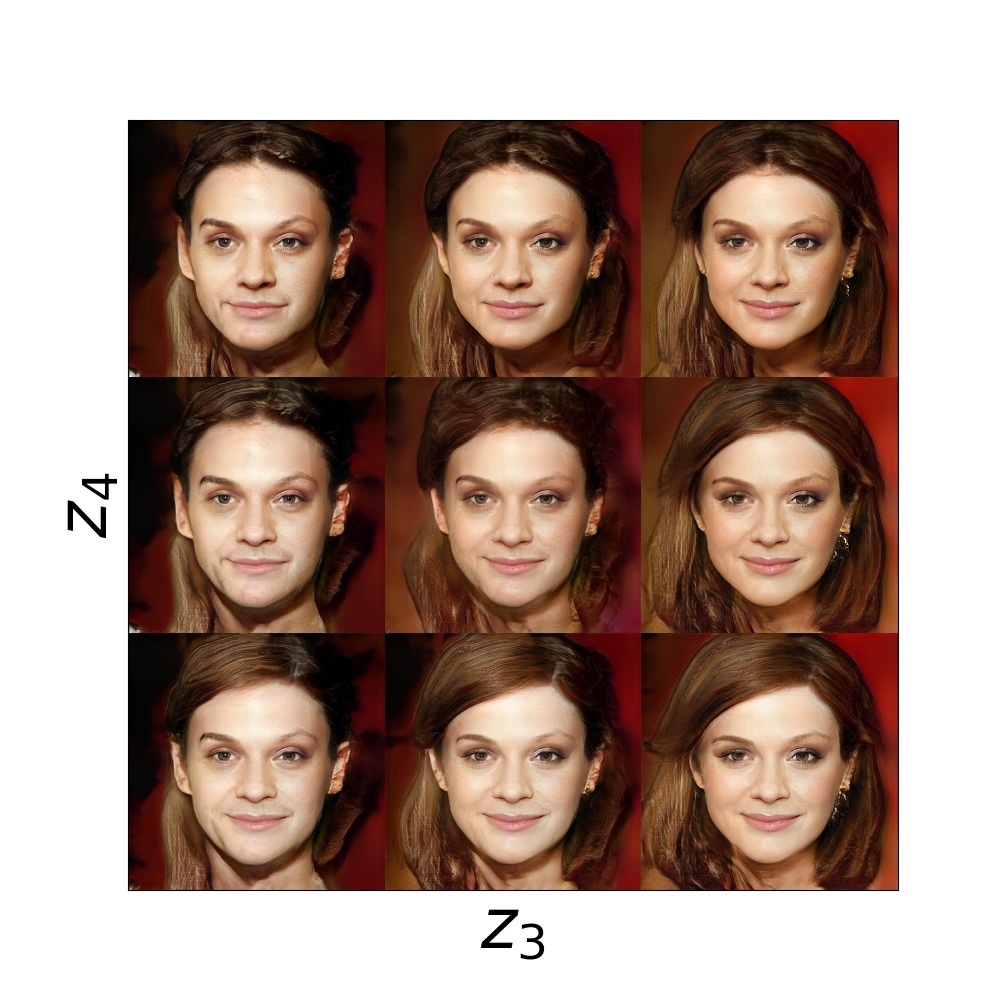}
    \includegraphics[width=0.24\linewidth, trim={1.7cm 1.0cm 3.0cm 3.1cm}, clip]{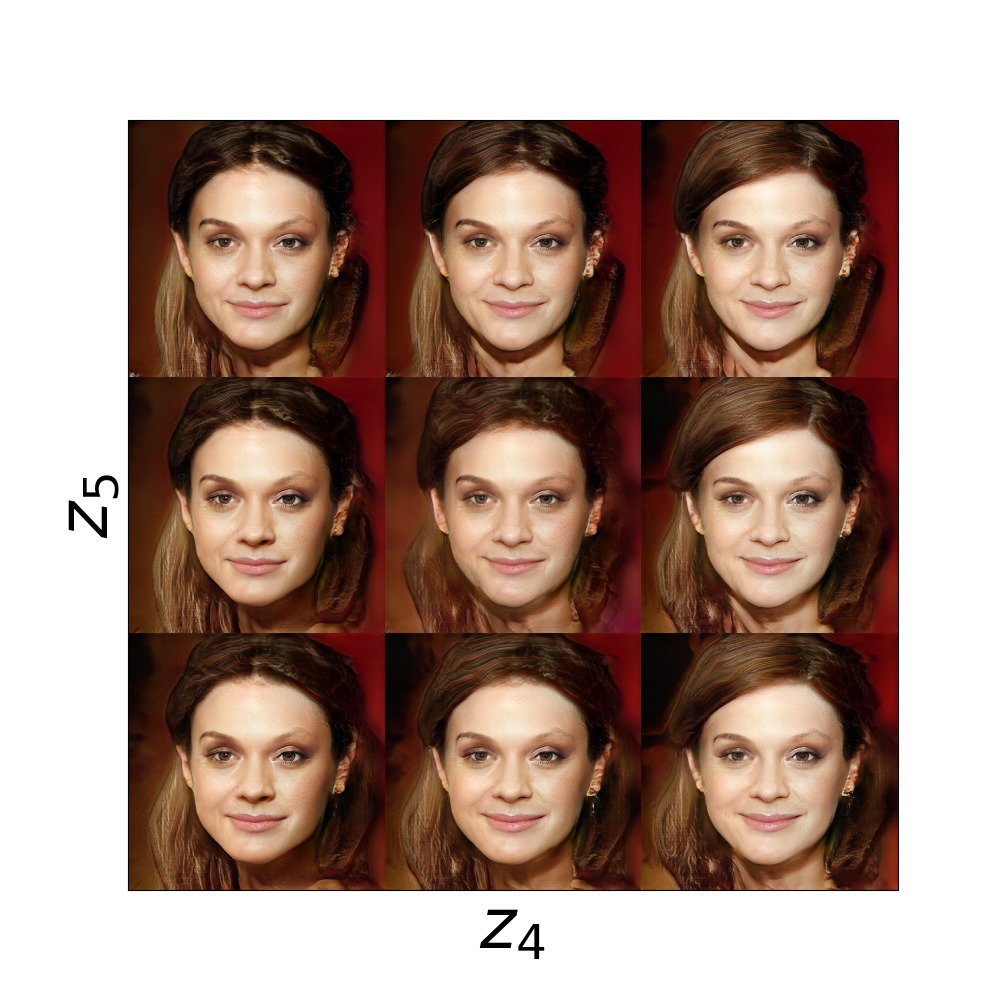}
    \end{tabularx}
    \vspace{-9pt}
    \subcaption{\normalsize PCAAE}
    \end{minipage}
    \vspace{-9pt}
    \caption{Interpolation in latent space of five parameters of $\beta$-TCVAE, FactorVAE and the PCAAE for the pre-train PGAN. Two components are adjusted along two axes, others are set to zeros. We can see that the PCAAE shows that each component of the proposed latent space represents one attribute of the generated images(\textit{e.g.} $z_1,z_2,z_3$ correspond to the hair colours, head poses and gender).
    }
    
    \label{fig:pca_gan_ex1}
    \vspace{-10pt}
\end{figure}

We now show our results of PCAAEs for organising the latent space of the pre-trained PGAN \cite{karras2017progressive}. To demonstrate the performance of our algorithm, we have trained the standard AE, the aforementioned VAE-based methods and our proposed PCAAE, using the procedure described in Section~\ref{sec:PCA_GAN}. Table~\ref{tab:table_evaluation_gan} shows the numeric evaluation of the methods and Figure \ref{fig:pca_gan_ex1} shows the generated images of the generator of PGAN from the latent spaces of the other approaches and of our proposed PCAAE. The other methods construct a latent space where the attributes of the generated images are correlated with more than one component. For example, we can see that the latent space of AE mixes up the attributes. In addition, it can be seen that the fourth parameter of $\beta$-TCVAE controls the hair color and the gender of generated images simultaneously. Respectively, the first parameter of FactorVAE changes the head pose. Then, the third one of this model still corresponds to the head pose. Indeed, the absolute PCC of this model for the head pose is correlated to the first and third components of the latent space. Our proposed PCAAE yields a disentangled latent space which is organised in a hierarchical fashion: the first component corresponds to the colour hair of the generated images, the second one represents head poses (\textit{e.g.} turning left and right), the third parameter corresponds to hair thickness and the last one is mildy correlated to skin tone.). Our PCAAE is able to efficiently separate the different facial attributes and rank them according to their importance in the reconstruction. Thus, the latent space created by our method is easier to interpret and navigate than the original GAN latent space.

We highlight that this procedure can be applied to any pre-trained model, so that the disentangling and organisation of the latent space can be carried out after the initial, computationally expensive, training of a GAN.

{}

\section{Conclusion}
\label{sec:conclusion}

In this paper, we have presented a novel autoencoder where the latent space is organised according to decreasing importance, and where these components are statistically independent. We refer to this network as a Principal Component Analysis Autoencoder - PCAAE. The autoencoder is trained with latent spaces of increasing sizes to ensure that we capture the properties of the data in decreasing order of importance, in an unsupervised manner. Furthermore, we have imposed statistical independence of the latent components by employing a covariance loss term, which we add to the standard autoencoder cost, to encourage a disentangled latent space. We have used synthetic data to illustrate that the PCAAE learns a latent space which is interpretable and which can be interpolated in a meaningful manner with respect to the properties inherent in the data. We have applied our autoencoder to high quality face data, and have shown that this efficiently disentangles the latent space of a powerful pre-trained GAN by projecting it to another smaller, interpretable, latent space. The resulting model can manipulate one facial attribute on each component. Furthermore, the proposed method can be applied to any pre-trained generative model, so that the initial time-consuming training of a powerful model and the organisation of its latent space can be carried out separately. We hope that this work will contribute to the interpretation and manipulation of latent spaces of complex data.

\section{Acknowlegement}
This work was funded by the DIGICOSME project.

%
%
\bibliographystyle{splncs}
\bibliography{refs_autoencoders}

\newpage
\clearpage

\section{Appendix}

\subsection{Extension of the PCAAE as a generative model: PCAWAE}

Recently, Wasserstein autoencoder (WAE) has been proposed as a new algorithm for building a generative model based on the latent variable \cite{tolstikhin2018wasserstein}. The WAE proposes to use the Kantorovich-Rubinstein duality \cite{arjovsky2017wasserstein} as an adversarial objective on the latent space. In this work, we apply the GAN-based penalty in \cite{tolstikhin2018wasserstein} in order to extend our PCAAE as a generative model.

Let us denote a prior $\mathcal{Z}_p$ (\textit{e.g.} a Gaussian distribution) and a random code $z \in \mathcal{R}^n$ which is generated randomly from $\mathcal{Z}_p$. In order to match the latent space of the proposed PCAAE $\mathcal{Z}$ with the prior distribution $\mathcal{Z}_p$, we apply the GAN-based penalty in \cite{tolstikhin2018wasserstein} by using a set of discriminators $C_i$, where $i=1,...,n$. Concretely, the $i^{th}$ discriminator attempts to distinguish the random codes which are sampled from $\mathcal{Z}_p$ and the generated latent codes of the $i^{th}$ autoencoder by ascending the following loss function:

\begin{equation}
\mathcal{L}_{C_i}(x) = \log(C_i(z_i)) + \log(1-C_i(E_i(x)))
\label{Equ:Ci}
\end{equation}

Meanwhile, the objective of the $i^{th}$ autoencoder is to minimise three loss functions: the reconstruction loss, covariance loss and adversarial loss, described as following:
\begin{equation}
\mathcal{L}_{AE_i}(x) = \lVert x - D_i \circ E_i(x)\rVert_2^2  \; + \; \sum_{j=1}^{i-1} \lambda_{cov}\mathcal{L}_{\text{cov}}(E_i(x),E_j(x)) - \lambda_{adv} \log(C_i(E_i(x)))
\label{Equ:AEi_sup}
\end{equation}

Thus, the min-max game between the $i^{th}$ discriminator and the $i^{th}$ autoencoder attempts to impose a prior distribution into the first $i$ components of the latent space of PCAAE. When the maximum size is reached (\textit{i.e.} $n$), the whole latent space of the autoencoder will be matched with the prior. We refer this method as PCAWAE.

\subsection{Limitation and Future works}
One limitation of our algorithm is that we increase the latent space size by one at each step. This can be problematic in some cases, where the autoencoder needs a certain amount of freedom to learn a useful representation. Therefore, we could consider increasing the latent space by small packets of codes, to give it the freedom it needs. It is clear that the use of the $\ell_2$ norm is not optimal to define the importance of a latent component. Indeed, in the case of the CelebA dataset as shown in Figure \ref{fig:pca_ae_celeba}, applying the PCAAE directly to the image data leads to very blurry results. Replacing the $\ell_2$ norm reconstruction loss by an alternative, perceptual, metric could provide better results. Finally, the application of a PCAAE trained in one region of a GAN latent space is not necessarily valid for another region. A future challenge will be to create a PCAAE which is applicable to the whole space of the GAN.

\begin{algorithm}[t]
\textbf{Require:}\\
Regularisation coefficient $\lambda_{cov}>0$. \\
Maximum latent space size $n$.\\
Initialise the parameters of the encoders $E_{\theta_i}$ and the decoders $D_{\theta_i}$.
	\BlankLine
	\While{$\theta_1$ not converged}{
		Sample $\left\{x_1, . . . , x_N\right\}$ from the training set. \\
		Update $E_{\theta_1}$ and $D_{\theta_1}$ by descending:\\
		\begin{equation*}
		\frac{1}{N}\sum_{k=1}^N ( x_k-D_{\theta_1}\circ E_{\theta_1}(x_k))^2 
		\end{equation*}
	}
	\For{$ i = 2 \dots n$}
	{
		\While{$\theta_i$ not converged}{
		Sample $\left\{x_1, . . . , x_N\right\}$ from the training set. \\
		Update $E_{\theta_i}$ and $D_{\theta_i}$ by descending:\\
		\begin{equation*}
		\frac{1}{N}\sum_{k=1}^N (x_k-D_{\theta_i}\circ E_{\theta_i}(x_k))^2_2 + \frac{\lambda_{cov}}{N(i-1)} \sum_{j=1}^{i-1} \sum_{k=1}^N E_{\theta_i}(x_k)E_{\theta_j}(x_k)
		\end{equation*}
	}
	}
\caption{PCAAE algorithm. Note, we have described the algorithm with a simple gradient descent, but any descent-based optimisation can be used (Adam, Adagrad etc)}
\label{algo:pcaAutoencoder}
\end{algorithm}

\begin{algorithm}[t]
\textbf{Require:}\\
Regularisation coefficient $\lambda_{cov}>0$, $\lambda_{adv}>0$ \\
Maximum latent space size $n$.\\
Initialise the parameters of the encoders $E_{\theta_i}$ and the decoders $D_{\theta_i}$. \\
Initialise the parameters of the discriminators $C_{\phi_i}$
	\BlankLine
	\While{$\theta_1$ not converged}{
		Sample $\left\{x_1, . . . , x_N\right\}$ from the training set. \\
		Sample $\left\{z_1, . . . , z_N\right\}$ from the prior. \\
		Update $C_{\phi_1}$ by ascending:
		\begin{equation*}
		\frac{\lambda_{adv}}{N}\sum_{k=1}^N \log(C_{\phi_1}(z_k)) +  \log(1-C_{\phi_1}(E_{\theta_1}(x_k)))
		\end{equation*}		
		\\
		Update $E_{\theta_1}$ and $D_{\theta_1}$ by descending:\\
		\begin{equation*}
		\frac{1}{N}\sum_{k=1}^N ( x_k-D_{\theta_1}\circ E_{\theta_1}(x_k))^2  - \lambda_{adv} \log(C_{\phi_1}(E_{\theta_1}(x_k)))
		\end{equation*}
	}
	\For{$ i = 2 \dots n$}
	{
		\While{$\theta_i$ not converged}{
		Sample $\left\{x_1, . . . , x_N\right\}$ from the training set. \\
		Sample $\left\{z_1, . . . , z_N\right\}$ from the prior. \\
		Update $C_{\phi_i}$ by ascending:
		\begin{equation*}
		\frac{\lambda_{adv}}{N}\sum_{k=1}^N \log(C_{\phi_i}(z_k)) +  \log(1-C_{\phi_i}(E_{\theta_i}(x_k)))
		\end{equation*}		
		\\		
		Update $E_{\theta_i}$ and $D_{\theta_i}$ by descending:\\
		\begin{equation*}
		\frac{1}{N}\sum_{k=1}^N (x_k-D_{\theta_i}\circ E_{\theta_i}(x_k))^2_2 - \lambda_{adv} \log(C_{\phi_i}(E_{\theta_i}(x_k))) + \frac{\lambda_{cov}}{N(i-1)} \sum_{j=1}^{i-1} \sum_{k=1}^N E_{\theta_i}(x_k)E_{\theta_j}(x_k)
		\end{equation*}
	}
	}
\caption{PCAWAE algorithm.}
\label{algo:pcawae}
\end{algorithm}

\subsection{Architecture of the proposed methods}

The pseudo-code for our algorithms can be seen in Algorithm~\ref{algo:pcaAutoencoder} and Algorithm~\ref{algo:pcawae}. Note that in this pseudo-code, we have used a standard gradient descent, but any gradient-descent based algorithm can be used (we used Adam \cite{kingma2014adam}).

For the geometrical structures, our autoencoder is a simple CNN with Leaky ReLUs ($\alpha=0.2$) and strided convolutions.  The size of the input image 64$\times$64 with 6 layers leading to a geometrical size of 1. The number of features is $(32,16,8,4,2,1)$ and all kernels of size 4. The decoder is symmetrical to that except that its number of features is multiplied by the current size of the latent space. Note that, for the purposes of fair comparison with other approaches, we used the same achitecture for the decoder of all methods.
  
For the latent variable, our discriminators contain 5 fully connected layers with Leaky ReLUs. The number features of the first discriminator is (8,8,8,8,1). The last activation is a sigmoid function. Then, the number of features of the next discriminators is multiplied those of the first network by the current size of the latent space.
  
In the case of the manipulation of the latent space of PGAN we use fully connected layers. Indeed, we are applying the PCAAE directly to a latent space, therefore convolutions are not appropriate in this case. The number of layers is 2, and again, the decoder is symmetrical to the encoder. The number of features in the case of the latent space of PGAN can be seen in Table~\ref{tab:ArchiTable_PCAGAN}. 
  
\begin{table}[b]
\footnotesize{
\begin{minipage}{0.48\linewidth}
\begin{tabular}{ |c|c|c| } 
 \hline
 \textbf{Encoder} & Act. & Output shape \\ 
 \hline
 Noise input & - & 512 \\ 
 Fully connected & LReLU & 64 \\ 
 Fully connected & LReLU & 1 \\ 
 BN (not training) & - & 1 \\
 \hline
\end{tabular}
\end{minipage}
\begin{minipage}{0.48\linewidth}
\begin{tabular}{ |c|c|c| } 
 \hline
 \textbf{Decoder} & Act. & Output shape \\ 
 \hline
 Latent space & - & n\_code \\ 
 Fully connected & LReLU & 64*(n\_code) \\ 
 Fully connected & - & 512 \\ 
 \hline
\end{tabular}
\end{minipage}
}
\vspace{0.2cm}
\caption{Architectures of the encoder and decoder of the PCAAE for a noise latent space with 512 parameters of the PGAN. n\_code denotes the number of components of the latent space. n\_code is set to 5 in all experiments.}
\label{tab:ArchiTable_PCAGAN}
\end{table}

\subsection{Results of the proposed methods for synthetic data}
\begin{figure}
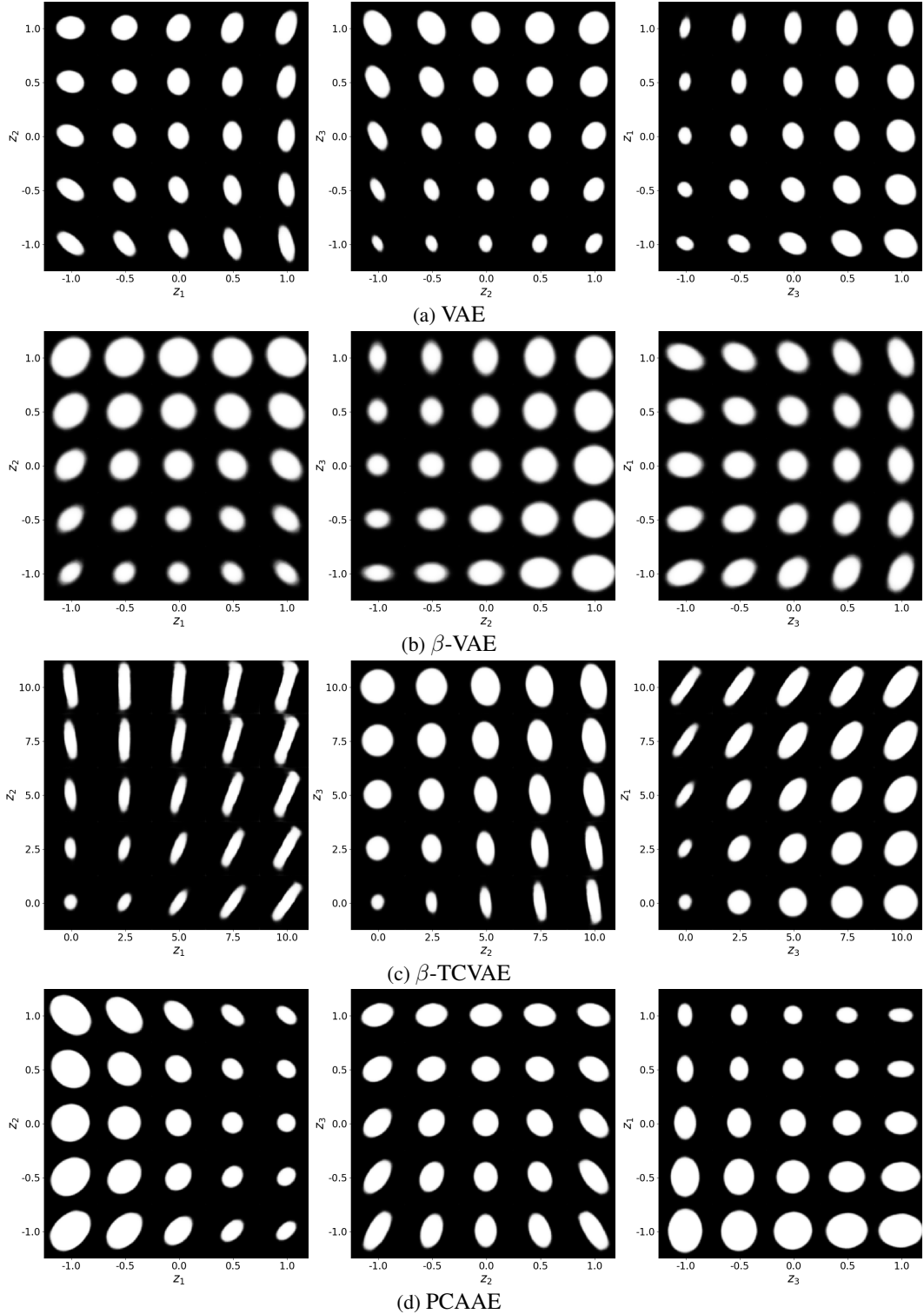

    \centering
    \begin{minipage}{1\textwidth}
    \begin{tabularx}{\linewidth}{ccc}
    \includegraphics[width=0.33\linewidth, trim={0.5cm 0.8cm 3.0cm 3.1cm}, clip]{images/pca_autoencoder/comparison/VAE/latent_z1z2.png}
    \includegraphics[width=0.33\linewidth, trim={0.5cm 0.8cm 3.0cm 3.1cm}, clip]{images/pca_autoencoder/comparison/VAE/latent_z2z3.png}
    \includegraphics[width=0.33\linewidth, trim={0.5cm 0.8cm 3.0cm 3.1cm}, clip]{images/pca_autoencoder/comparison/VAE/latent_z3z1.png}
    \end{tabularx}
    \vspace{-7pt}
    \subcaption{\normalsize VAE}
    \end{minipage}
    \\
    \begin{minipage}{1\textwidth}
    \begin{tabularx}{\linewidth}{ccc}
    \includegraphics[width=0.33\linewidth, trim={0.5cm 0.8cm 3.0cm 3.1cm}, clip]{images/pca_autoencoder/comparison/B_VAE/latent_z1z2.png}
    \includegraphics[width=0.33\linewidth, trim={0.5cm 0.8cm 3.0cm 3.1cm}, clip]{images/pca_autoencoder/comparison/B_VAE/latent_z2z3.png}
    
    \includegraphics[width=0.33\linewidth, trim={0.5cm 0.8cm 3.0cm 3.1cm}, clip]{images/pca_autoencoder/comparison/B_VAE/latent_z3z1.png}
    \end{tabularx}
    \vspace{-7pt}
    \subcaption{\normalsize $\beta$-VAE}
    \end{minipage}
    \\
    \begin{minipage}{\textwidth}
    \begin{tabularx}{\linewidth}{ccc}
    \includegraphics[width=0.33\linewidth, trim={0.5cm 0.8cm 3.0cm 3.1cm}, clip]{images/pca_autoencoder/comparison/btcvae/latent_z1z2.png}
    
    \includegraphics[width=0.33\linewidth, trim={0.5cm 0.8cm 3.0cm 3.1cm}, clip]{images/pca_autoencoder/comparison/btcvae/latent_z2z3.png}
    
    \includegraphics[width=0.33\linewidth, trim={0.5cm 0.8cm 3.0cm 3.1cm}, clip]{images/pca_autoencoder/comparison/btcvae/latent_z3z1.png}
    \end{tabularx}
    \vspace{-7pt}
    \subcaption{\normalsize $\beta$-TCVAE}
    \end{minipage}
    \\
    \begin{minipage}{1\textwidth}
    \begin{tabularx}{\linewidth}{ccc}
    \includegraphics[width=0.33\linewidth, trim={0.5cm 0.8cm 3.0cm 3.1cm}, clip]{images/pca_autoencoder/comparison/PCA_AE/latent_z1z2.png}
    \includegraphics[width=0.33\linewidth, trim={0.5cm 0.8cm 3.0cm 3.1cm}, clip]{images/pca_autoencoder/comparison/PCA_AE/latent_z2z3.png}
    \includegraphics[width=0.33\linewidth, trim={0.5cm 0.8cm 3.0cm 3.1cm}, clip]{images/pca_autoencoder/comparison/PCA_AE/latent_z3z1.png}
    \end{tabularx}
    \vspace{-7pt}
    \subcaption{\normalsize PCAAE}
    \end{minipage}
    \vspace{-7pt}
    \caption{Interpolation in latent space with respect to image reconstruction, ellipses with rotation (three parameters) of VAE, $\beta$-VAE, $\beta$-TCVAE and our proposed method. While VAE changes the area of ellipses by controlling the second and the third component of the latent space, disentangling methods and our proposed PCAAE use just one component for the area (\textit{e.g.} $\beta$-VAE and $\beta$-TCVAE take $Z_2$ and $Z_3$ for the area, respectively). Thus, the area is systematically found in the latent space of all methods which are proposed for unsupervised learning of disentangled representations.
    }
    
    \label{fig:3D_ellipses_rotation_apendix}
\end{figure}

We show the evaluations of the compared methods: those of AE, VAE \cite{kingma2014adam}, WAE \cite{tolstikhin2018wasserstein} , $\beta$-VAE$_B$ \cite{higgins2017beta}, $\beta$-VAE$_H$ \cite{burgess2018understanding}, FactorVAE \cite{kim2018disentangling}, $\beta$-TCVAE \cite{chen2018isolating} and our proposed PCAAE in Table~\ref{table:evaluation_ellipses_sup} and Figure~\ref{fig:3D_ellipses_rotation_apendix}. For a fair comparison, we use the same architecture for all decoders. Please see the code attached for more details. One can see that the first dimension ($z_1$) of our PCAAE always controls the area of the reconstructed ellipse. The last two components correspond to the orientation of the ellipse. The illustrations show how any two dimensions of our latent space have independent actions with respect to each other. In any column  of the grids shown in Figure 4 of our main paper, the effect of traveling up and down the axis is the same as in any other column (and similarly for the lines). Note that all unsupervised learning methods of disentangled representations such as $\beta$-VAE, FactorVAE, $\beta$-TCVAE and our proposed methods take systematically one component of the latent space for the area of ellipses.

Table \ref{table:evaluation_ellipses_sup} also shows an ablation study which compares the PCAAE with the baselines such as a standard AE, WAE and our PCAAE with no covariance loss (\textit{i.e.} $\lambda_{cov}=0$). We can see that more than one component of the latent space of AE, WAE and the PCAAE with no covariance loss controls the area of the ellipses. In the case of the PCAAE with no covariance loss, the first and the third component of its latent space correspond to the area attribute simultaneously. This confirms the need of the proposed covariance loss.

Thus, our method has efficiently organised and disentangled the latent space of the ellipses. This can be tested in our demo code.

\begin{table}[t]
  \centering
\begin{tabular}{|c|c|c|c|c|c|c|c|c|c|}
\hline  
\multirow{2}{*}{Com.}  &  \multicolumn{3}{|c|}{VAE} &  \multicolumn{3}{|c|}{$\beta$-VAE$_B$} &  \multicolumn{3}{|c|}{WAE} \\
\cline{2-10}

&A&R1&R2&A&R1&R2&A&R1&R2\\
\hline        
 $Z_1$    &  0.0043 &   \textbf{0.7826} & 0.1291 &  0.1542 &  0.0098 &  \textbf{0.6458} &  0.0381 & 0.5190 & \textbf{0.7844}  \\
 $Z_2$    &  0.3343 &  0.2094 &  \textbf{0.6746} &   \textbf{0.8950} &  0.1387 & 0.4236 &  0.4655 & \textbf{0.7213} & 0.3323 \\
 $Z_3$    &  \textbf{0.8284} &  0.2764 & 0.1503 &  0.0604 &   \textbf{0.7556} & 0.1604 & \textbf{0.7341} & 0.2857 & 0.4644  \\
\hline 
\hline 
\multirow{2}{*}{Com.}  &  \multicolumn{3}{|c|}{$\beta$-TCVAE} &  \multicolumn{3}{|c|}{FactorVAE} &  \multicolumn{3}{|c|}{Vanilla AE}\\
\cline{2-10}

&A&R1&R2&A&R1&R2&A&R1&R2\\
\hline        
 $Z_1$    &  0.1020 & 0.3523 & \textbf{0.5387} & 0.0139 & \textbf{0.7576} & 0.2787 & 0.5153 & 0.3668 & \textbf{0.9231} \\
 $Z_2$    &  0.1393 & \textbf{0.6640} & 0.2784 & 0.0732 & 0.3509 & \textbf{0.5838} & 0.0064 & \textbf{0.7372} & 0.2981 \\
 $Z_3$    &  \textbf{0.8581} & 0.1514 & 0.2340 & \textbf{0.8860} & 0.1727 & 0.3529 & \textbf{0.6427} & 0.3222 & 0.3642 \\
\hline 
\hline 
\multirow{2}{*}{Com.}  &  \multicolumn{3}{|c|}{PCAWAE} &  \multicolumn{3}{|c|}{PCAAE ($\lambda_{cov}=0$)} &  \multicolumn{3}{|c|}{PCAAE}\\
\cline{2-10}

&A&R1&R2&A&R1&R2&A&R1&R2\\
\hline        
 $Z_1$  & \textbf{0.9897} & 0.1072 & 0.1053 & \textbf{0.9907} & 0.1052 & 0.0073 & \textbf{0.9863} & 0.1497 & 0.1622 \\
 $Z_2$  & 0.1057 & \textbf{0.7694} & 0.1260 & 0.0075 & \textbf{0.6064} & 0.4800 & 0.0028 & 0.0566 & \textbf{0.6116} \\
 $Z_3$  &  0.1076 & 0.0528 & \textbf{0.8740}  & 0.5533 & 0.3355 & \textbf{0.5972} & 0.0035 & \textbf{0.7185} & 0.0591 \\
\hline 
\end{tabular}
\vspace{9pt}
\caption{Evaluation of the absolute PCC between the attributes of ellipses with respect to three components ($Z_1$,  $Z_2$ and  $Z_3$) of the trained latent space. We consider three attributes: the area (A), the ratio of two diameters towards vertical and horizontal directions (R1), the ratio of two diameters towards diagonal directions (R2). The strongest correlation of the components and the attributes is in bold font. We can see that each component of PCAAE and the extended version PCAWAE are correlated with only one ellipse attribute. Compared to other methods, our proposed methods choose to represent the area using the first component of the latent space: it has indeed organised the latent space in a hierarchical manner. Please note that there are many valid parameterisations of ellipses, and there is no reason for the other approaches to find R1 and R2. However, the area is systematically found in the latent space of all the disentangling methods.}
\label{table:evaluation_ellipses_sup}
\end{table}

{}

\subsection{Results of PCAAE for CelebA dataset}

\begin{figure*}
    \centering
    \begin{tabularx}{\linewidth}{cccccccccc}
    $z_1$:
    &
    \includegraphics[width=0.075\linewidth]{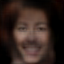}
    &
    \includegraphics[width=0.075\linewidth]{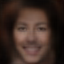}
    &
    \includegraphics[width=0.075\linewidth]{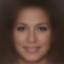}
    &
    \includegraphics[width=0.075\linewidth]{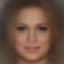}
    &
    \includegraphics[width=0.075\linewidth]{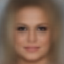}
    &
    \includegraphics[width=0.075\linewidth]{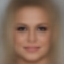}
    &
    \includegraphics[width=0.075\linewidth]{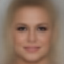}
    &
    \includegraphics[width=0.075\linewidth]{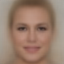}
    &
    \includegraphics[width=0.075\linewidth]{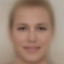}
    \\
     $z_2$:
    &
    \includegraphics[width=0.075\linewidth]{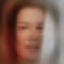}
    &
    \includegraphics[width=0.075\linewidth]{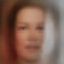}
    &
    \includegraphics[width=0.075\linewidth]{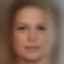}
    &
    \includegraphics[width=0.075\linewidth]{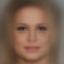}
    &
    \includegraphics[width=0.075\linewidth]{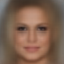}
    &
    \includegraphics[width=0.075\linewidth]{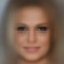}
    &
    \includegraphics[width=0.075\linewidth]{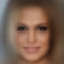}
    &
    \includegraphics[width=0.075\linewidth]{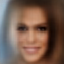}
    &
    \includegraphics[width=0.075\linewidth]{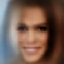}
    \\
    $z_3$:
    &
    \includegraphics[width=0.075\linewidth]{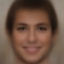}
    &
    \includegraphics[width=0.075\linewidth]{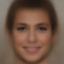}
    &
    \includegraphics[width=0.075\linewidth]{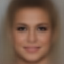}
    &
    \includegraphics[width=0.075\linewidth]{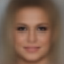}
    &
    \includegraphics[width=0.075\linewidth]{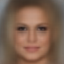}
    &
    \includegraphics[width=0.075\linewidth]{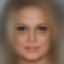}
    &
    \includegraphics[width=0.075\linewidth]{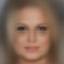}
    &
    \includegraphics[width=0.075\linewidth]{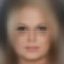}
    &
    \includegraphics[width=0.075\linewidth]{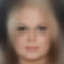}
    \\
    $z_4$:
    &
    \includegraphics[width=0.075\linewidth]{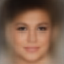}
    &
    \includegraphics[width=0.075\linewidth]{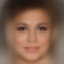}
    &
    \includegraphics[width=0.075\linewidth]{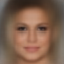}
    &
    \includegraphics[width=0.075\linewidth]{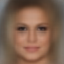}
    &
    \includegraphics[width=0.075\linewidth]{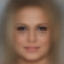}
    &
    \includegraphics[width=0.075\linewidth]{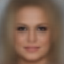}
    &
    \includegraphics[width=0.075\linewidth]{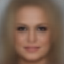}
    &
    \includegraphics[width=0.075\linewidth]{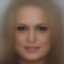}
    &
    \includegraphics[width=0.075\linewidth]{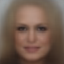}
    \\
    $z_5$:
    &
    \includegraphics[width=0.075\linewidth]{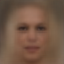}
    &
    \includegraphics[width=0.075\linewidth]{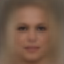}
    &
    \includegraphics[width=0.075\linewidth]{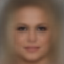}
    &
    \includegraphics[width=0.075\linewidth]{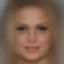}
    &
    \includegraphics[width=0.075\linewidth]{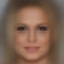}
    &
    \includegraphics[width=0.075\linewidth]{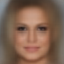}
    &
    \includegraphics[width=0.075\linewidth]{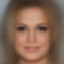}
    &
    \includegraphics[width=0.075\linewidth]{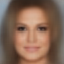}
    &
    \includegraphics[width=0.075\linewidth]{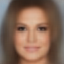}
    \\
    $z_6$:
    &
    \includegraphics[width=0.075\linewidth]{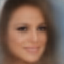}
    &
    \includegraphics[width=0.075\linewidth]{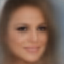}
    &
    \includegraphics[width=0.075\linewidth]{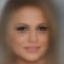}
    &
    \includegraphics[width=0.075\linewidth]{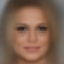}
    &
    \includegraphics[width=0.075\linewidth]{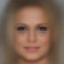}
    &
    \includegraphics[width=0.075\linewidth]{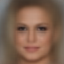}
    &
    \includegraphics[width=0.075\linewidth]{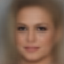}
    &
    \includegraphics[width=0.075\linewidth]{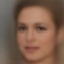}
    &
    \includegraphics[width=0.075\linewidth]{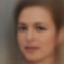}
    \\
    $z_7$:
    &
    \includegraphics[width=0.075\linewidth]{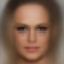}
    &
    \includegraphics[width=0.075\linewidth]{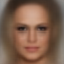}
    &
    \includegraphics[width=0.075\linewidth]{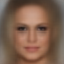}
    &
    \includegraphics[width=0.075\linewidth]{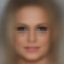}
    &
    \includegraphics[width=0.075\linewidth]{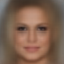}
    &
    \includegraphics[width=0.075\linewidth]{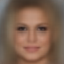}
    &
    \includegraphics[width=0.075\linewidth]{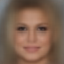}
    &
    \includegraphics[width=0.075\linewidth]{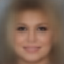}
    &
    \includegraphics[width=0.075\linewidth]{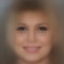}
    \\
    $z_8$:
    &
    \includegraphics[width=0.075\linewidth]{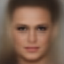}
    &
    \includegraphics[width=0.075\linewidth]{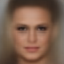}
    &
    \includegraphics[width=0.075\linewidth]{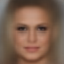}
    &
    \includegraphics[width=0.075\linewidth]{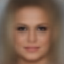}
    &
    \includegraphics[width=0.075\linewidth]{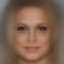}
    &
    \includegraphics[width=0.075\linewidth]{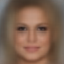}
    &
    \includegraphics[width=0.075\linewidth]{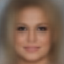}
    &
    \includegraphics[width=0.075\linewidth]{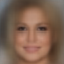}
    &
    \includegraphics[width=0.075\linewidth]{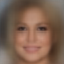}
    \\
    $z_9$:
    &
    \includegraphics[width=0.075\linewidth]{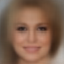}
    &
    \includegraphics[width=0.075\linewidth]{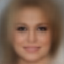}
    &
    \includegraphics[width=0.075\linewidth]{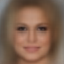}
    &
    \includegraphics[width=0.075\linewidth]{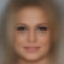}
    &
    \includegraphics[width=0.075\linewidth]{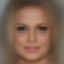}
    &
    \includegraphics[width=0.075\linewidth]{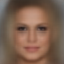}
    &
    \includegraphics[width=0.075\linewidth]{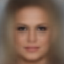}
    &
    \includegraphics[width=0.075\linewidth]{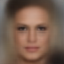}
    &
    \includegraphics[width=0.075\linewidth]{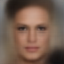}
    \\
    $z_{10}$:
    &
    \includegraphics[width=0.075\linewidth]{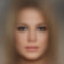}
    &
    \includegraphics[width=0.075\linewidth]{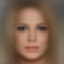}
    &
    \includegraphics[width=0.075\linewidth]{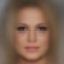}
    &
    \includegraphics[width=0.075\linewidth]{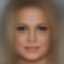}
    &
    \includegraphics[width=0.075\linewidth]{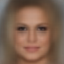}
    &
    \includegraphics[width=0.075\linewidth]{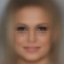}
    &
    \includegraphics[width=0.075\linewidth]{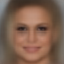}
    &
    \includegraphics[width=0.075\linewidth]{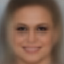}
    &
    \includegraphics[width=0.075\linewidth]{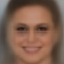}
    \end{tabularx}
    \caption{\textbf{Interpolation in latent space of ten parameters of a PCA autoencoder for CelebA dataset with the size image of $64 \times 64$}. The code shown in the left side is used to adjusted, other codes are set to zeros. The middle column corresponding the images with the codes of all zeros.}
    \label{fig:pca_ae_celeba}
\end{figure*}

In Figure \ref{fig:pca_ae_celeba}, we show the results of the PCAAE applied directly to images from the CelebA dataset. We can see that, while the PCAAE correctly organises the latent space (changing the average colour of the images in the first latent space component, for example), the results are overly smoothed. This is due to the complex nature of the CelebA dataset. Therefore, we found that a better approach was to apply the PCAAE to the latent space of a pre-trained GAN which is known to produce reliable results. These approaches can be seen in Section 4 \textbf{``PCAAE for GAN''} of our main paper.

\subsection{Interpolation in the original (entangled) latent space of PGAN}

\begin{figure}[h]
    \centering
    \tiny
    \begin{tabularx}{\linewidth}{ccccccccc}
    $\overline{\eta}_1-1.0$ & $\overline{\eta}_1-0.75$ & $\overline{\eta}_1-0.5$ & $\overline{\eta}_1-0.25$ & $\overline{\eta}_1$ & $\overline{\eta}_1+0.25$ & $\overline{\eta}_1+0.5$ & $\overline{\eta}_1+0.75$ & $\overline{\eta}_1+1.0$ \\
    \includegraphics[width=0.08\linewidth]{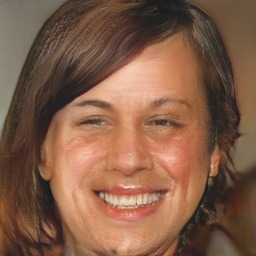}
    &
    \includegraphics[width=0.08\linewidth]{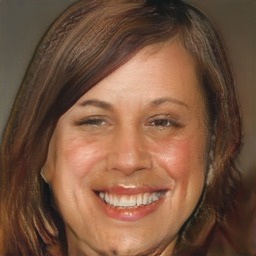}
    &
    \includegraphics[width=0.08\linewidth]{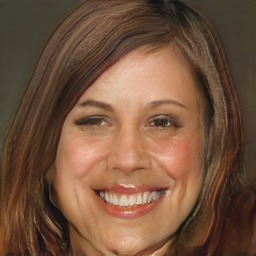}
    &
    \includegraphics[width=0.08\linewidth]{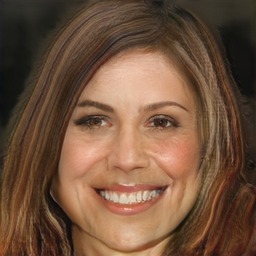}
    &
    \includegraphics[width=0.08\linewidth]{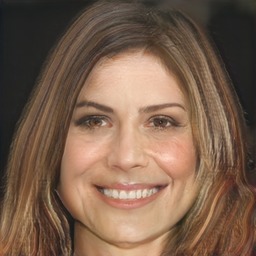}
    &
    \includegraphics[width=0.08\linewidth]{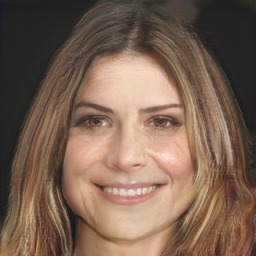}
    &
    \includegraphics[width=0.08\linewidth]{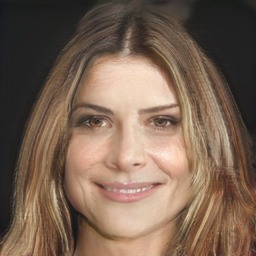}
    &
    \includegraphics[width=0.08\linewidth]{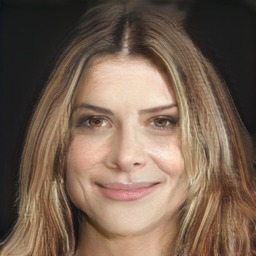}
    &
    \includegraphics[width=0.08\linewidth]{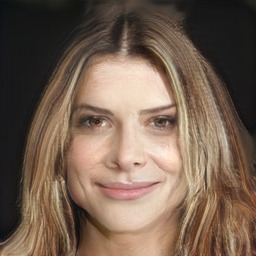}
    \\
    $\overline{\eta}_2-1.0$ & $\overline{\eta}_2-0.75$ & $\overline{\eta}_2-0.5$ & $\overline{\eta}_2-0.25$ & $\overline{\eta}_2$ & $\overline{\eta}_2+0.25$ & $\overline{\eta}_2+0.5$ & $\overline{\eta}_2+0.75$ & $\overline{\eta}_2+1.0$ \\
    \includegraphics[width=0.08\linewidth]{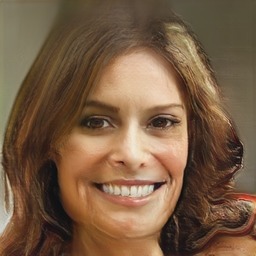}
    &
    \includegraphics[width=0.08\linewidth]{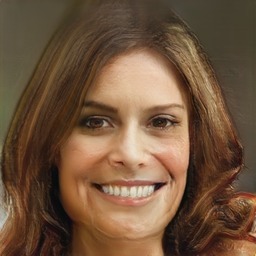}
    &
    \includegraphics[width=0.08\linewidth]{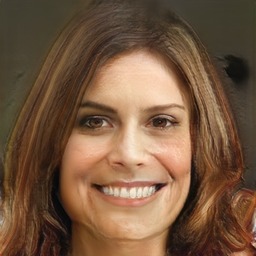}
    &
    \includegraphics[width=0.08\linewidth]{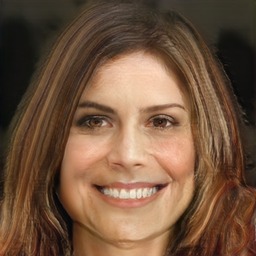}
    &
    \includegraphics[width=0.08\linewidth]{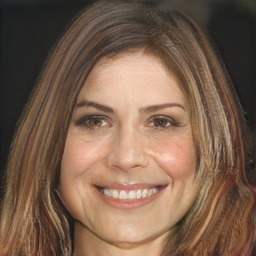}
    &
    \includegraphics[width=0.08\linewidth]{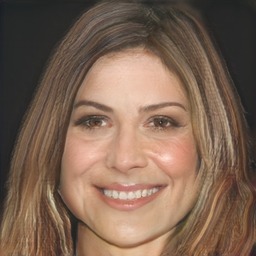}
    &
    \includegraphics[width=0.08\linewidth]{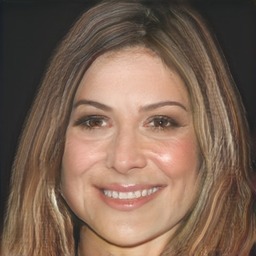}
    &
    \includegraphics[width=0.08\linewidth]{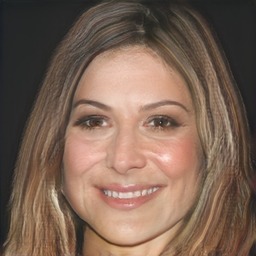}
    &
    \includegraphics[width=0.08\linewidth]{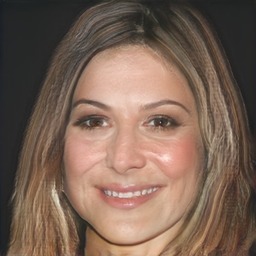}
    \\
    $\overline{\eta}_3-1.0$ & $\overline{\eta}_3-0.75$ & $\overline{\eta}_3-0.5$ & $\overline{\eta}_3-0.25$ & $\overline{\eta}_3$ & $\overline{\eta}_3+0.25$ & $\overline{\eta}_3+0.5$ & $\overline{\eta}_3+0.75$ & $\overline{\eta}_3+1.0$ \\
    \includegraphics[width=0.08\linewidth]{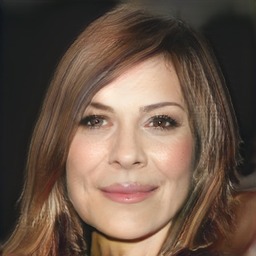}
    &
    \includegraphics[width=0.08\linewidth]{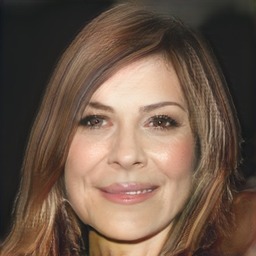}
    &
    \includegraphics[width=0.08\linewidth]{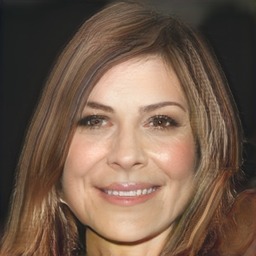}
    &
    \includegraphics[width=0.08\linewidth]{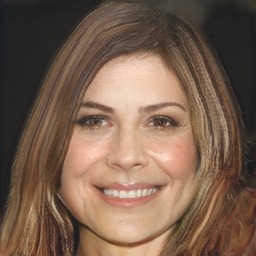}
    &
    \includegraphics[width=0.08\linewidth]{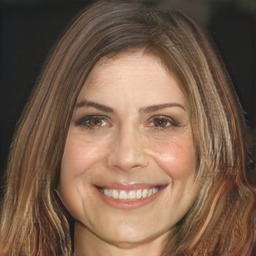}
    &
    \includegraphics[width=0.08\linewidth]{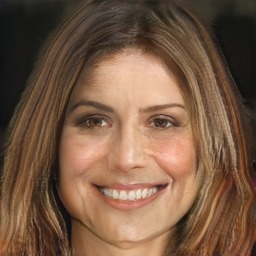}
    &
    \includegraphics[width=0.08\linewidth]{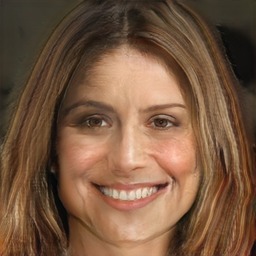}
    &
    \includegraphics[width=0.08\linewidth]{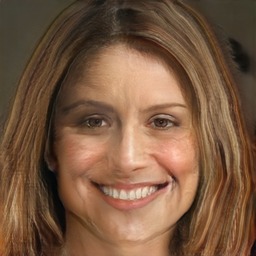}
    &
    \includegraphics[width=0.08\linewidth]{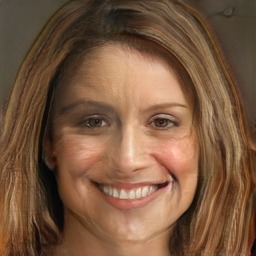}
    \\
    $\overline{\eta}_4-1.0$ & $\overline{\eta}_4-0.75$ & $\overline{\eta}_4-0.5$ & $\overline{\eta}_4-0.25$ & $\overline{\eta}_4$ & $\overline{\eta}_4+0.25$ & $\overline{\eta}_4+0.5$ & $\overline{\eta}_4+0.75$ & $\overline{\eta}_4+1.0$ \\
    \includegraphics[width=0.08\linewidth]{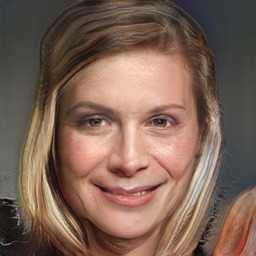}
    &
    \includegraphics[width=0.08\linewidth]{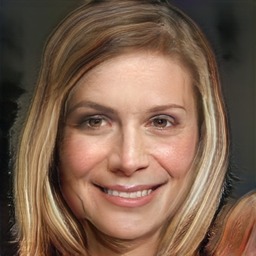}
    &
    \includegraphics[width=0.08\linewidth]{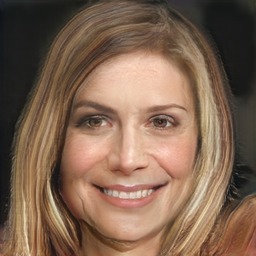}
    &
    \includegraphics[width=0.08\linewidth]{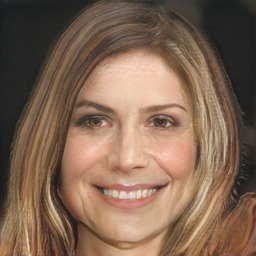}
    &
    \includegraphics[width=0.08\linewidth]{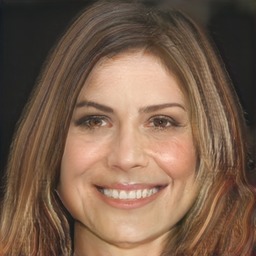}
    &
    \includegraphics[width=0.08\linewidth]{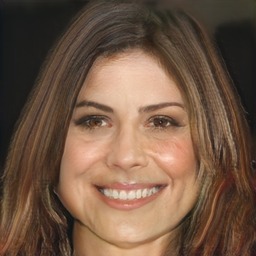}
    &
    \includegraphics[width=0.08\linewidth]{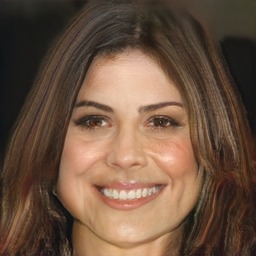}
    &
    \includegraphics[width=0.08\linewidth]{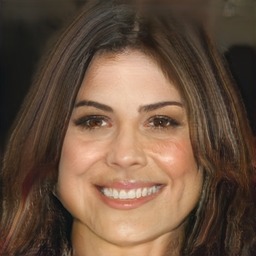}
    &
    \includegraphics[width=0.08\linewidth]{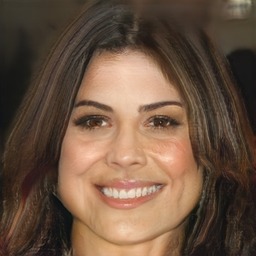}
    \\
    $\overline{\eta}_5-1.0$ & $\overline{\eta}_5-0.75$ & $\overline{\eta}_5-0.5$ & $\overline{\eta}_5-0.25$ & $\overline{\eta}_5$ & $\overline{\eta}_5+0.25$ & $\overline{\eta}_5+0.5$ & $\overline{\eta}_5+0.75$ & $\overline{\eta}_5+1.0$ \\
    \includegraphics[width=0.08\linewidth]{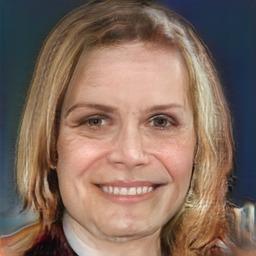}
    &
    \includegraphics[width=0.08\linewidth]{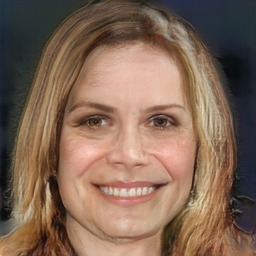}
    &
    \includegraphics[width=0.08\linewidth]{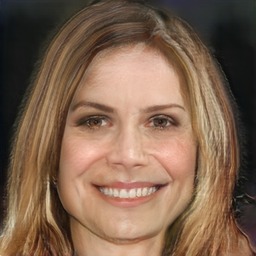}
    &
    \includegraphics[width=0.08\linewidth]{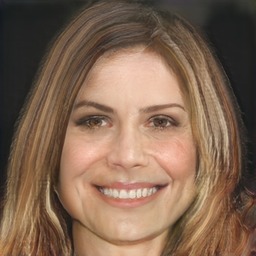}
    &
    \includegraphics[width=0.08\linewidth]{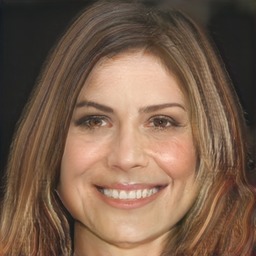}
    &
    \includegraphics[width=0.08\linewidth]{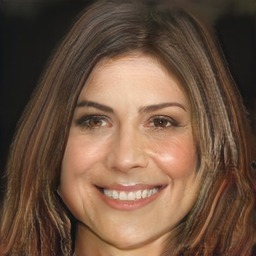}
    &
    \includegraphics[width=0.08\linewidth]{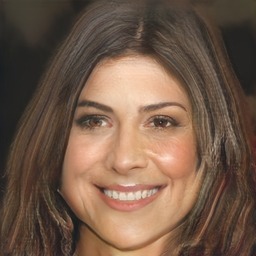}
    &
    \includegraphics[width=0.08\linewidth]{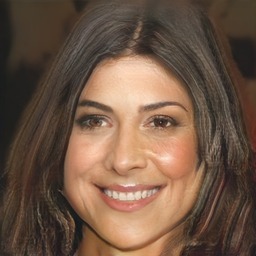}
    &
    \includegraphics[width=0.08\linewidth]{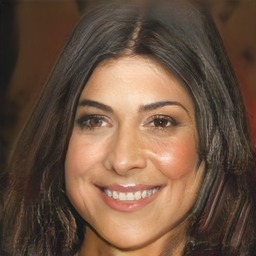}
    \\
    $\overline{\eta}_6-1.0$ & $\overline{\eta}_6-0.75$ & $\overline{\eta}_6-0.5$ & $\overline{\eta}_6-0.25$ & $\overline{\eta}_6$ & $\overline{\eta}_6+0.25$ & $\overline{\eta}_6+0.5$ & $\overline{\eta}_6+0.75$ & $\overline{\eta}_6+1.0$ \\
    \includegraphics[width=0.08\linewidth]{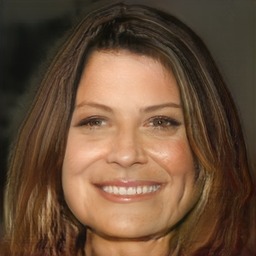}
    &
    \includegraphics[width=0.08\linewidth]{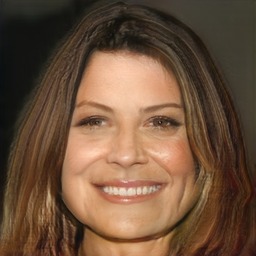}
    &
    \includegraphics[width=0.08\linewidth]{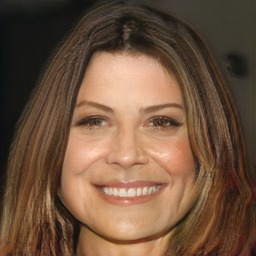}
    &
    \includegraphics[width=0.08\linewidth]{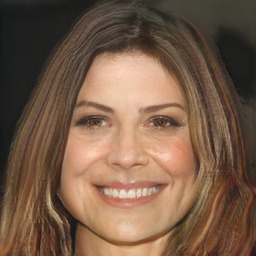}
    &
    \includegraphics[width=0.08\linewidth]{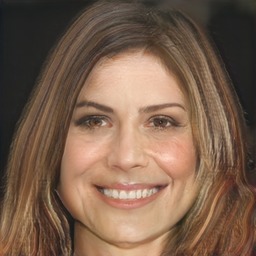}
    &
    \includegraphics[width=0.08\linewidth]{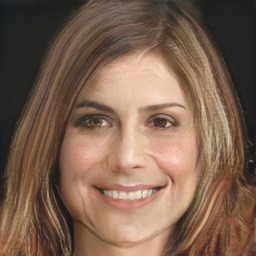}
    &
    \includegraphics[width=0.08\linewidth]{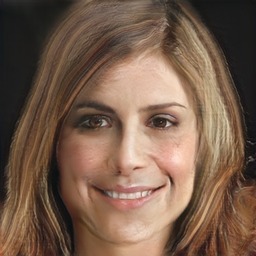}
    &
    \includegraphics[width=0.08\linewidth]{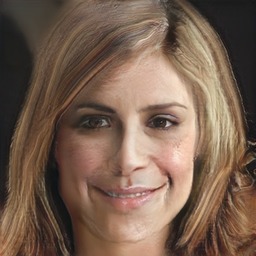}
    &
    \includegraphics[width=0.08\linewidth]{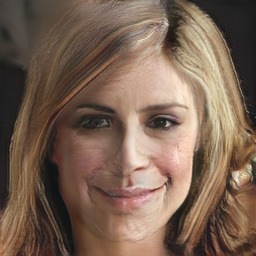}
    \\
    $...$ & $...$ & $...$ & $...$ & $...$ & $...$ & $...$ & $...$ & $...$
    \\
    $\overline{\eta}_{507}-1.0$ & $\overline{\eta}_{507}-0.75$ & $\overline{\eta}_{507}-0.5$ & $\overline{\eta}_{507}-0.25$ & $\overline{\eta}_{507}$ & $\overline{\eta}_{507}+0.25$ & $\overline{\eta}_{507}+0.5$ & $\overline{\eta}_{507}+0.75$ & $\overline{\eta}_{507}+1.0$ \\
    \includegraphics[width=0.08\linewidth]{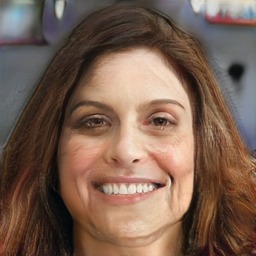}
    &
    \includegraphics[width=0.08\linewidth]{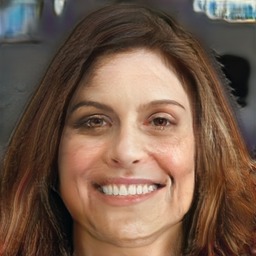}
    &
    \includegraphics[width=0.08\linewidth]{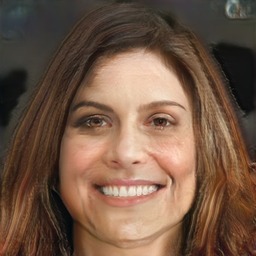}
    &
    \includegraphics[width=0.08\linewidth]{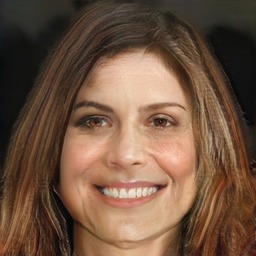}
    &
    \includegraphics[width=0.08\linewidth]{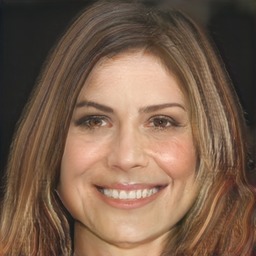}
    &
    \includegraphics[width=0.08\linewidth]{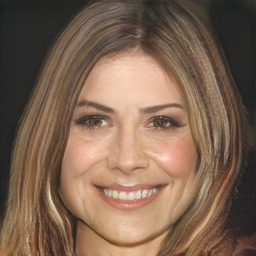}
    &
    \includegraphics[width=0.08\linewidth]{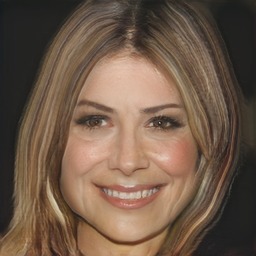}
    &
    \includegraphics[width=0.08\linewidth]{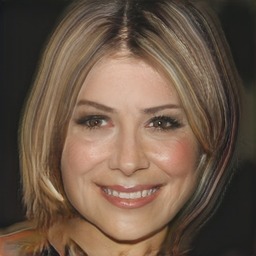}
    &
    \includegraphics[width=0.08\linewidth]{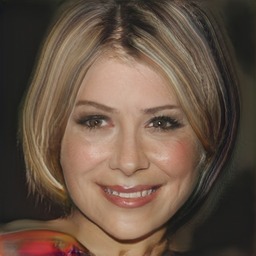}
    \\
    $\overline{\eta}_{508}-1.0$ & $\overline{\eta}_{508}-0.75$ & $\overline{\eta}_{508}-0.5$ & $\overline{\eta}_{508}-0.25$ & $\overline{\eta}_{508}$ & $\overline{\eta}_{508}+0.25$ & $\overline{\eta}_{508}+0.5$ & $\overline{\eta}_{508}+0.75$ & $\overline{\eta}_{508}+1.0$ \\
    \includegraphics[width=0.08\linewidth]{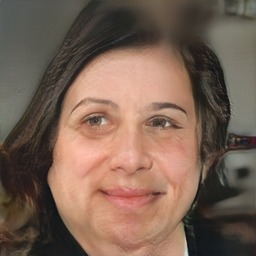}
    &
    \includegraphics[width=0.08\linewidth]{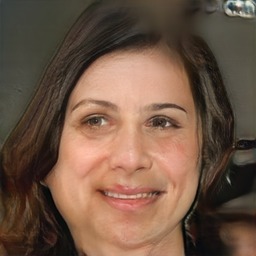}
    &
    \includegraphics[width=0.08\linewidth]{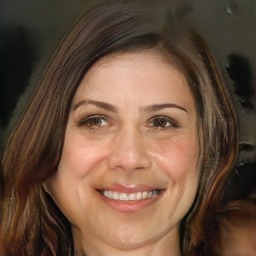}
    &
    \includegraphics[width=0.08\linewidth]{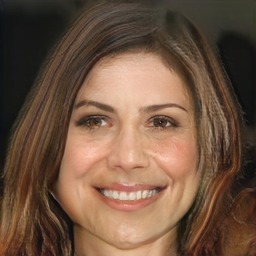}
    &
    \includegraphics[width=0.08\linewidth]{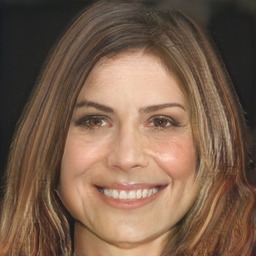}
    &
    \includegraphics[width=0.08\linewidth]{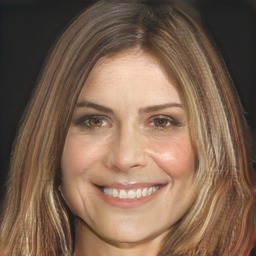}
    &
    \includegraphics[width=0.08\linewidth]{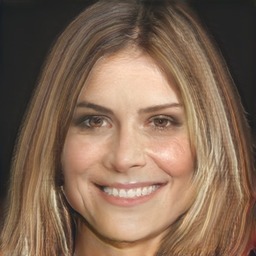}
    &
    \includegraphics[width=0.08\linewidth]{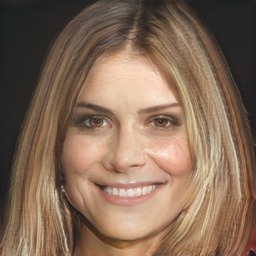}
    &
    \includegraphics[width=0.08\linewidth]{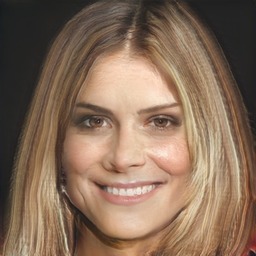}
    \\
    $\overline{\eta}_{509}-1.0$ & $\overline{\eta}_{509}-0.75$ & $\overline{\eta}_{509}-0.5$ & $\overline{\eta}_{509}-0.25$ & $\overline{\eta}_{509}$ & $\overline{\eta}_{509}+0.25$ & $\overline{\eta}_{509}+0.5$ & $\overline{\eta}_{509}+0.75$ & $\overline{\eta}_{509}+1.0$ \\
    \includegraphics[width=0.08\linewidth]{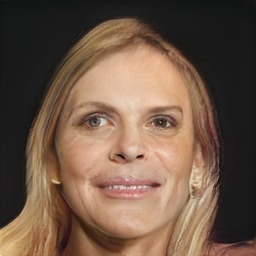}
    &
    \includegraphics[width=0.08\linewidth]{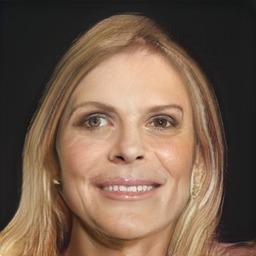}
    &
    \includegraphics[width=0.08\linewidth]{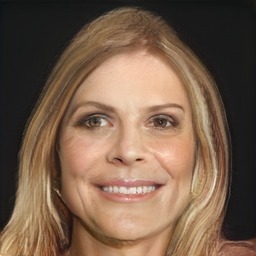}
    &
    \includegraphics[width=0.08\linewidth]{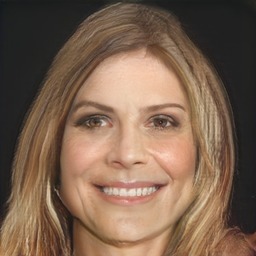}
    &
    \includegraphics[width=0.08\linewidth]{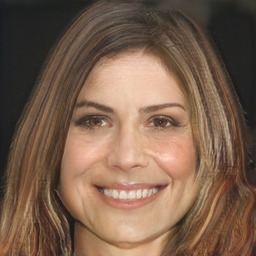}
    &
    \includegraphics[width=0.08\linewidth]{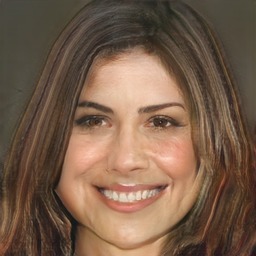}
    &
    \includegraphics[width=0.08\linewidth]{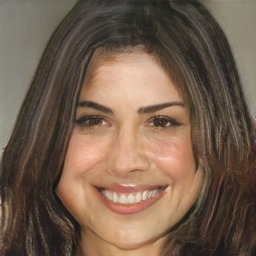}
    &
    \includegraphics[width=0.08\linewidth]{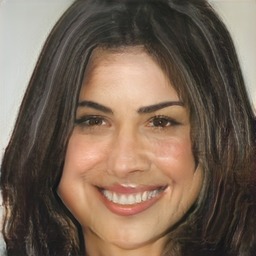}
    &
    \includegraphics[width=0.08\linewidth]{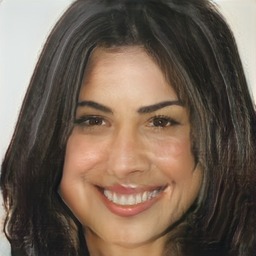}
    \\
    $\overline{\eta}_{510}-1.0$ & $\overline{\eta}_{510}-0.75$ & $\overline{\eta}_{510}-0.5$ & $\overline{\eta}_{510}-0.25$ & $\overline{\eta}_{510}$ & $\overline{\eta}_{510}+0.25$ & $\overline{\eta}_{510}+0.5$ & $\overline{\eta}_{510}+0.75$ & $\overline{\eta}_{510}+1.0$ \\
    \includegraphics[width=0.08\linewidth]{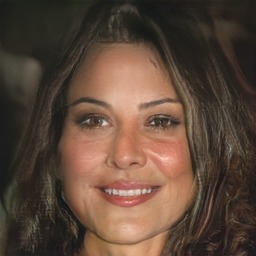}
    &
    \includegraphics[width=0.08\linewidth]{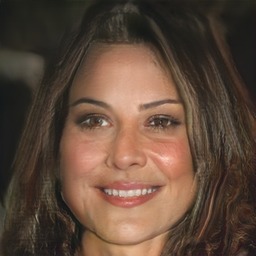}
    &
    \includegraphics[width=0.08\linewidth]{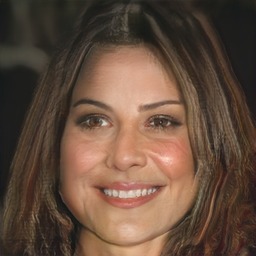}
    &
    \includegraphics[width=0.08\linewidth]{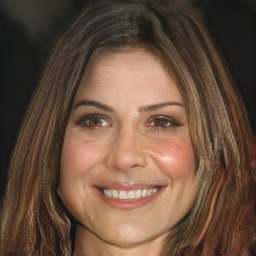}
    &
    \includegraphics[width=0.08\linewidth]{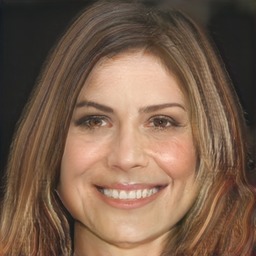}
    &
    \includegraphics[width=0.08\linewidth]{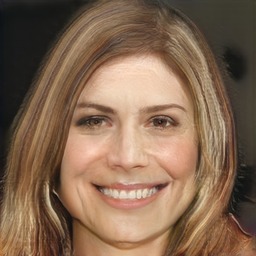}
    &
    \includegraphics[width=0.08\linewidth]{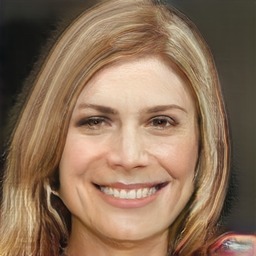}
    &
    \includegraphics[width=0.08\linewidth]{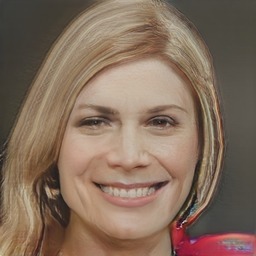}
    &
    \includegraphics[width=0.08\linewidth]{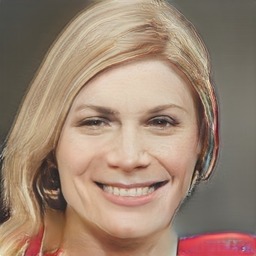}
    \\
    $\overline{\eta}_{511}-1.0$ & $\overline{\eta}_{511}-0.75$ & $\overline{\eta}_{511}-0.5$ & $\overline{\eta}_{511}-0.25$ & $\overline{\eta}_{511}$ & $\overline{\eta}_{511}+0.25$ & $\overline{\eta}_{511}+0.5$ & $\overline{\eta}_{511}+0.75$ & $\overline{\eta}_{511}+1.0$ \\
    \includegraphics[width=0.08\linewidth]{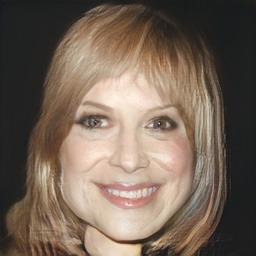}
    &
    \includegraphics[width=0.08\linewidth]{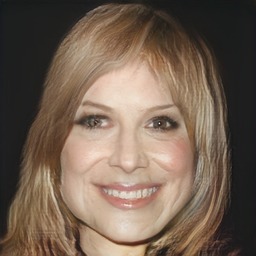}
    &
    \includegraphics[width=0.08\linewidth]{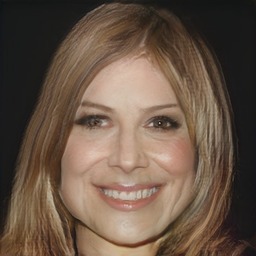}
    &
    \includegraphics[width=0.08\linewidth]{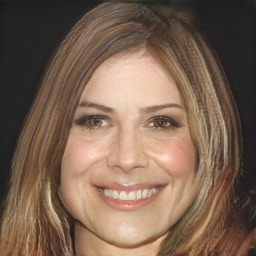}
    &
    \includegraphics[width=0.08\linewidth]{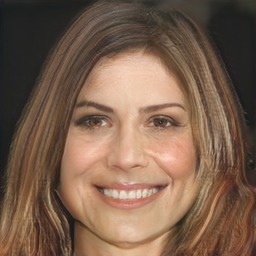}
    &
    \includegraphics[width=0.08\linewidth]{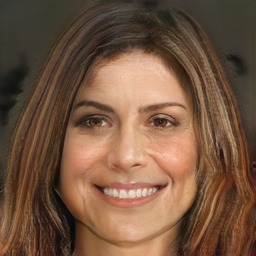}
    &
    \includegraphics[width=0.08\linewidth]{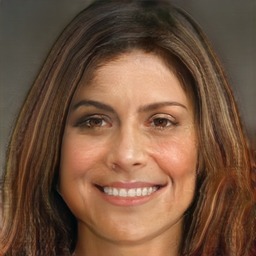}
    &
    \includegraphics[width=0.08\linewidth]{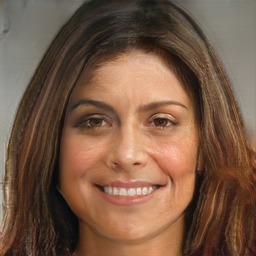}
    &
    \includegraphics[width=0.08\linewidth]{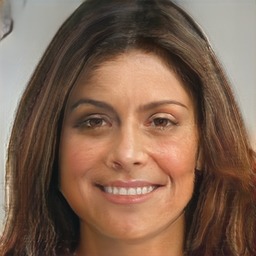}
    \\
    $\overline{\eta}_{512}-1.0$ & $\overline{\eta}_{512}-0.75$ & $\overline{\eta}_{512}-0.5$ & $\overline{\eta}_{512}-0.25$ & $\overline{\eta}_{512}$ & $\overline{\eta}_{512}+0.25$ & $\overline{\eta}_{512}+0.5$ & $\overline{\eta}_{512}+0.75$ & $\overline{\eta}_{512}+1.0$ \\
    \includegraphics[width=0.08\linewidth]{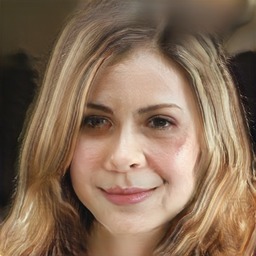}
    &
    \includegraphics[width=0.08\linewidth]{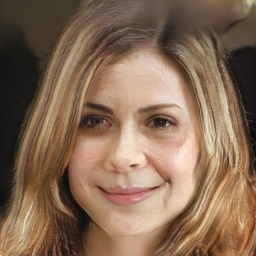}
    &
    \includegraphics[width=0.08\linewidth]{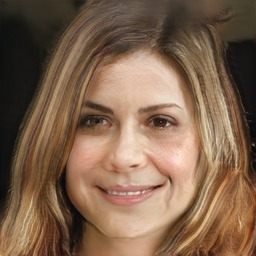}
    &
    \includegraphics[width=0.08\linewidth]{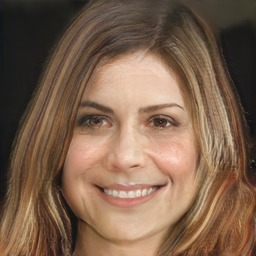}
    &
    \includegraphics[width=0.08\linewidth]{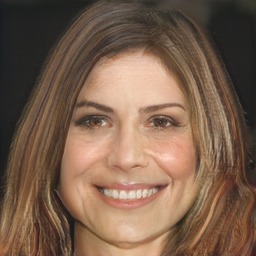}
    &
    \includegraphics[width=0.08\linewidth]{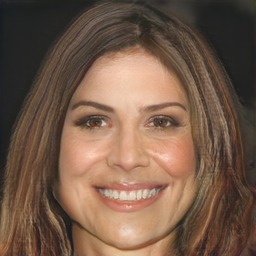}
    &
    \includegraphics[width=0.08\linewidth]{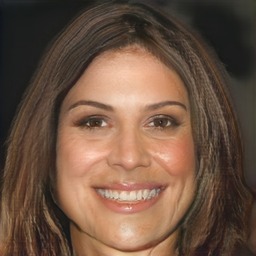}
    &
    \includegraphics[width=0.08\linewidth]{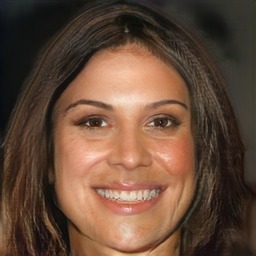}
    &
    \includegraphics[width=0.08\linewidth]{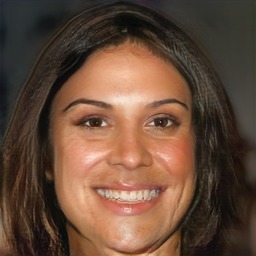}
    \end{tabularx}
    \caption{\textbf{Interpolation in the original latent space of PGAN (with 512 parameters)}. From the initial code $\overline{\eta}=[\overline{\eta}_1,\overline{\eta}_2,\overline{\eta}_3,...,\overline{\eta}_{512}]$ as shown in the middle column, we adjust the $n^{th}$ component by adding a constant shown above the image, other codes are not shown that are fixed. We can see that it is difficult to interpret this latent space. Several attributes such as hair colour or head pose are varied within the same component of the latent space of PGAN.}
    \label{fig:interpolation_gan_extend}
\end{figure}

In Figure \ref{fig:interpolation_gan_extend}  we show several examples of interpolation in the original latent space of PGAN, before applying our PCAAE. We visualise images generated by PGAN while varying one latent component at a time. We can see that it is difficult to interpret this latent space. For example, we can see a blond woman at both the first and the last parameters of the latent space, and the woman in the generated images changes the pose of her head when either the second and last components are varied. It is clear that this latent space is heavily entangled, with several characteristics modified by changing one component. This makes it difficult to understand, navigate and manipulate the latent space. Addressing these problems is precisely the goal of the present work.

\subsection{Further examples of navigating the PGAN latent space}

\begin{figure}
    \centering
    \begin{minipage}{\textwidth}
    \begin{tabularx}{\linewidth}{ccc}
    \includegraphics[width=0.24\linewidth, trim={1.7cm 1.0cm 3.0cm 3.1cm}, clip]{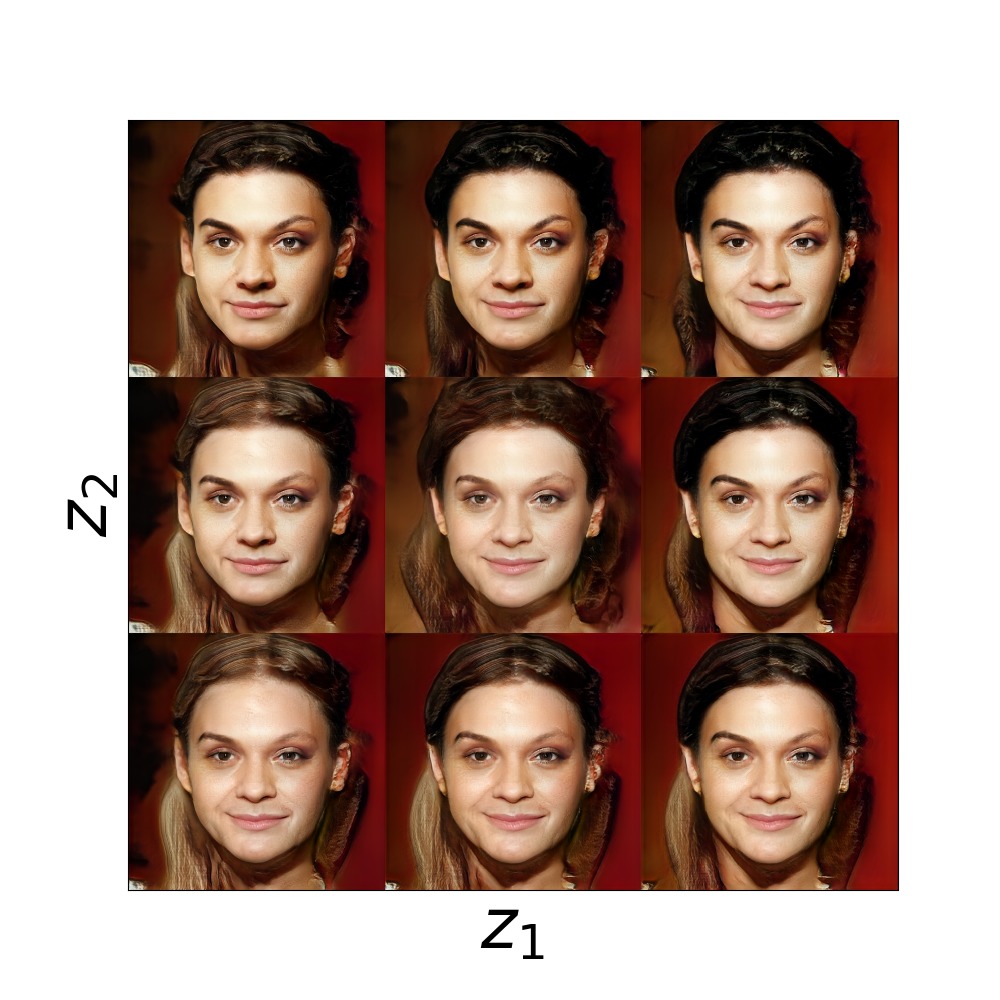}
    \includegraphics[width=0.24\linewidth, trim={1.7cm 1.0cm 3.0cm 3.1cm}, clip]{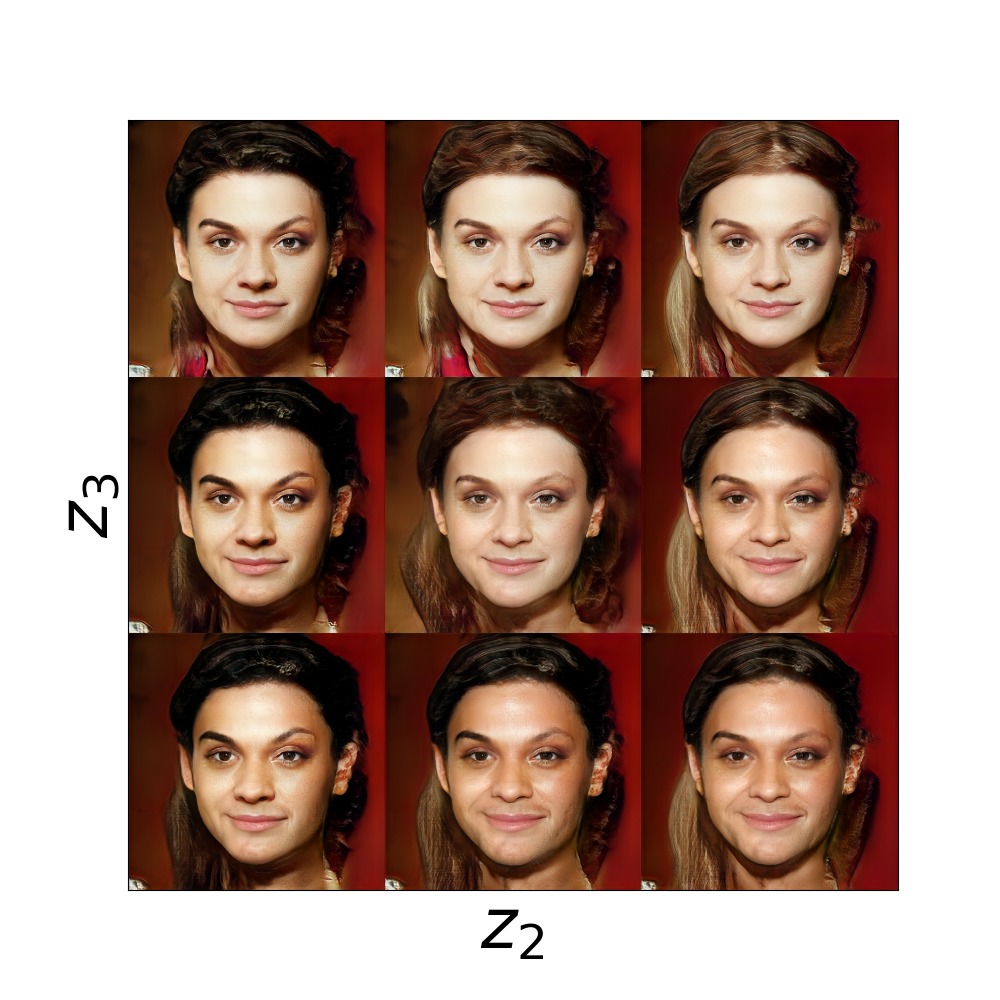}
    \includegraphics[width=0.24\linewidth, trim={1.7cm 1.0cm 3.0cm 3.1cm}, clip]{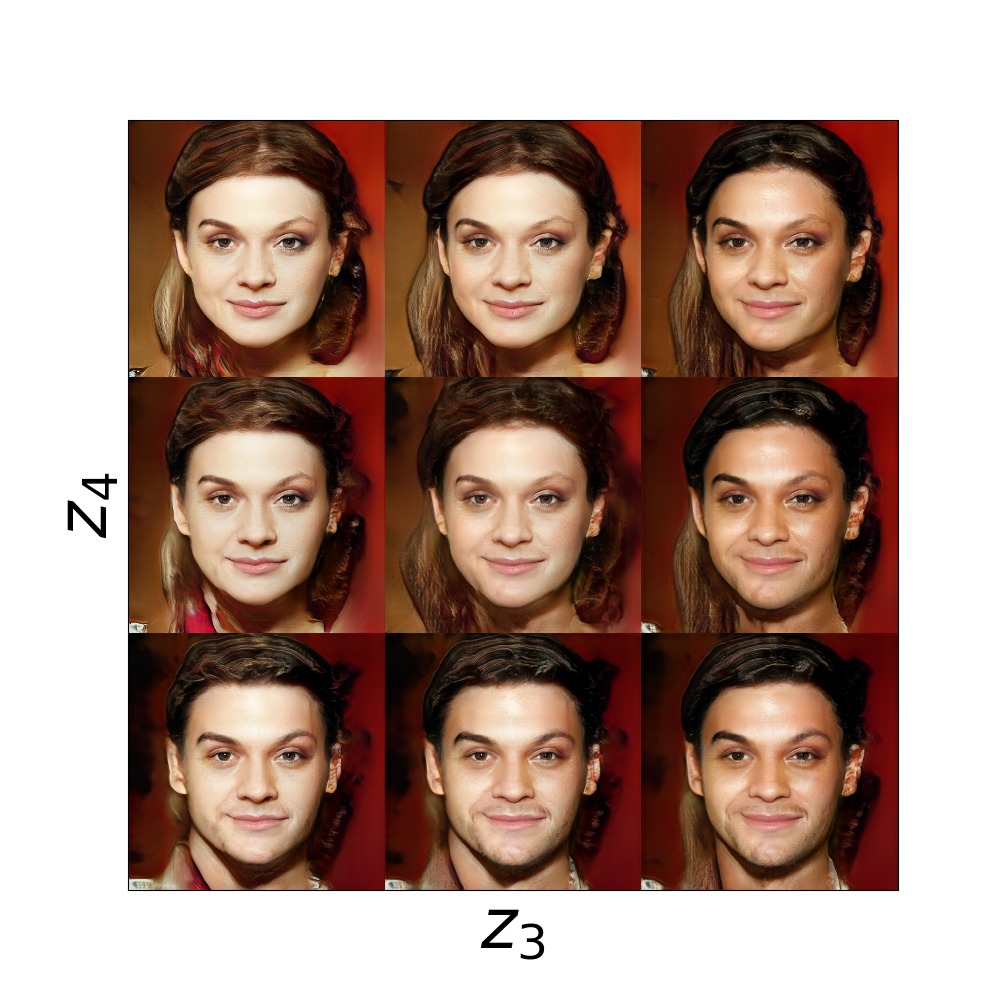}
    \includegraphics[width=0.24\linewidth, trim={1.7cm 1.0cm 3.0cm 3.1cm}, clip]{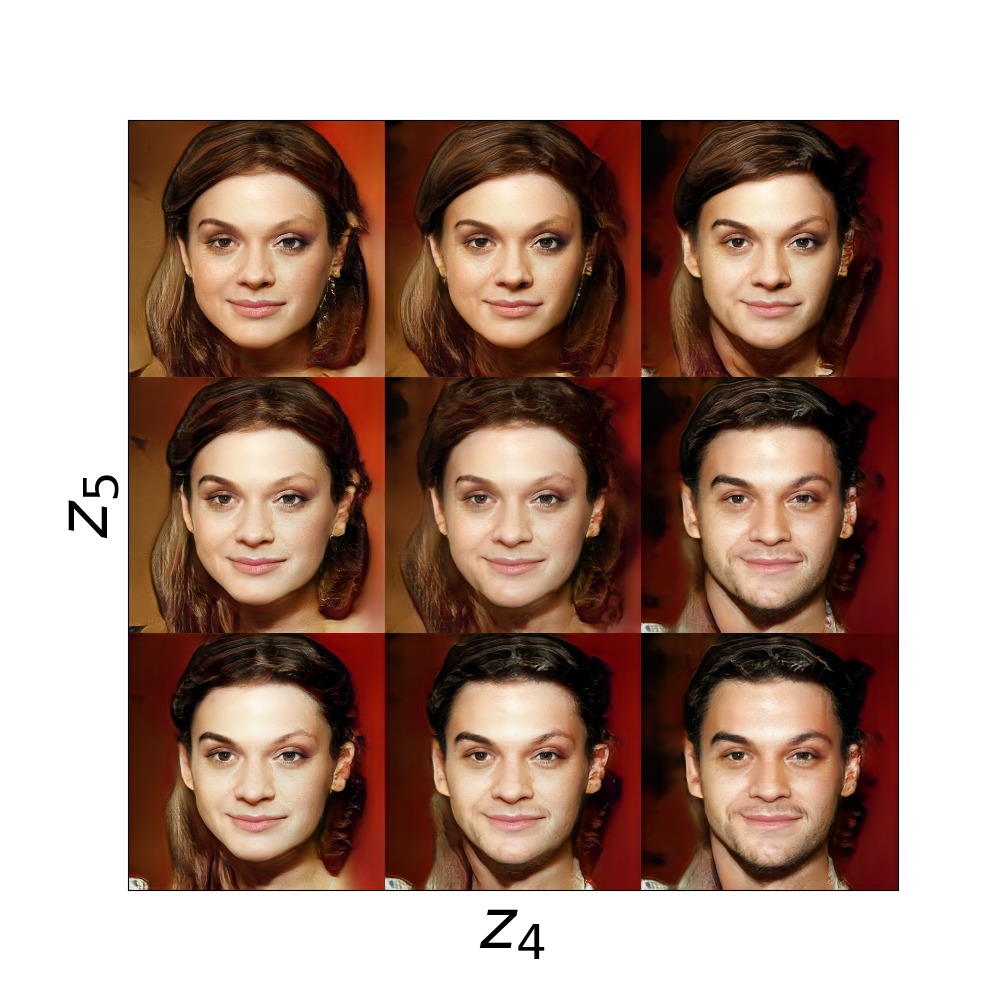}
    \end{tabularx}
    \vspace{-9pt}
    \subcaption{\normalsize AE}
    \end{minipage}
    \\
    \begin{minipage}{\textwidth}
    \begin{tabularx}{\linewidth}{ccc}
    \includegraphics[width=0.24\linewidth, trim={1.7cm 1.0cm 3.0cm 3.1cm}, clip]{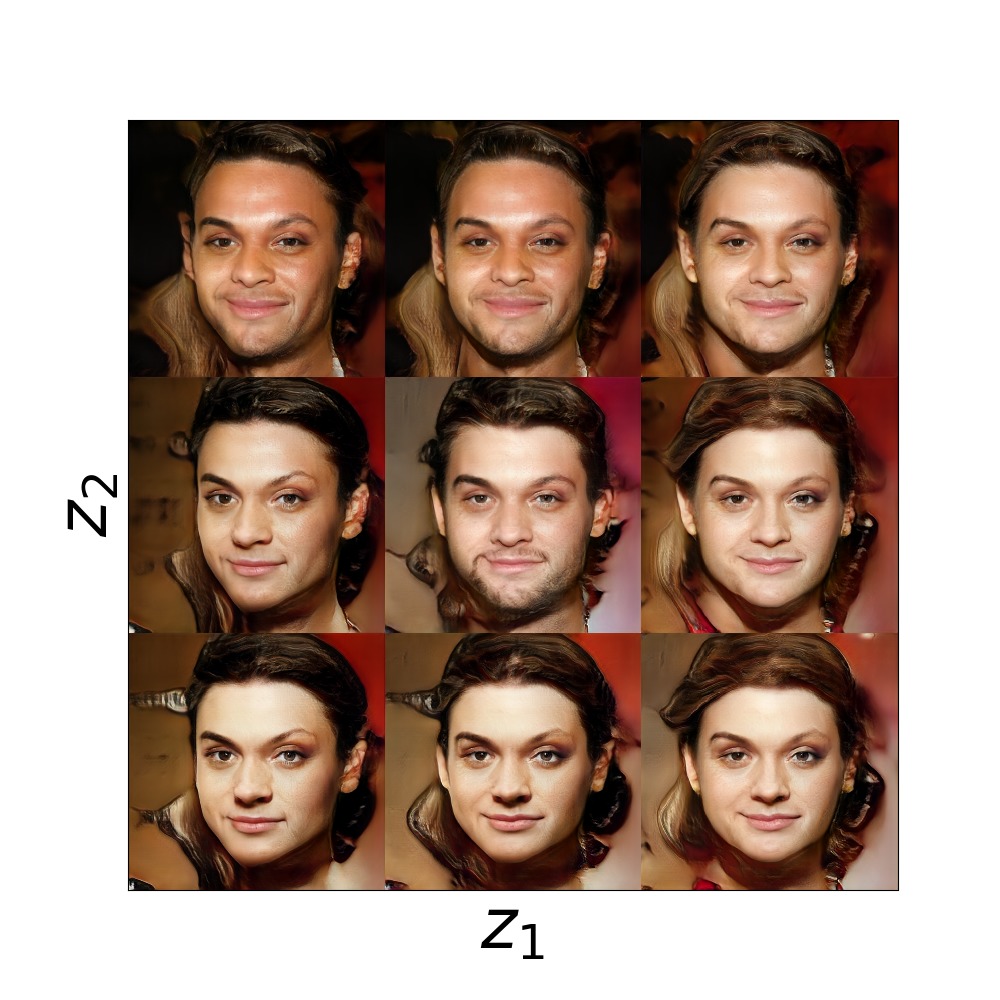}
    \includegraphics[width=0.24\linewidth, trim={1.7cm 1.0cm 3.0cm 3.1cm}, clip]{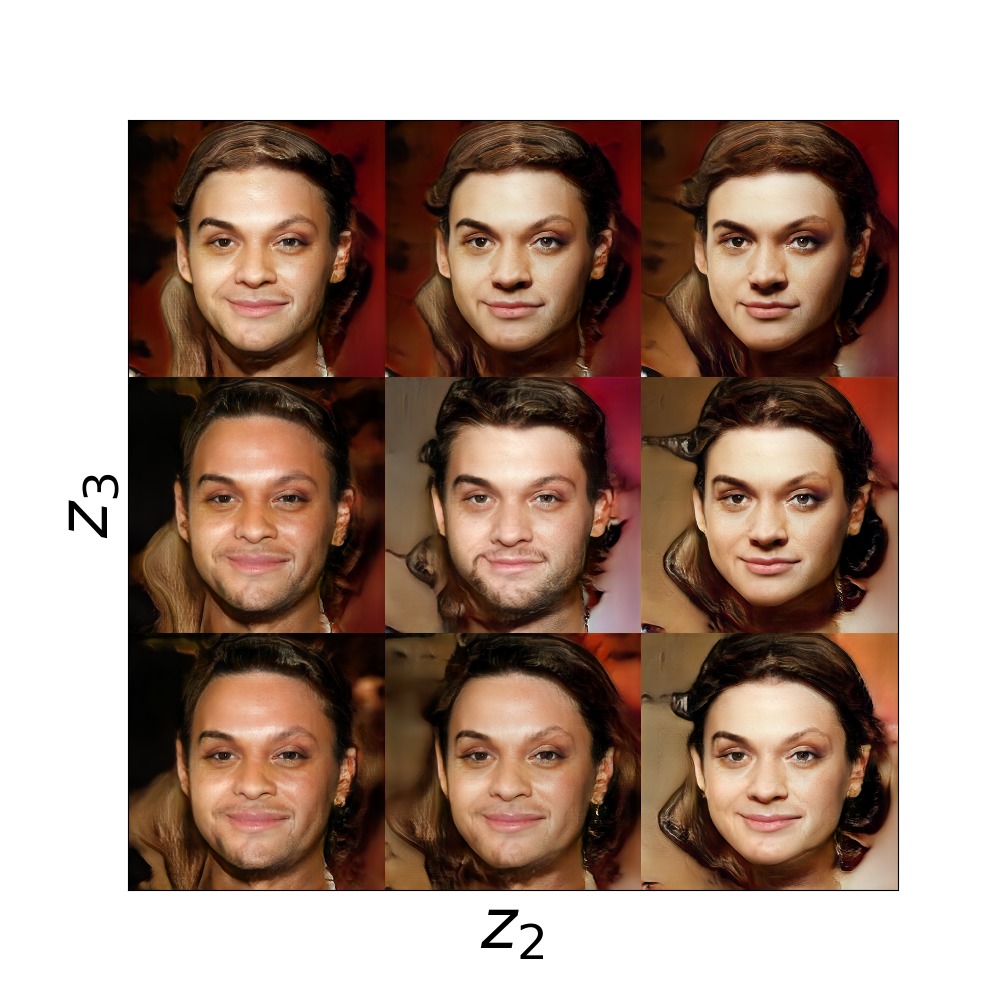}
    \includegraphics[width=0.24\linewidth, trim={1.7cm 1.0cm 3.0cm 3.1cm}, clip]{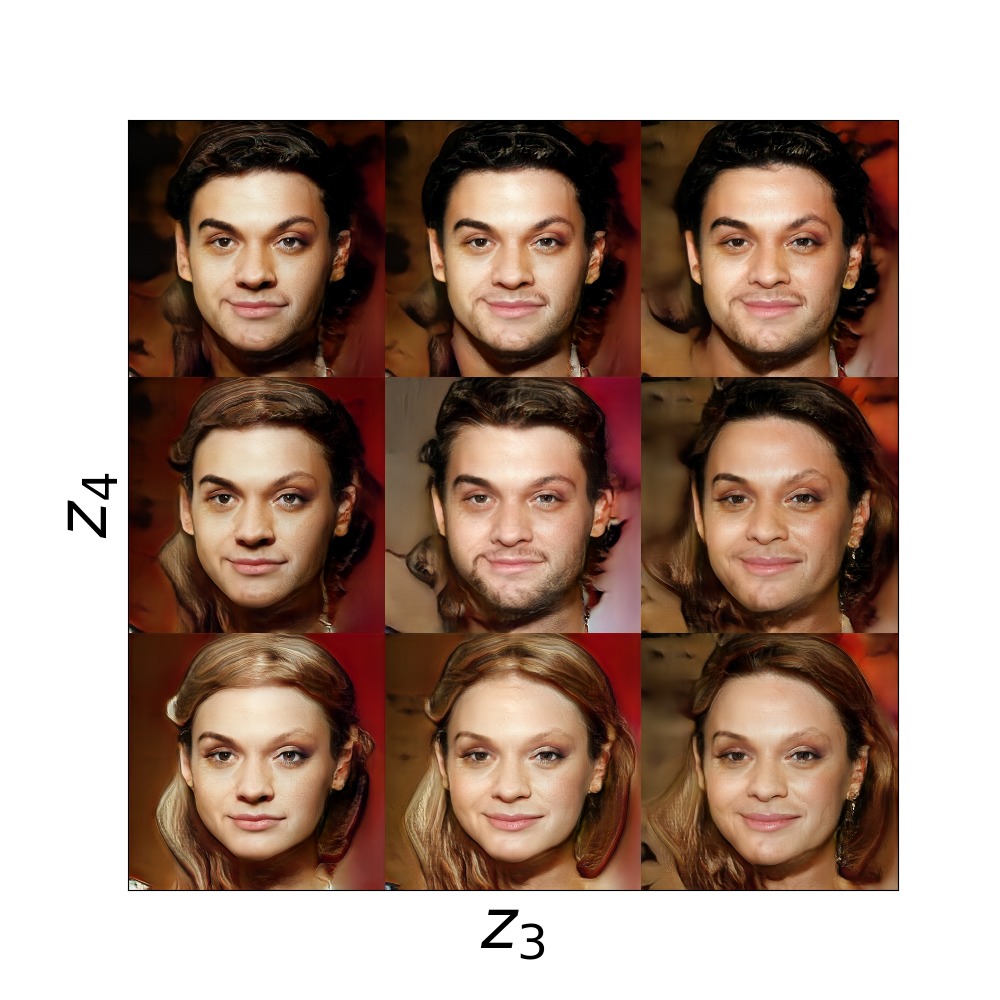}
    \includegraphics[width=0.24\linewidth, trim={1.7cm 1.0cm 3.0cm 3.1cm}, clip]{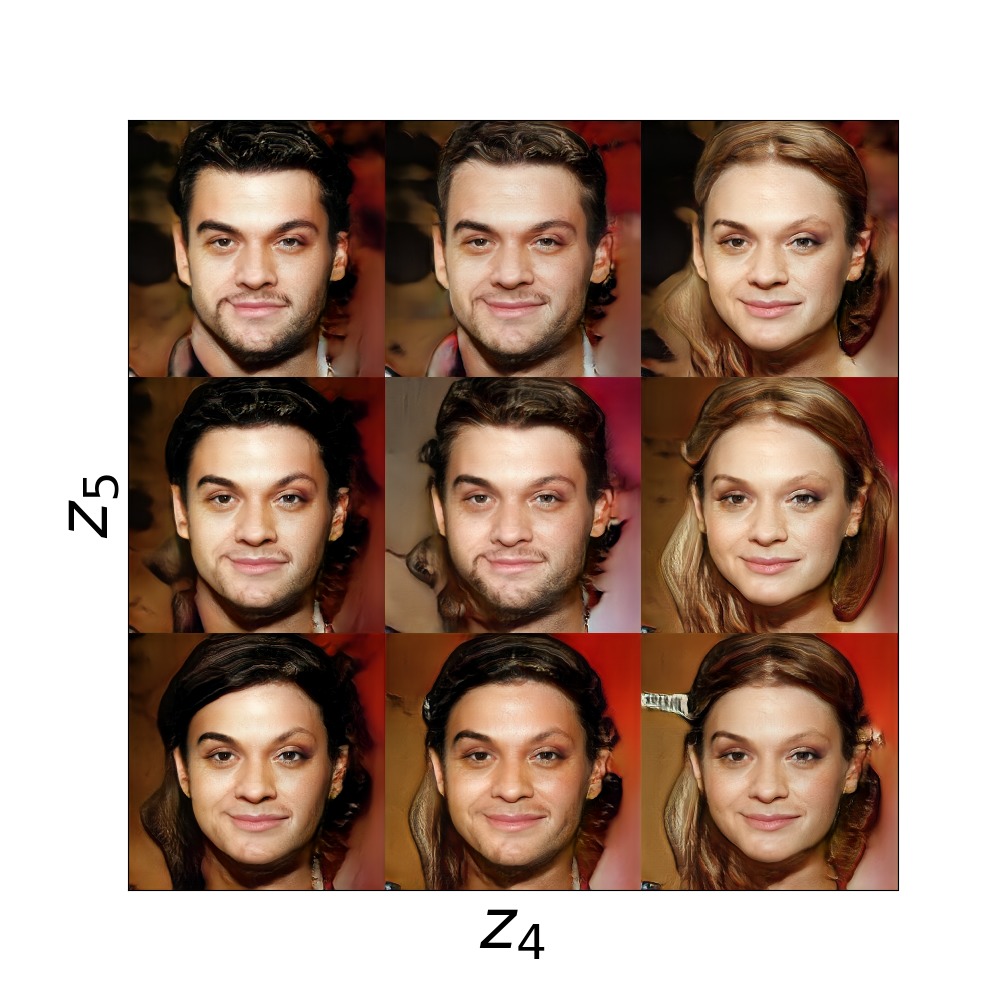}
    \end{tabularx}
    \vspace{-9pt}
    \subcaption{\normalsize VAE}
    \end{minipage}
    \\
    \begin{minipage}{\textwidth}
    \begin{tabularx}{\linewidth}{ccc}
    \includegraphics[width=0.24\linewidth, trim={1.7cm 1.0cm 3.0cm 3.1cm}, clip]{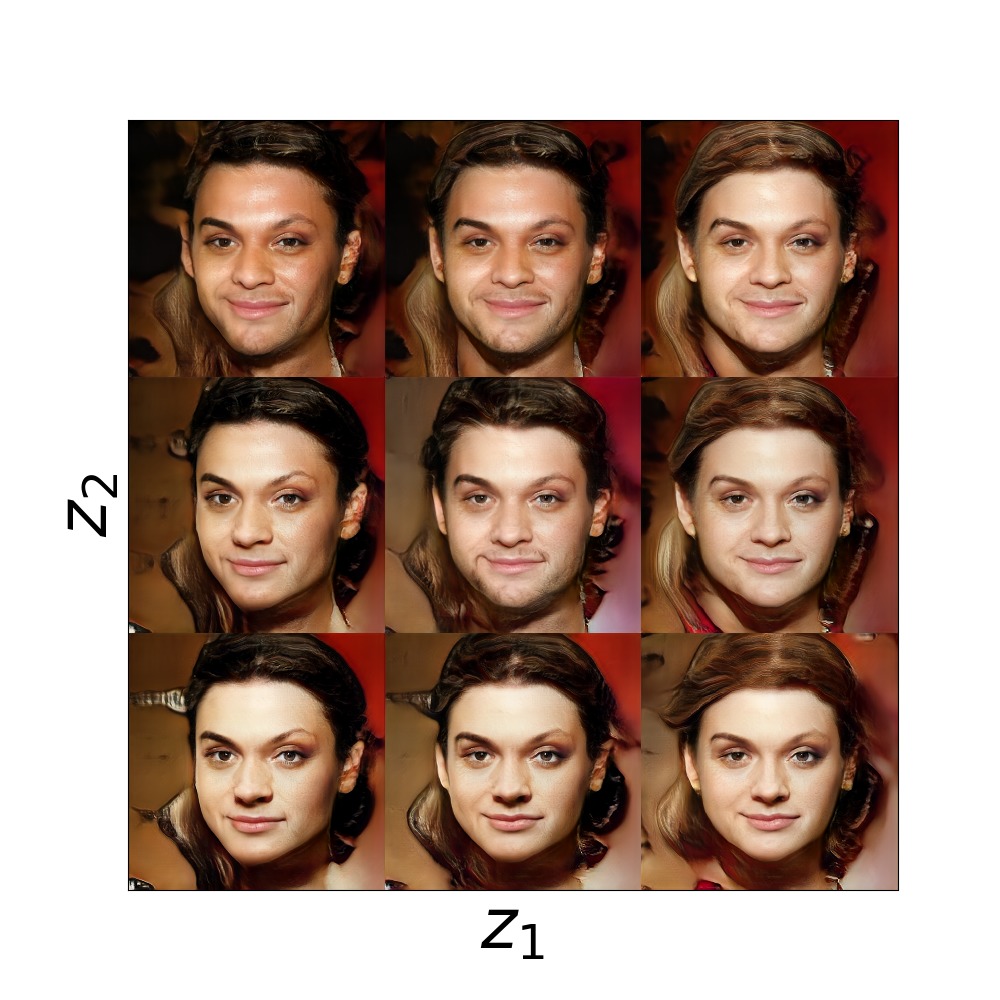}
    \includegraphics[width=0.24\linewidth, trim={1.7cm 1.0cm 3.0cm 3.1cm}, clip]{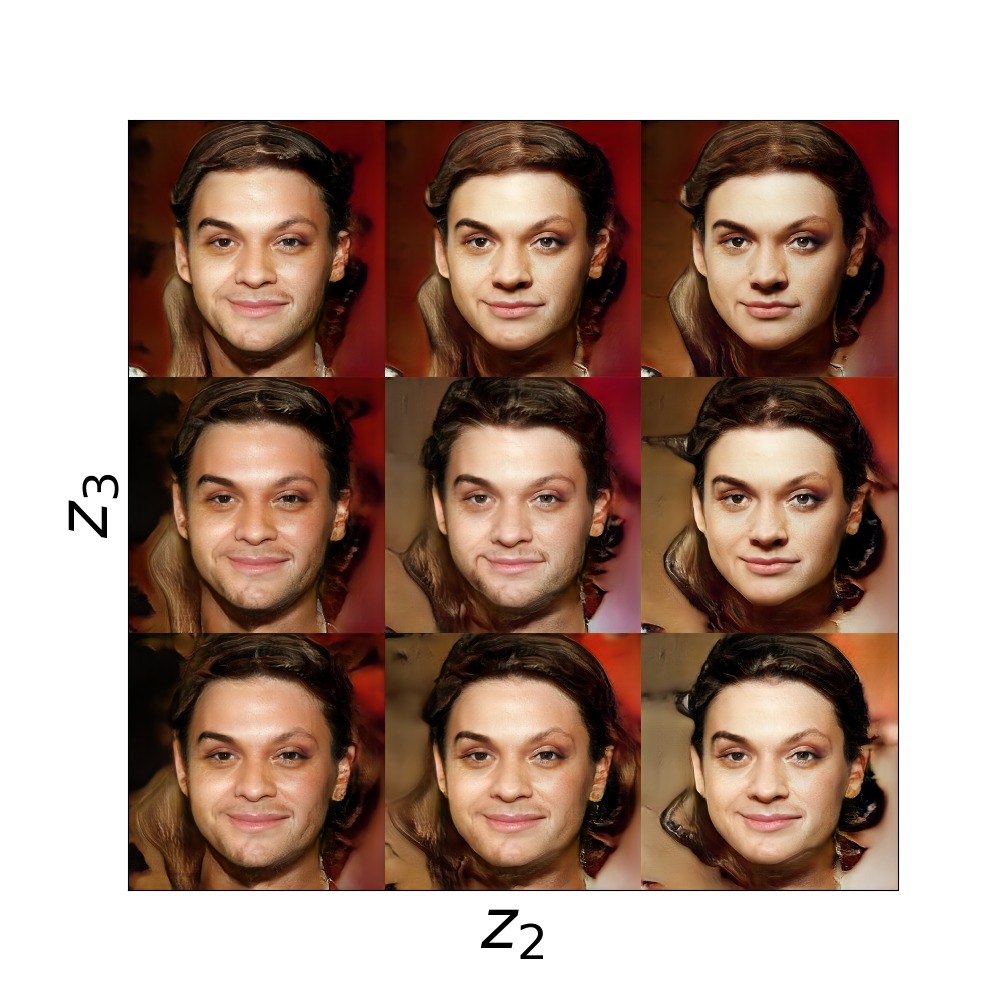}
    \includegraphics[width=0.24\linewidth, trim={1.7cm 1.0cm 3.0cm 3.1cm}, clip]{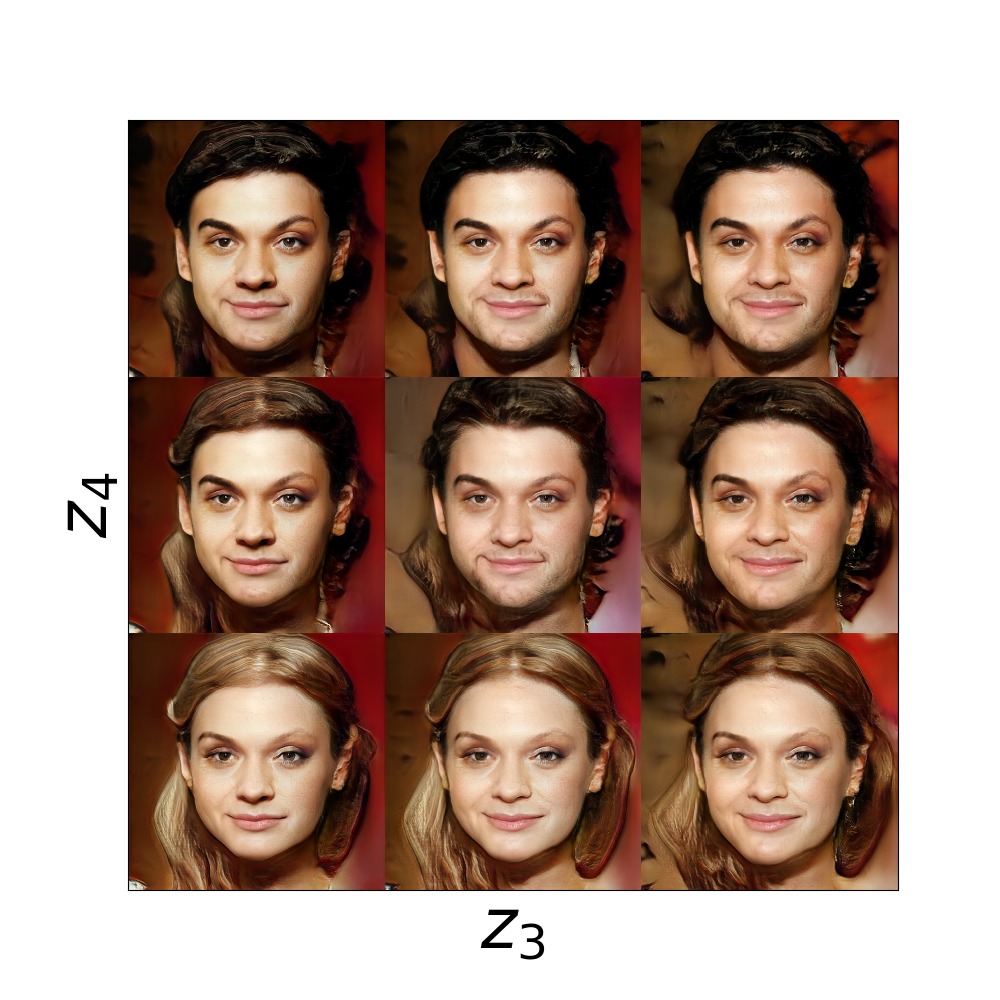}
    \includegraphics[width=0.24\linewidth, trim={1.7cm 1.0cm 3.0cm 3.1cm}, clip]{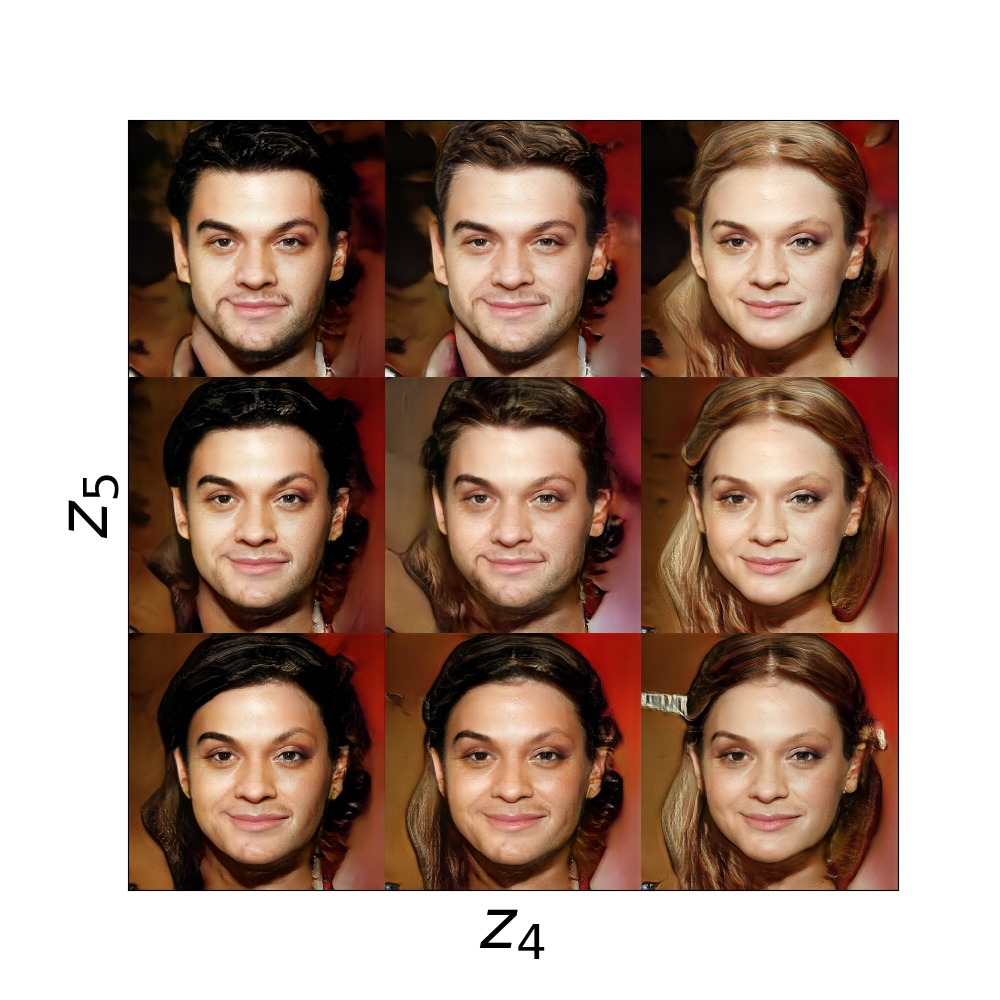}
    \end{tabularx}
    \vspace{-9pt}
    \subcaption{\normalsize $\beta$-VAE$_B$}
    \end{minipage}
    \\ 
    \begin{minipage}{\textwidth}
    \begin{tabularx}{\linewidth}{ccc}
    \includegraphics[width=0.24\linewidth, trim={1.7cm 1.0cm 3.0cm 3.1cm}, clip]{images/pca_gan/compare/factor/latent_z1z2.jpg}
    \includegraphics[width=0.24\linewidth, trim={1.7cm 1.0cm 3.0cm 3.1cm}, clip]{images/pca_gan/compare/factor/latent_z2z3.jpg}
    \includegraphics[width=0.24\linewidth, trim={1.7cm 1.0cm 3.0cm 3.1cm}, clip]{images/pca_gan/compare/factor/latent_z3z4.jpg}
    \includegraphics[width=0.24\linewidth, trim={1.7cm 1.0cm 3.0cm 3.1cm}, clip]{images/pca_gan/compare/factor/latent_z4z5.jpg}
    \end{tabularx}
    \vspace{-9pt}
    \subcaption{\normalsize FactorVAE}
    \end{minipage}
    \\ 
    \begin{minipage}{\textwidth}
    \begin{tabularx}{\linewidth}{ccc}
    \includegraphics[width=0.24\linewidth, trim={1.7cm 1.0cm 3.0cm 3.1cm}, clip]{images/pca_gan/compare/pcaae/latent_z1z2.jpg}
    \includegraphics[width=0.24\linewidth, trim={1.7cm 1.0cm 3.0cm 3.1cm}, clip]{images/pca_gan/compare/pcaae/latent_z2z3.jpg}
    \includegraphics[width=0.24\linewidth, trim={1.7cm 1.0cm 3.0cm 3.1cm}, clip]{images/pca_gan/compare/pcaae/latent_z3z4.jpg}
    \includegraphics[width=0.24\linewidth, trim={1.7cm 1.0cm 3.0cm 3.1cm}, clip]{images/pca_gan/compare/pcaae/latent_z4z5.jpg}
    \end{tabularx}
    \vspace{-9pt}
    \subcaption{\normalsize PCAAE}
    \end{minipage}
    \vspace{-9pt}
    \caption{Interpolation in latent space of five parameters of AE, VAE, $\beta$-VAE$_B$, FactorVAE and the PCAAE for the pre-train PGAN \cite{karras2017progressive}. Two components are adjusted along two axes, others are set to zeros. We can see that VAE, $\beta$-VAE$_B$ mixes hair colour along two components $z_1$ and $z_4$, $z_2$ and $z_5$ respectively. Head pose corresponds to the components $z_1$ and $z_4$ of FactorVAE. Whereas our method shows that each component of our proposed latent space represents one attribute of the generated images. For example, $z_1,z_2,z_3$ correspond to hair colours, head poses and gender.
    }
    \label{fig:pca_gan_ex1_appendix}
\end{figure}

We report a further comparison of the methods in Table \ref{tab:table_evaluation_gan_full} and Figure \ref{fig:pca_gan_ex1_appendix}. We can see that several methods utilise two or more components of the latent space to control one attribute. The proposed method PCAAE and its extension PCAWAE take only one component for each attribute. For instance, they choose the first component for hair colour, the second one for head pose and the third one for gender. 

\begin{figure}[b]
    \centering
    \tiny
    \begin{tabularx}{\linewidth}{ccccccccc}
    $z_1=-8$ & $z_1=-6$ & $z_1=-4$ & $z_1=-2$ & $z_1=0$ & $z_1=2$ & $z_1=4$ & $z_1=6$ & $z_1=8$ \\
    \includegraphics[width=0.08\linewidth]{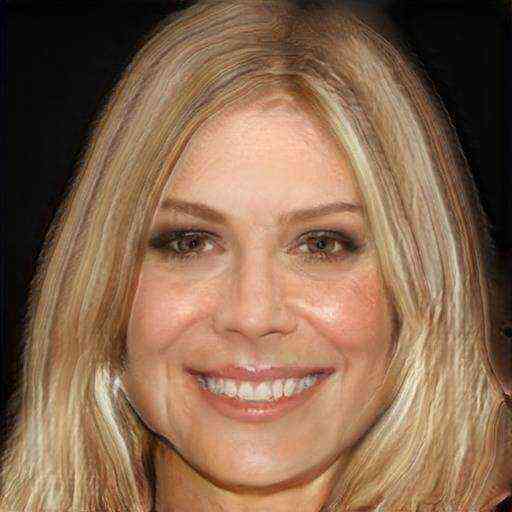}
    &
    \includegraphics[width=0.08\linewidth]{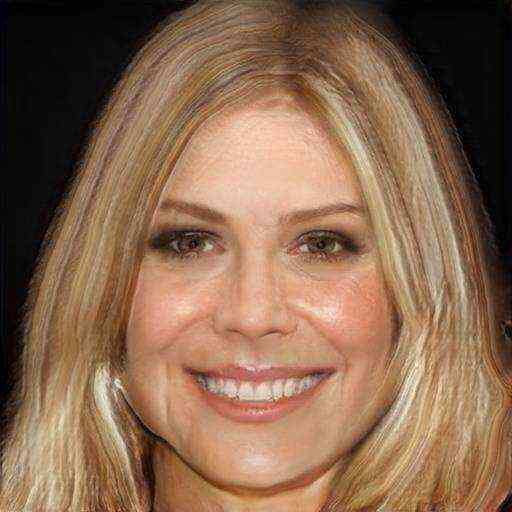}
    &
    \includegraphics[width=0.08\linewidth]{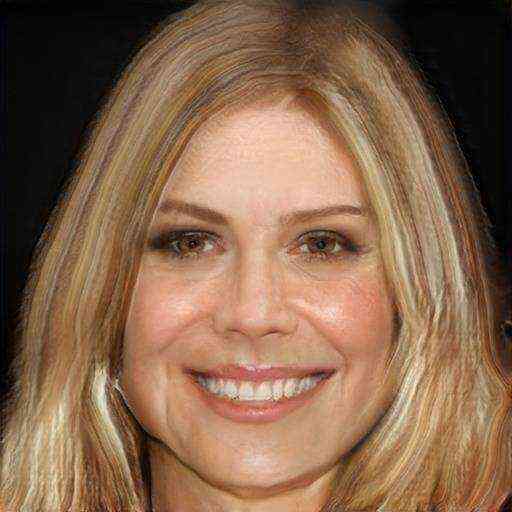}
    &
    \includegraphics[width=0.08\linewidth]{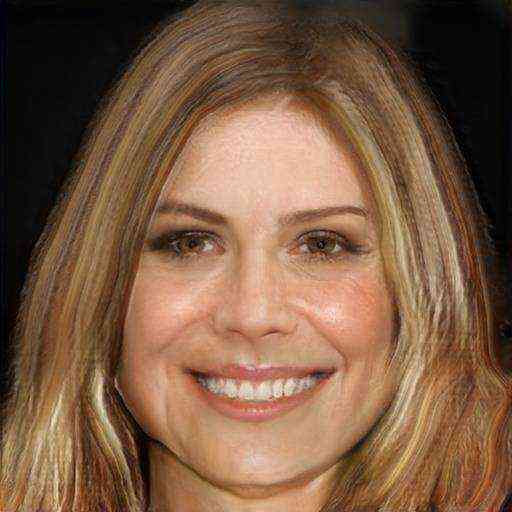}
    &
    \includegraphics[width=0.08\linewidth]{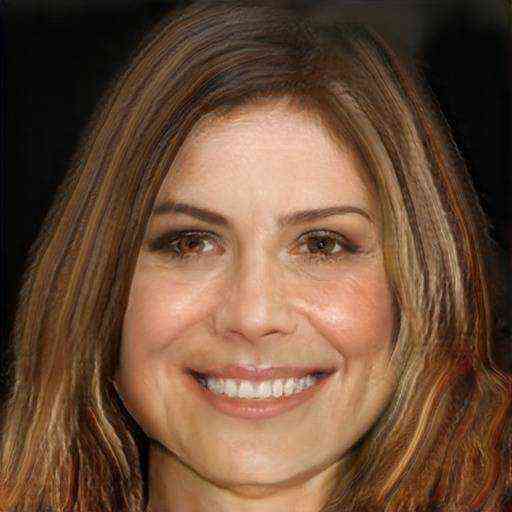}
    &
    \includegraphics[width=0.08\linewidth]{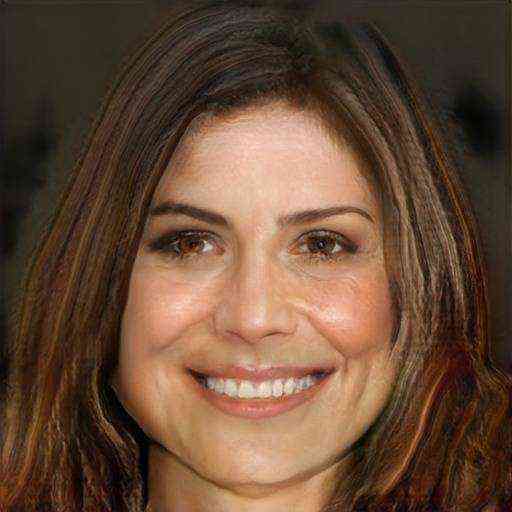}
    &
    \includegraphics[width=0.08\linewidth]{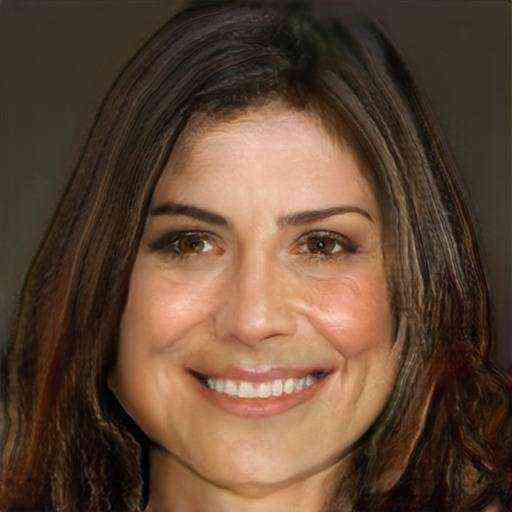}
    &
    \includegraphics[width=0.08\linewidth]{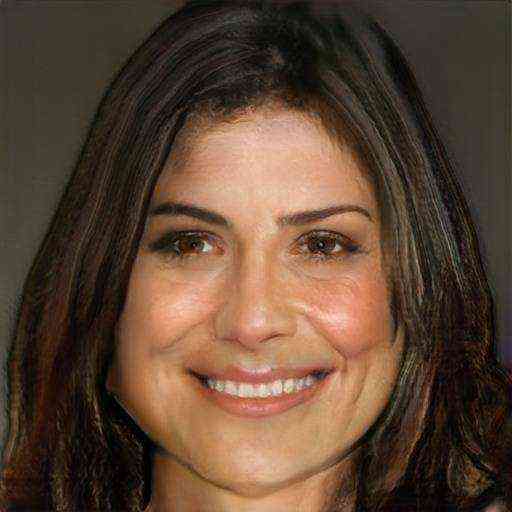}
    &
    \includegraphics[width=0.08\linewidth]{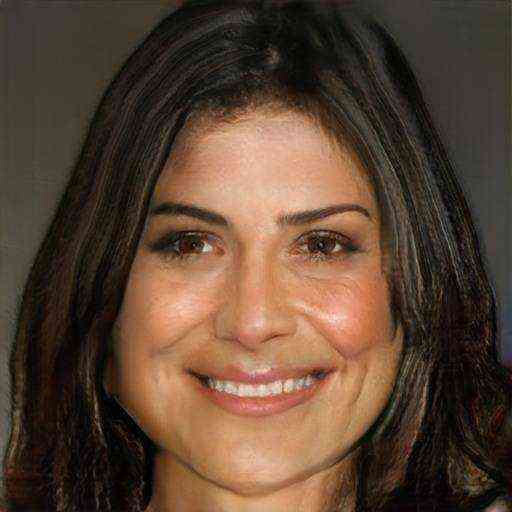}
    \\
    $z_2=-8$ & $z_2=-6$ & $z_2=-4$ & $z_2=-2$ & $z_2=0$ & $z_2=2$ & $z_2=4$ & $z_2=6$ & $z_2=8$ \\
    \includegraphics[width=0.08\linewidth]{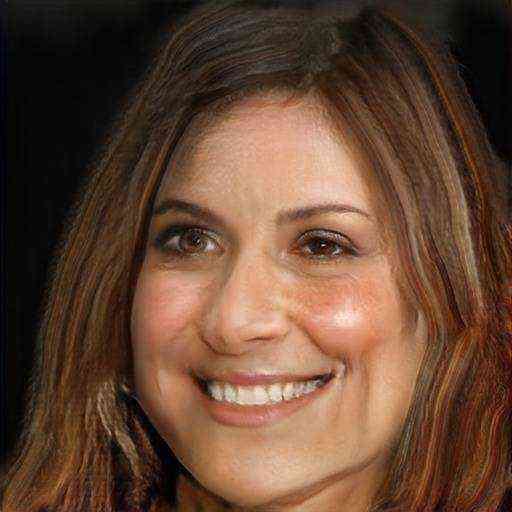}
    &
    \includegraphics[width=0.08\linewidth]{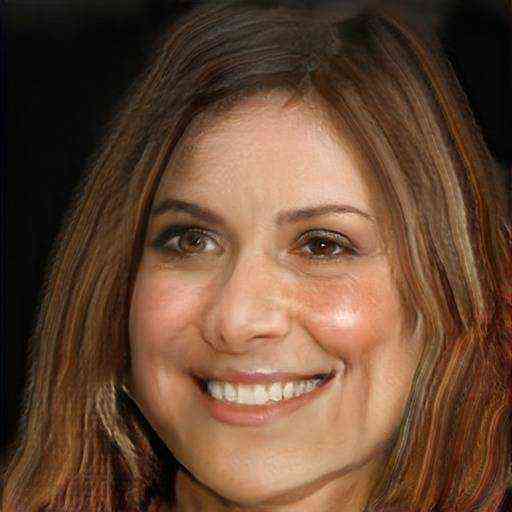}
    &
    \includegraphics[width=0.08\linewidth]{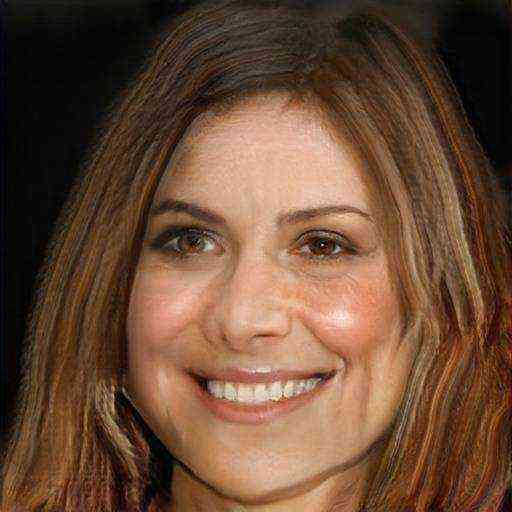}
    &
    \includegraphics[width=0.08\linewidth]{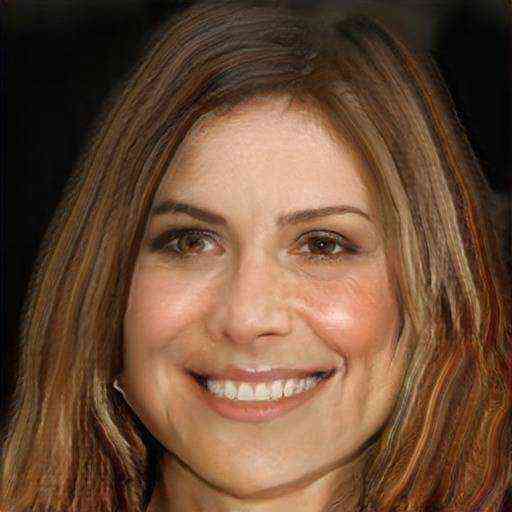}
    &
    \includegraphics[width=0.08\linewidth]{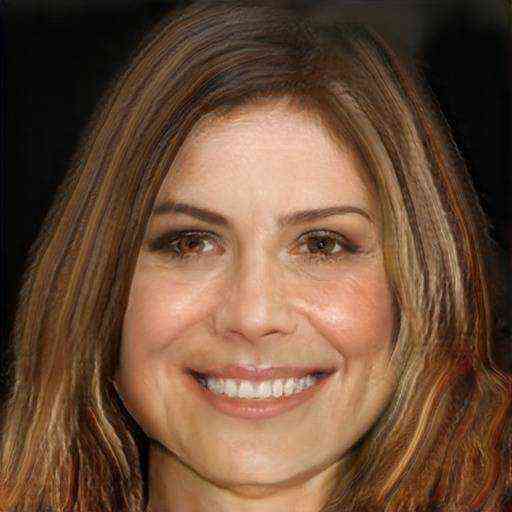}
    &
    \includegraphics[width=0.08\linewidth]{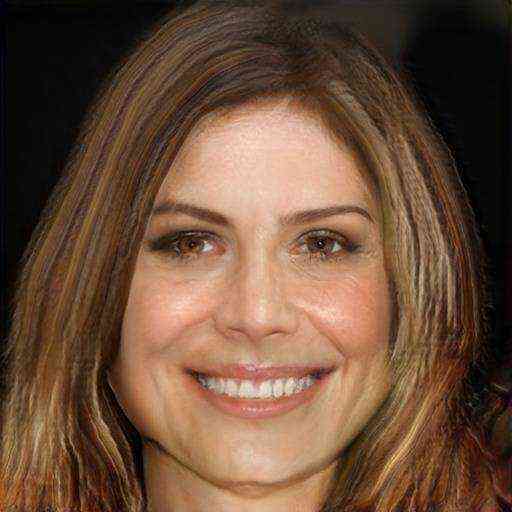}
    &
    \includegraphics[width=0.08\linewidth]{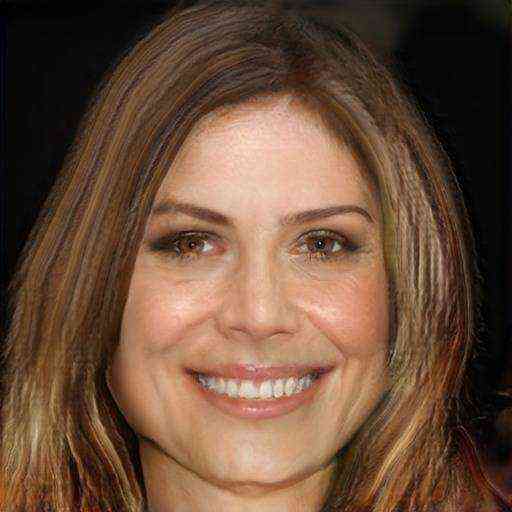}
    &
    \includegraphics[width=0.08\linewidth]{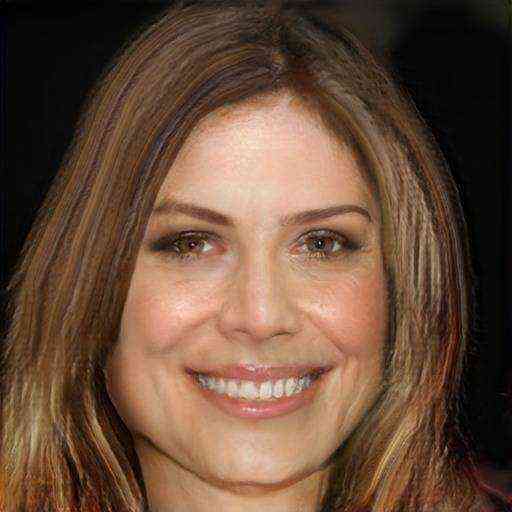}
    &
    \includegraphics[width=0.08\linewidth]{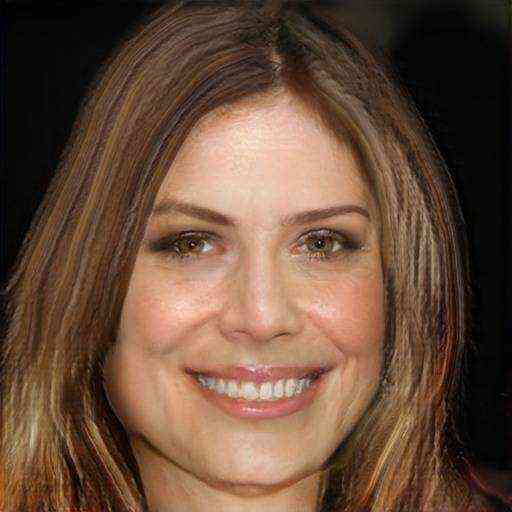}
    \\
    $z_3=-8$ & $z_3=-6$ & $z_3=-4$ & $z_3=-2$ & $z_3=0$ & $z_3=2$ & $z_3=4$ & $z_3=6$ & $z_3=8$ \\
    \includegraphics[width=0.08\linewidth]{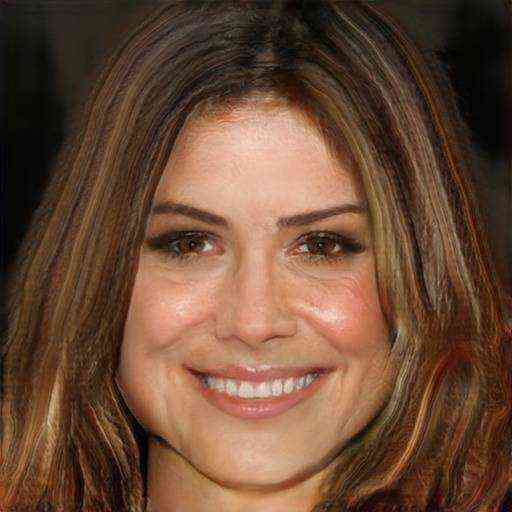}
    &
    \includegraphics[width=0.08\linewidth]{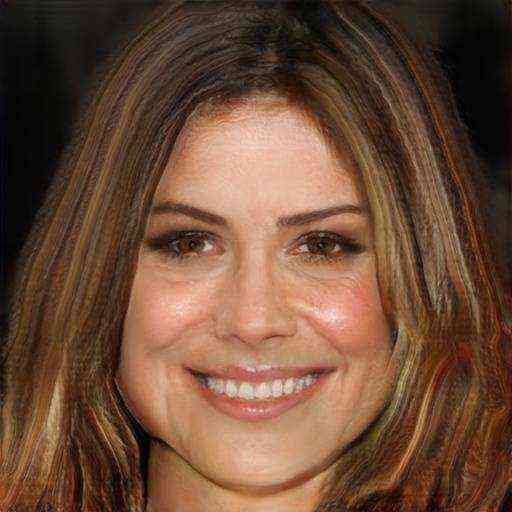}
    &
    \includegraphics[width=0.08\linewidth]{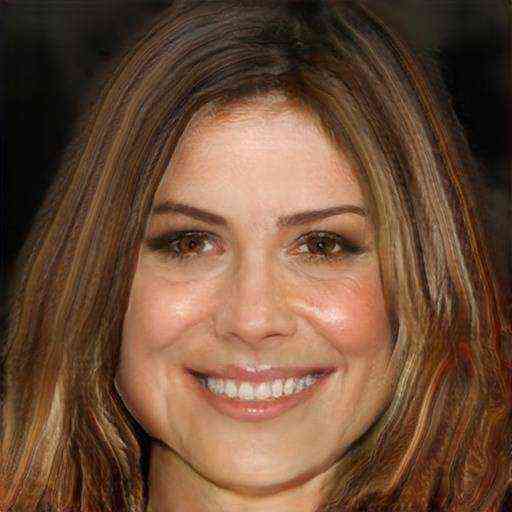}
    &
    \includegraphics[width=0.08\linewidth]{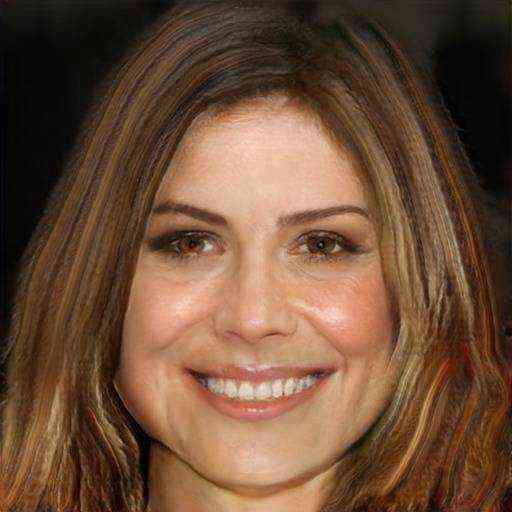}
    &
    \includegraphics[width=0.08\linewidth]{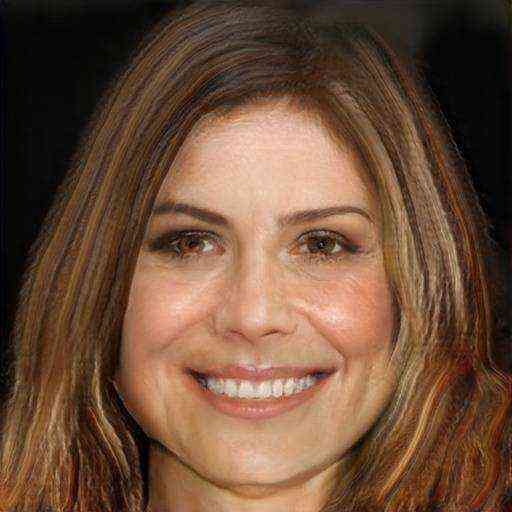}
    &
    \includegraphics[width=0.08\linewidth]{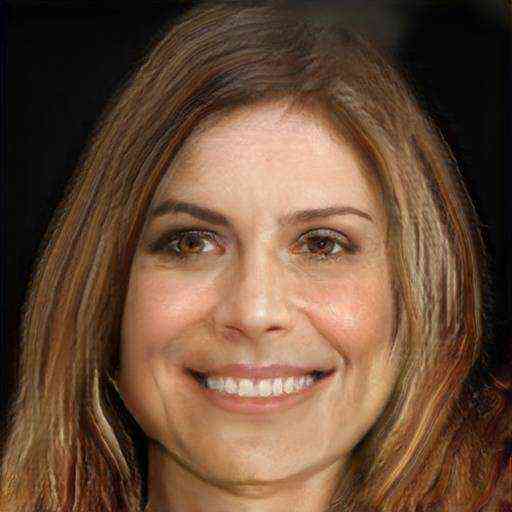}
    &
    \includegraphics[width=0.08\linewidth]{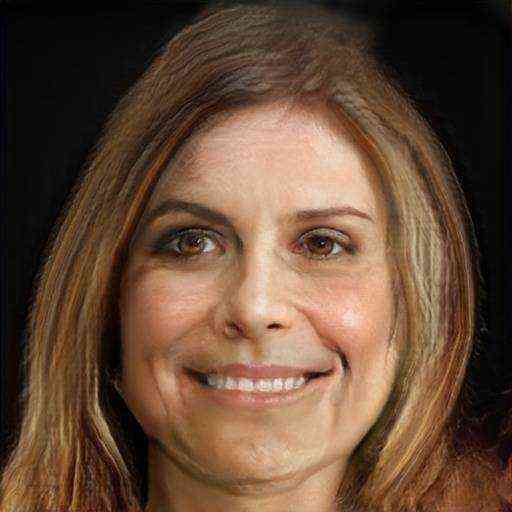}
    &
    \includegraphics[width=0.08\linewidth]{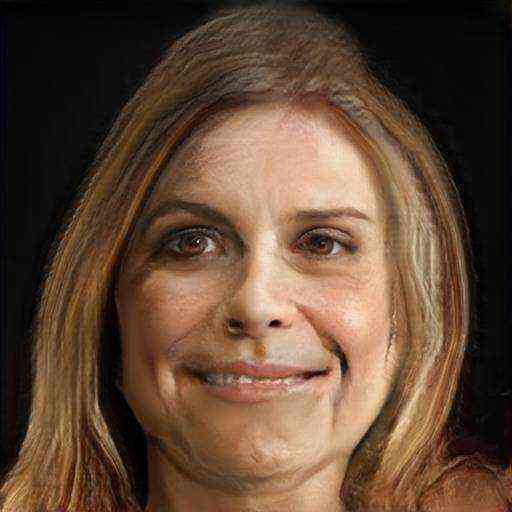}
    &
    \includegraphics[width=0.08\linewidth]{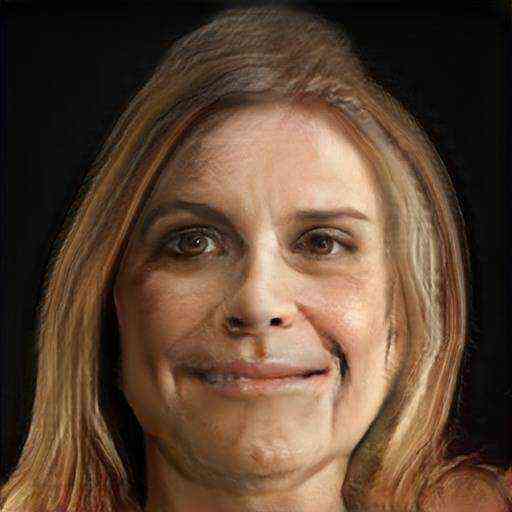}
    \\
    $z_4=-8$ & $z_4=-6$ & $z_4=-4$ & $z_4=-2$ & $z_4=0$ & $z_4=2$ & $z_4=4$ & $z_4=6$ & $z_4=8$ \\
    \includegraphics[width=0.08\linewidth]{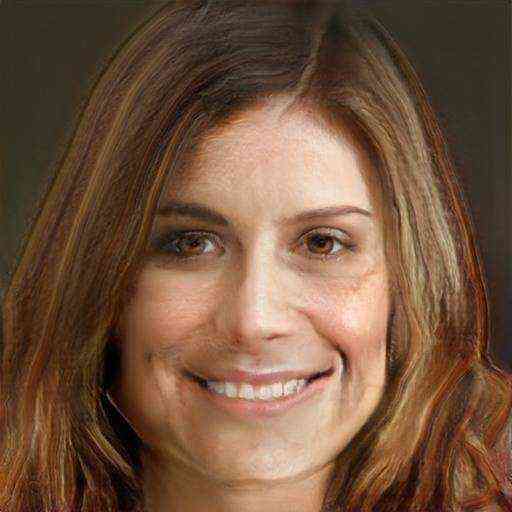}
    &
    \includegraphics[width=0.08\linewidth]{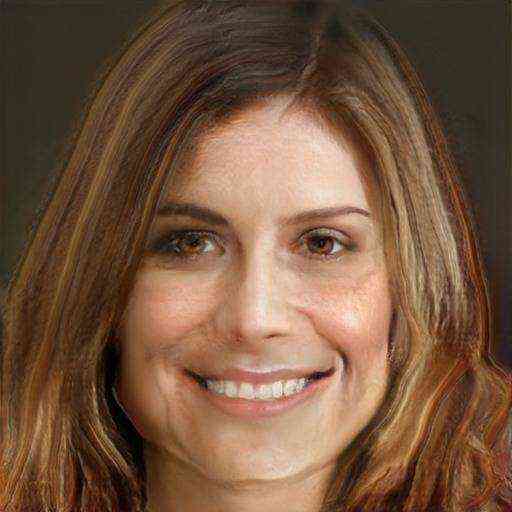}
    &
    \includegraphics[width=0.08\linewidth]{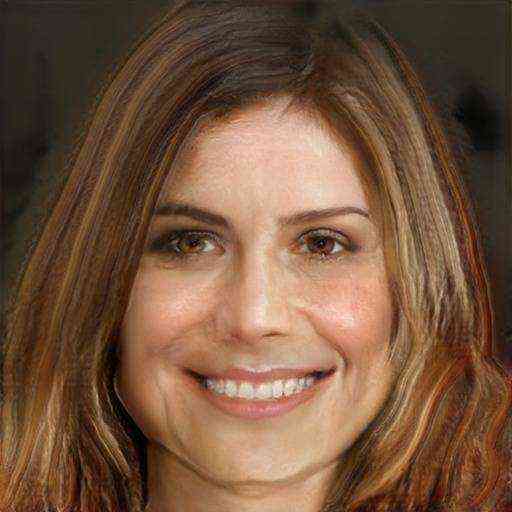}
    &
    \includegraphics[width=0.08\linewidth]{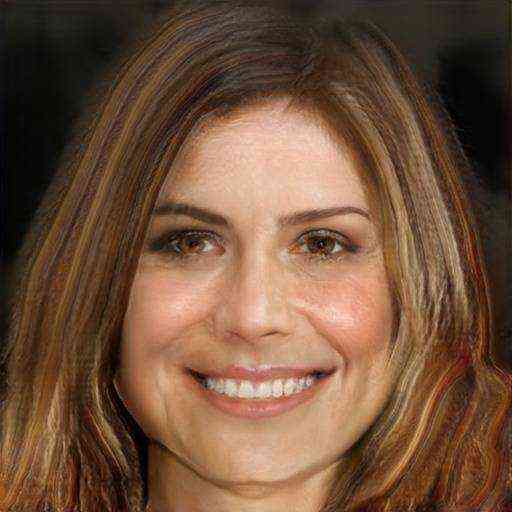}
    &
    \includegraphics[width=0.08\linewidth]{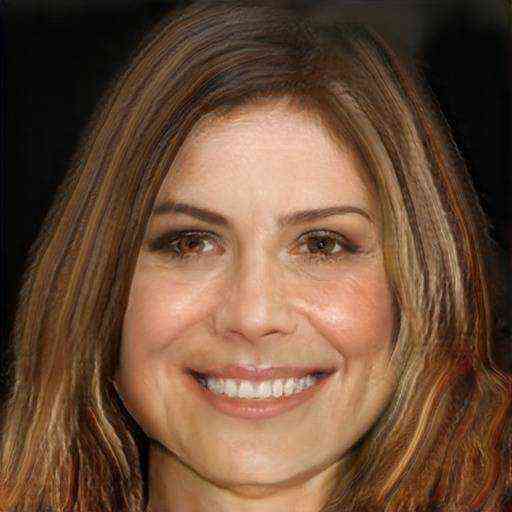}
    &
    \includegraphics[width=0.08\linewidth]{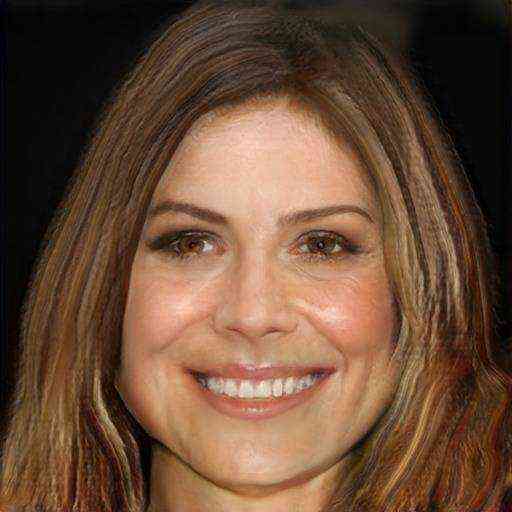}
    &
    \includegraphics[width=0.08\linewidth]{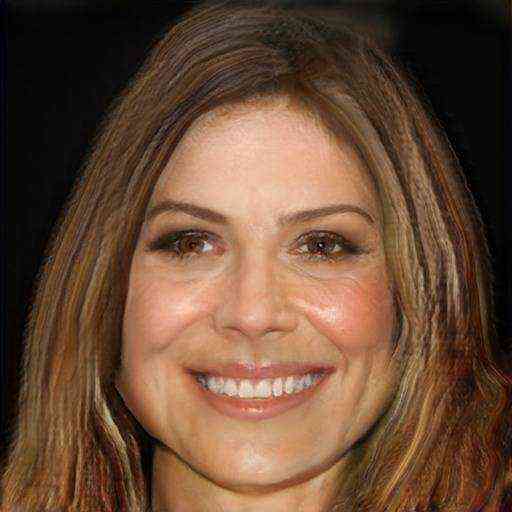}
    &
    \includegraphics[width=0.08\linewidth]{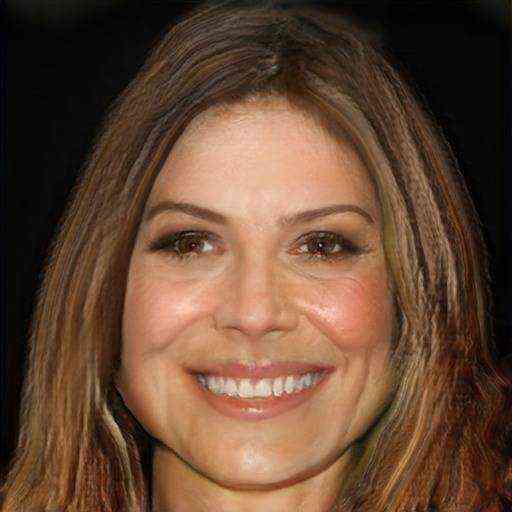}
    &
    \includegraphics[width=0.08\linewidth]{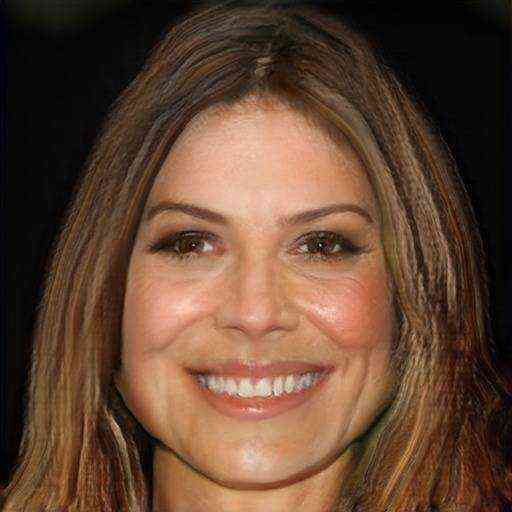}
    \\
    $z_5=-8$ & $z_5=-6$ & $z_5=-4$ & $z_5=-2$ & $z_5=0$ & $z_5=2$ & $z_5=4$ & $z_5=6$ & $z_5=8$ \\
    \includegraphics[width=0.08\linewidth]{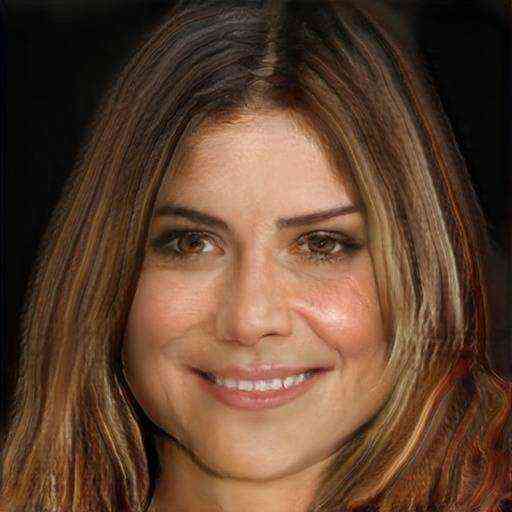}
    &
    \includegraphics[width=0.08\linewidth]{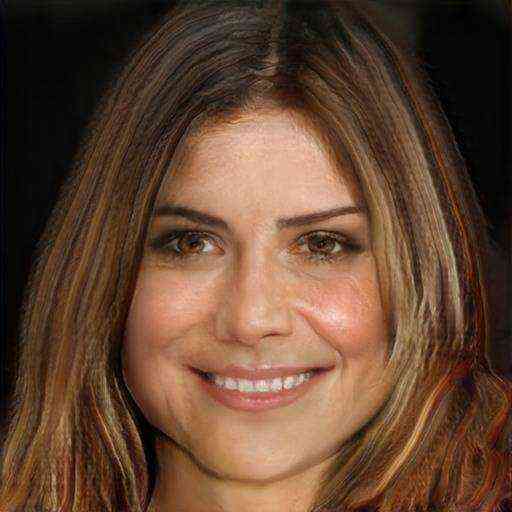}
    &
    \includegraphics[width=0.08\linewidth]{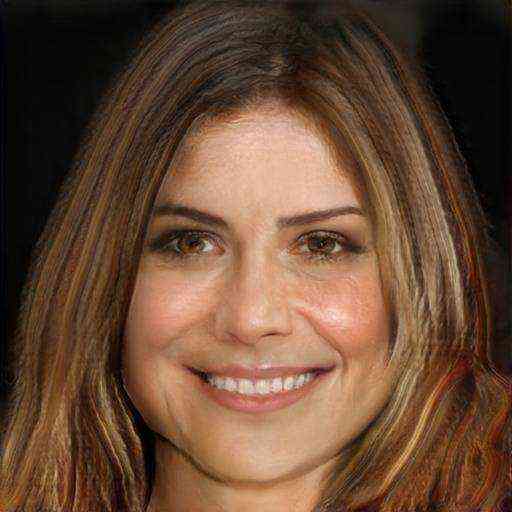}
    &
    \includegraphics[width=0.08\linewidth]{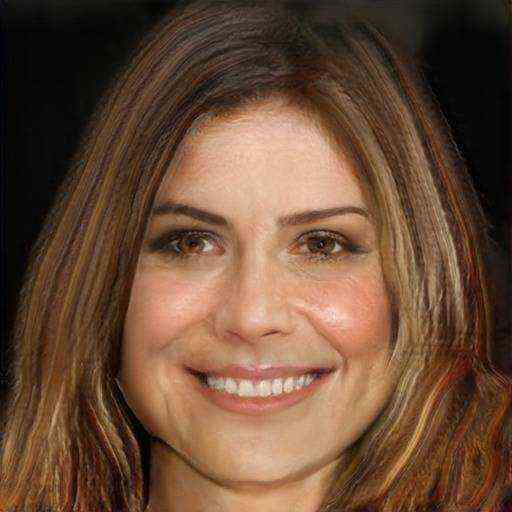}
    &
    \includegraphics[width=0.08\linewidth]{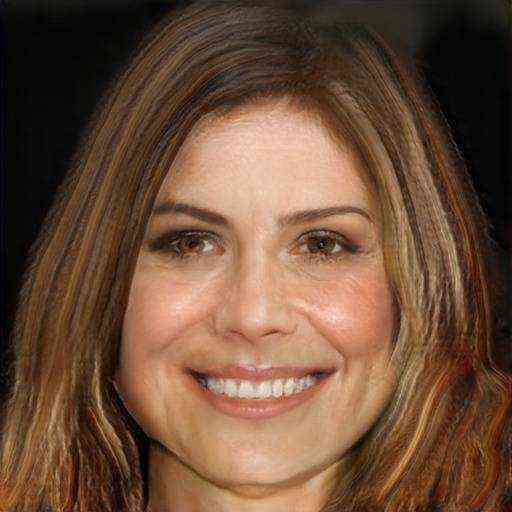}
    &
    \includegraphics[width=0.08\linewidth]{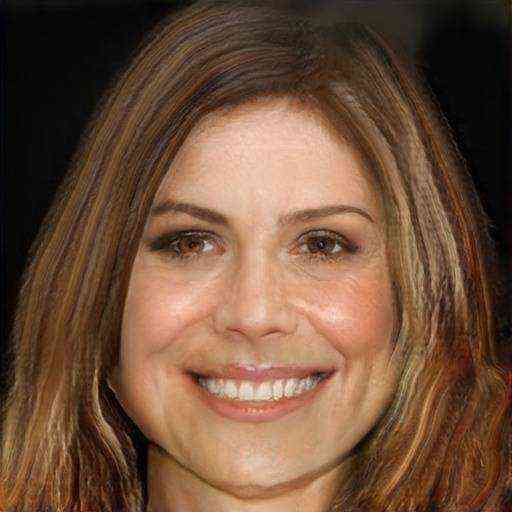}
    &
    \includegraphics[width=0.08\linewidth]{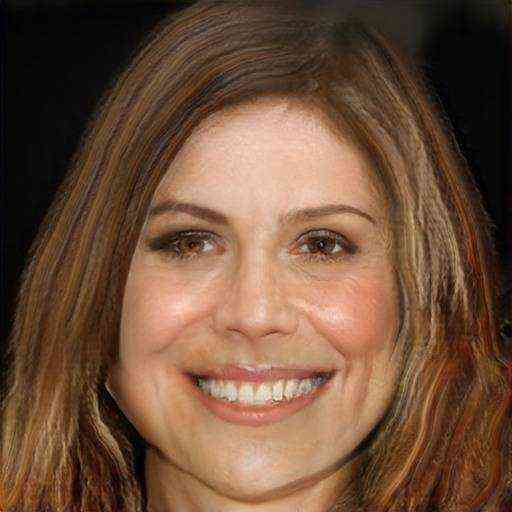}
    &
    \includegraphics[width=0.08\linewidth]{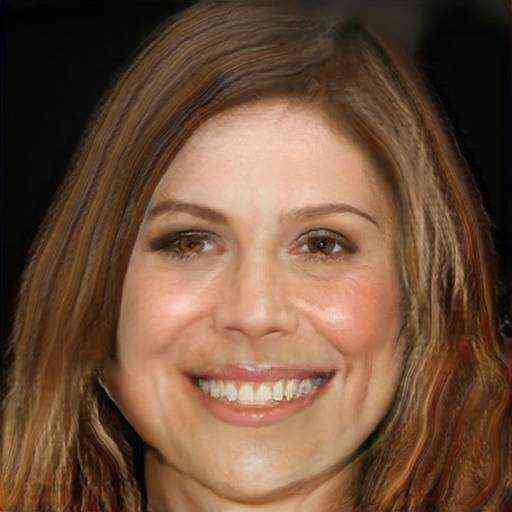}
    &
    \includegraphics[width=0.08\linewidth]{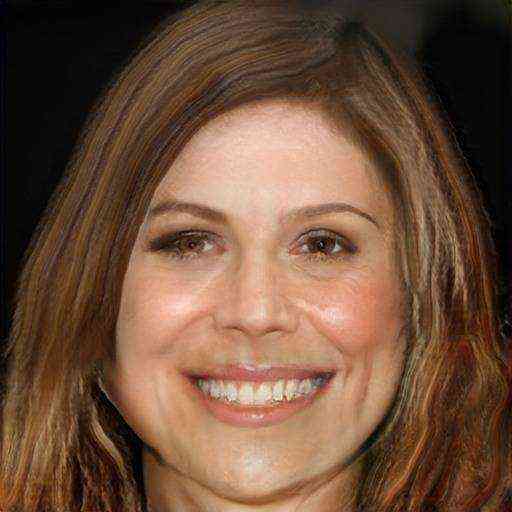}
    \end{tabularx}
    \caption{Further results of navigation in the latent space of the PCAAE for a pre-trained PGAN. We trained this PCAAE around the code $\overline{\eta}$ corresponding to the middle column. On each row, we have modified a single component (the other components are set to 0). We see that the component $z_1$ represents hair colour, while $z_2$ corresponds head poses, and in this case $z_3$ corresponds to gender and $z_5$ to the mouth posture. Note that in this experiment, we used a larger $\sigma$ to explore a larger part of the latent space.}
    \label{fig:pca_gan_ex2}
\end{figure}

We show another example of the automatic navigation of the latent space of the PGAN in Figure \ref{fig:pca_gan_ex2}. It is generated by training our PCAAE around the code generating the image at the middle of the three grids. We can see that for this example, the first component ($z_1$) corresponds to the hair colour from black to blond, the second one ($z_2$) controls the head poses and the third parameter ($z_3$) changes the gender.

\begin{figure}[htb]
    \centering
    \begin{minipage}{\textwidth}
    \begin{tabularx}{\linewidth}{c}
    \includegraphics[width=0.99\linewidth]{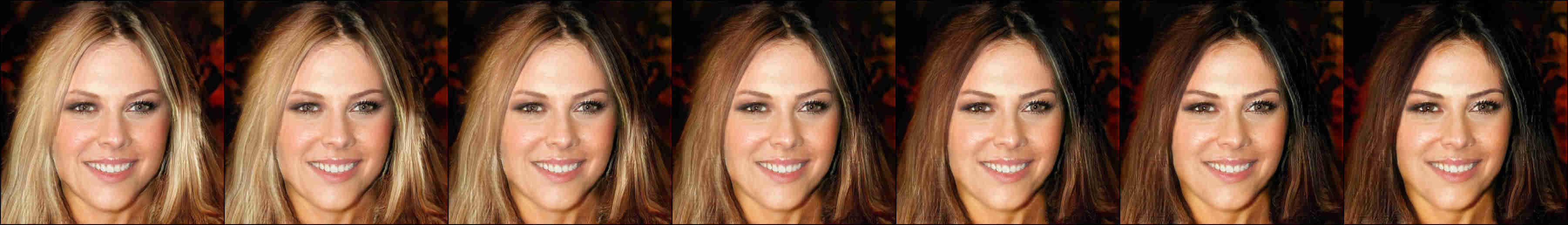} \\
    \hline     \vspace{-7pt}\\
    \includegraphics[width=0.99\linewidth]{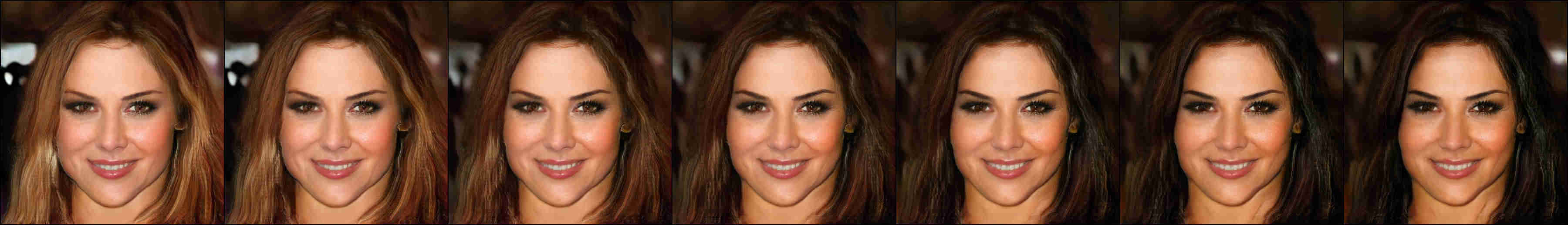} \\
    \includegraphics[width=0.99\linewidth]{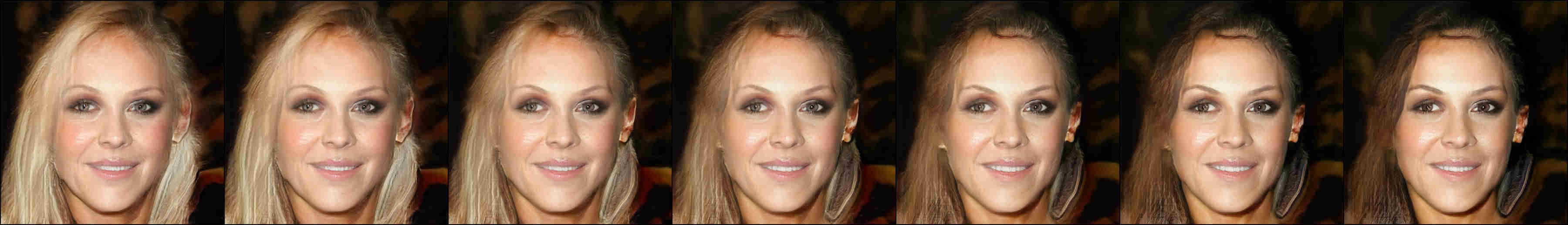} \\
    \includegraphics[width=0.99\linewidth]{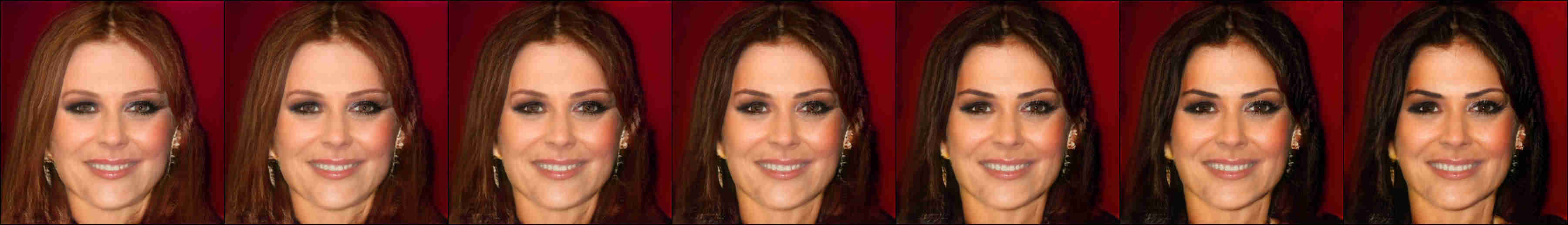} \\
    \end{tabularx}
    \vspace{-7pt}
    \subcaption{Hair colour controlling}
    \vspace{3pt}
    \end{minipage}
    \\
    \begin{minipage}{\textwidth}
    \begin{tabularx}{\linewidth}{c}
    \includegraphics[width=0.99\linewidth]{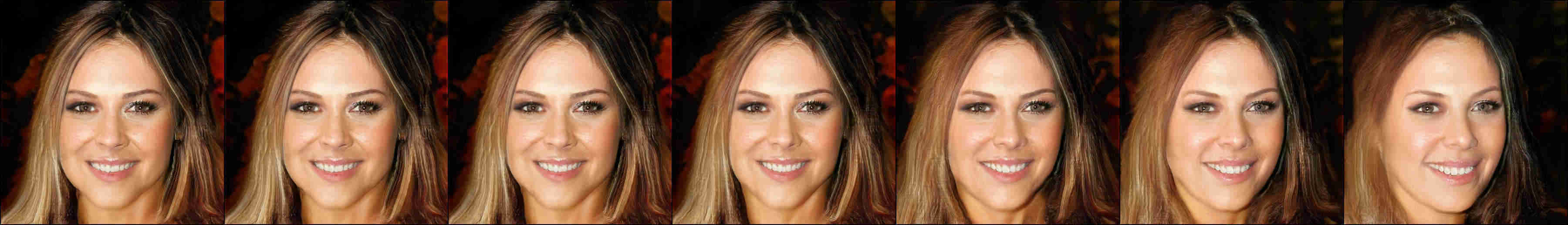} \\
    \hline \vspace{-7pt}\\
    \includegraphics[width=0.99\linewidth]{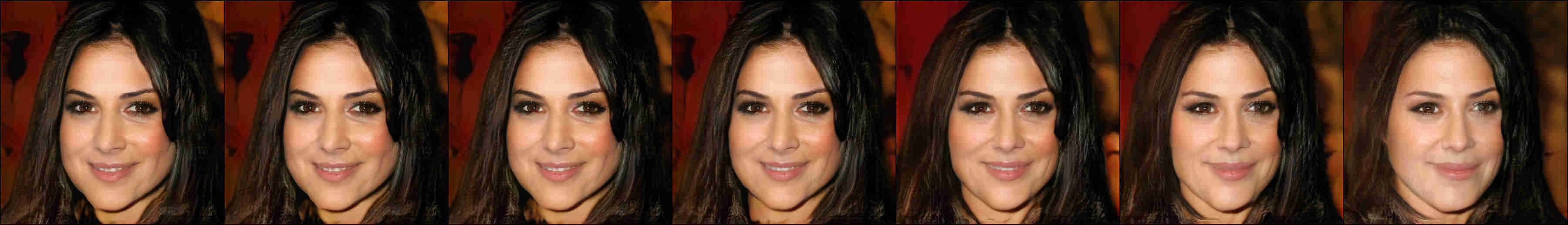} \\
    \includegraphics[width=0.99\linewidth]{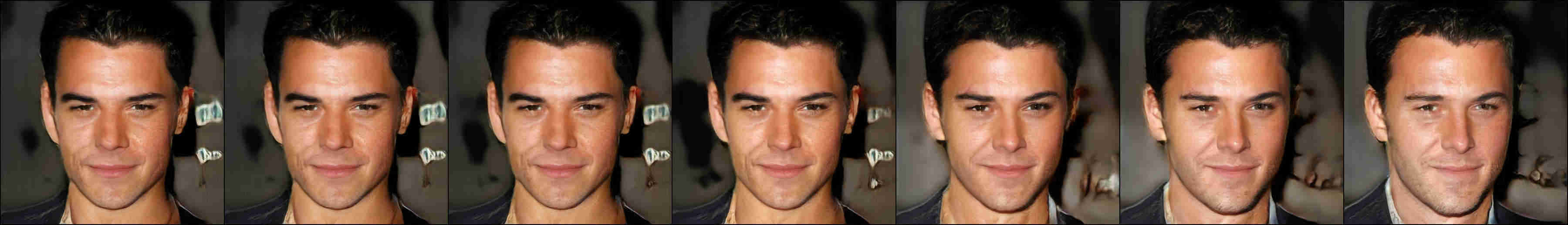} \\
    \includegraphics[width=0.99\linewidth]{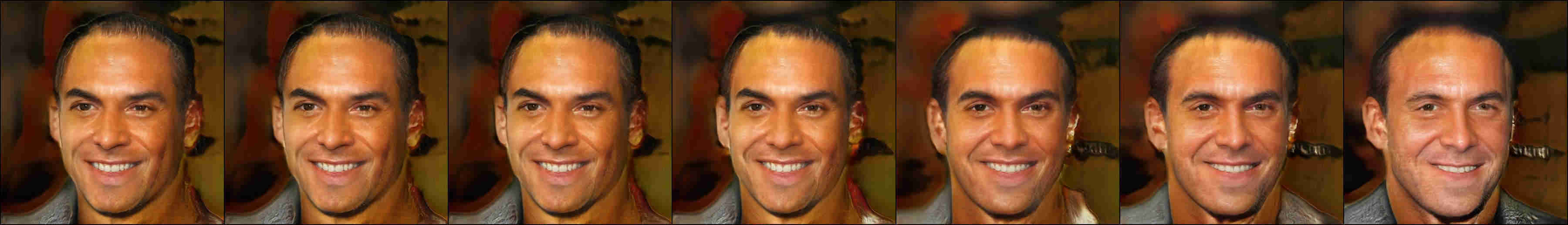} \\
    \end{tabularx}
    \vspace{-7pt}
    \subcaption{Head pose controlling}
    \end{minipage}    
    \caption{Application of PCAAE for PGAN: Transferring the learning attributes from the training code (the first row) to other testing codes (the last rows) for (a) hair colour and (b) head pose. The first row shows that we change the hair colour of the generated image from PGAN with respect to a training initial code $\overline{\eta}$ by adjusting the first component of the latent space of PCAAE. We test the trained PCAAE for PGAN from this code on other initial codes $\overline{\eta}$. We can see that the hair colours of generated images from other initial codes are also changed as those of the training code. Respectively, the head pose of generated images from testing code is also changed as those of the training code.}
    \label{fig:pca_gan_appli}
\end{figure}

In order to better visualise the results of the proposed method, we adjust two components which correspond to hair colour and head poses of generated images from the training initial code as shown in the first row of each sub figure of Figure \ref{fig:pca_gan_appli}. Then, we apply the trained model to other initial codes of the latent space of PGAN. We can see that the attribute of generated images from testing initial code also change as those of the training code (described as the last rows). 

Keep in mind that these results are obtained in a completely unsupervised manner with solely an $l_2$ norm as a guide. As said in Section 6 \textbf{``Limitations and future works''} of our main paper, it remains a challenge to find an universal PCAAE that could be used around each possible input of a GAN while maintaining the meaning of each dimension.   

{}

\begin{table}[h]
\begin{tabular}{|c|c|c|c|c|c|c|c|c|c|c|c|c|}
\hline  
\multirow{2}{*}{Co.}  &  \multicolumn{3}{|c|}{AE} &  \multicolumn{3}{|c|}{$\beta$-TCVAE} &  \multicolumn{3}{|c|}{FactorVAE} & \multicolumn{3}{|c|}{VAE}\\
\cline{2-13}
& HC & HP & GE & HC & HP & GE & HC & HP & GE & HC & HP & GE\\
\hline        
$Z_1$ & 0.20 & \textbf{0.53} & 0.02 & 0.35 & \textbf{0.80} & 0.04 & 0.07 & 0.46 & \textbf{0.53} & 0.35 & \textbf{0.81} & 0.07 \\
$Z_2$  & \textbf{0.36} & 0.21 & 0.04 & 0.05 & 0.26 & 0.25 & 0.04 & 0.17 & 0.03 & 0.05 & 0.24 & 0.24 \\
$Z_3$  & 0.28 & 0.36 & 0.23 & 0.13 & 0.04 & 0.04 & 0.09 & \textbf{0.66} & 0.37 & 0.13 & 0.02 & 0.02 \\
$Z_4$  & 0.20 & 0.19 & \textbf{0.58} & \textbf{0.53} & 0.19 & \textbf{0.53} & \textbf{0.70} & 0.11 & 0.14 & \textbf{0.59} & 0.23 & \textbf{0.57} \\
$Z_5$  & 0.34 & 0.21 & 0.49 & 0.07 & 0.15 & 0.33 & 0.01 & 0.28 & 0.12 & 0.07 & 0.11 & 0.35 \\
\hline
\multirow{2}{*}{Co.}  &  \multicolumn{3}{|c|}{$\beta$-VAE$_H$} &  \multicolumn{3}{|c|}{WAE} &  \multicolumn{3}{|c|}{PCAWAE} & \multicolumn{3}{|c|}{PCAAE}\\
\cline{2-13}
& HC & HP & GE & HC & HP & GE & HC & HP & GE & HC & HP & GE\\
\hline        
$Z_1$  & 0.33 & \textbf{0.80} & 0.05 & 0.19 & 0.33 & 0.01 & \textbf{0.73} & 0.01 & 0.30 &\textbf{0.70} & 0.03 & 0.14 \\
$Z_2$  & 0.02 &  0.29 &  0.22 & \textbf{0.54} & 0.10 &   \textbf{0.58} & 0.01 & \textbf{0.81} & 0.22 & 0.07 & \textbf{0.80} & 0.13 \\
$Z_3$  & 0.10 & 0.03 & 0.03 & 0.10 & \textbf{0.60} &  0.16 & 0.09 & 0.03 &  \textbf{0.37} & 0.16 & 0.15 & \textbf{0.56} \\
$Z_4$  & \textbf{0.56} & 0.19 & \textbf{0.57} & 0.36 & 0.37 & 0.36 & 0.01 & 0.20 &  0.23 & 0.03 & 0.24 & 0.05 \\
$Z_5$  & 0.07 & 0.13 & 0.33 & 0.26 & 0.29 & 0.08 & 0.07 & 0.22 & 0.19 & 0.05 & 0.22 & 0.09 \\
\hline
\end{tabular}
\vspace{2pt}
\caption{Quantitative evaluation of the correlation of latent components with high-level attributes. We have calculated the PCC between the latent components of AE, VAE-based methods, the proposed PCAAE and PCAWAE, and three attributes: head pose (HP), hair colour (HC) and gender (GE). We see that both PCAAE and the extended version PCAWAE choose to place the HC in the first component (the most important in terms of reconstruction error), followed by the HP and the GE. The components of PCAAE and PCAWAE are correlated with one attribute only, whereas other approaches have several components correlated with an attribute (HP with $Z_1$ and $Z_3$ of Factor-VAE, $Z_3$ of $\beta$-TCVAE and $\beta$-VAE controls HC and GE simultaneously, for example.)
}
\vspace{-10pt}
\label{tab:table_evaluation_gan_full}
\end{table}

{}

\end{document}